\documentclass[10pt,twocolumn,letterpaper]{article}
\usepackage{silence}
\usepackage{iccv}
\usepackage[pagebackref=true,breaklinks=true,colorlinks,bookmarks=false]{hyperref}
\usepackage{times}
\usepackage{epsfig}
\usepackage{graphicx}
\usepackage{amsmath}
\usepackage{amssymb}
\usepackage{pifont}
\usepackage[export]{adjustbox}
\usepackage{multirow}
\usepackage[font=footnotesize,skip=0pt]{caption}
\usepackage{subcaption}
\usepackage{xcolor,colortbl}
\usepackage{makecell}
\usepackage{graphicx,verbatimbox}

\usepackage[capitalize]{cleveref}
\usepackage{dsfont}
\crefname{section}{Sec.}{Secs.}
\Crefname{section}{Section}{Sections}
\Crefname{table}{Table}{Tables}
\crefname{table}{Tab.}{Tabs.}

\newcommand{\cmark}{\ding{51}}%
\newcommand{\xmark}{\ding{55}}%
\iccvfinalcopy 

\def\iccvPaperID{4712} 

\ificcvfinal\fi

\begin{document}

\title{Toward Unsupervised Realistic Visual Question Answering}

\author{%
  Yuwei Zhang\thanks{The first two authors contributed equally to this work.} \quad Chih-Hui Ho\footnotemark[1] \quad Nuno Vasconcelos \\
  Department of Electrical and Computer Engineering\\
  University of California, San Diego\\
  \texttt{\{yuz163,chh279, nvasconcelos\}@ucsd.edu} \\
}

\maketitle

\ificcvfinal\thispagestyle{empty}\fi

\begin{abstract}
The problem of realistic VQA (RVQA), where a model has to reject unanswerable questions (UQs) and answer answerable ones (AQs), is studied. We first point out 2 drawbacks in current RVQA research, where (1) datasets contain too many unchallenging UQs and (2) a large number of annotated UQs are required for training. To resolve the first drawback, we propose a new testing dataset, RGQA, which combines AQs from an existing VQA dataset with around 29K human-annotated UQs. These UQs consist of both fine-grained and coarse-grained image-question pairs generated with 2 approaches: CLIP-based and Perturbation-based. To address the second drawback, we introduce an unsupervised training approach. This combines pseudo UQs obtained by randomly pairing images and questions, with an RoI Mixup procedure to generate more fine-grained pseudo UQs, and model ensembling to regularize model confidence. Experiments show that using pseudo UQs significantly outperforms RVQA baselines. RoI Mixup and model ensembling further increase the gain. Finally, human evaluation reveals a performance gap between humans and models, showing that more RVQA research is needed.
\end{abstract}
\vspace{-10pt}

\section{Introduction}
\begin{figure}[t]
\centering
\setlength{\tabcolsep}{2pt}
\begin{tabular}{cc}
\includegraphics[width=0.8 \linewidth]{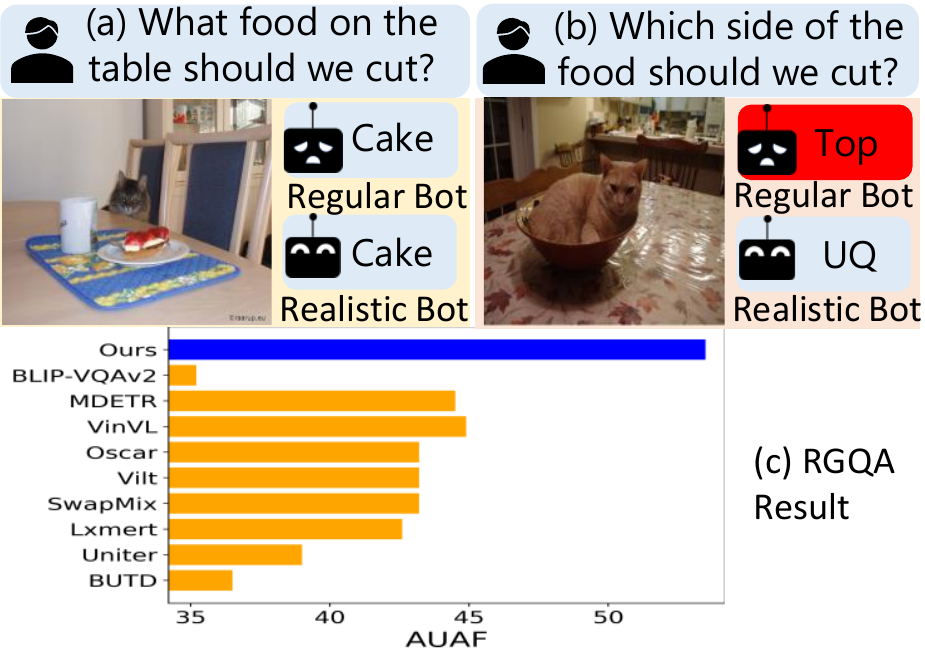}  
\end{tabular}

\caption{Realistic VQA. In VQA, a vision system answers a question by inspection of an image. However, existing approaches have no awareness if the question is {\it answerable} (AQ), such as in (a), or {\it unanswerable} (UQ), as in (b). A realistic VQA system only answers AQs. (c) RVQA performance of prior (yellow) vs. proposed (blue) models.}
\label{fig:intro}
\end{figure}

\vspace{-3pt}
\label{sec:intro}
Visual Question Answering (VQA) is a challenging task that requires a machine to understand a question in natural language, perceive an image, and produce an answer.
Despite extensive research in VQA~\cite{Antol15,Goyal2017MakingTV,hudson2018gqa,marino2019ok,Johnson2017CLEVRAD,tan2019lxmert,uniter,vinvl}, little attention has been given to VQA robustness. In this work, we consider robustness to {\it unanswerable questions} (UQs), which cannot be answered by image inspection, as in Fig.~\ref{fig:intro}(b). This is opposed to the traditional answerable questions (AQ), such as in Fig.~\ref{fig:intro}(a).

Lack of robustness to UQs is problematic because, in the absence of image information, the VQA system frequently resorts to the answer statistically most correlated with the question. In the figure, the absence of food in (b) entices the robot to pick the answer corresponding to the ``side of food" most commonly ``cut" in the dataset, which happens to be the ``top" (perhaps because the dataset is rich in cake images). The problem is that a decision by the robot to act on this answer would be catastrophic for the cat in the scene. More generally, the inability to reject UQs signals a deeper perceptual deficiency and exposes VQA systems to attacks. 

Vulnerability to UQs can create safety hazards for indoor robots~\cite{anderson2018navigation} or assistants for the visually impaired~\cite{gurari2018vizwiz} and reduces user trust in VQA models (see appendix for various examples from the recent large-scale BLIP model~\cite{li2022blip}). When faced with a UQ, the VQA system should refuse to answer or ask for more information. More precisely, it should assess the question,  decide to (a) ``accept" or ``reject" it, and only (b) answer the accepted questions.
Since this resembles the idea of a ``realistic model" for classification~\cite{Wu20DeepRTC,pei18}, we denote it {\it realistic VQA\/} (RVQA). 

Although some prior works have addressed RVQA, existing formulations are not conducive to practical RVQA systems, for three reasons. 
First, existing formulations address the {\it supervised\/} training of RVQA models. This, however, requires a significant number of annotated UQs~\cite{ray2016vtfq,mahendru2017qrpe,gurari2018vizwiz}. The collection of a set of annotated UQs large enough to train a modern VQA network is expensive, frequently not even plausible. This is compounded by the existence of many types of UQs: training on one type does not guarantee generalization to another. Second, prior datasets generate UQs by randomly paring images and questions from an existing VQA dataset~\cite{ray2016vtfq,mahendru2017qrpe,kafle2017tdiuc,toor2017c2vqa}. This, however, tends to produce obviously unrelated pairs of images and questions with low semantic similarity, that are easy to reject. In the real world, RVQA models must be able to handle both simple and challenging UQs. Finally,  the VQA datasets that support RVQA, such as VizWiz~\cite{gurari2018vizwiz}, are designed for a specific application domain, frequently containing images with few objects. This prevents the modeling of complex image-question relationships.
To address these drawbacks, we consider the problem of {\it unsupervised RVQA\/}. We start by curating a new evaluation dataset for this task,
based on \textit{testdev} set of the widely used graph-based VQA (GQA) dataset~\cite{hudson2018gqa}, a challenging dataset that involves multi-step reasoning. The new dataset, denoted as {\it realistic GQA\/} (RGQA),
is composed of $26,591$ AQs in the \textit{testdev} set of GQA and $29,046$ additional human-annotated UQs.
To penalize RVQA models that overfit on a specific type of UQs, we generate candidate UQ by two methods. {\it CLIP-based\/} UQ generation produces candidate UQs by retrieving questions sorted by CLIP~\cite{radford2021clip} similarity score between image and question. {\it Perturbation-based\/} (PT-based) UQ generation perturbs the object, attribute, and relation phrases in a question to change its meaning. 
For each method, we further generate a set of easy and a set of hard candidate UQs, leading to a total of four RGQA subsets. All candidate UQs are finally annotated by humans, to guarantee they are unanswerable.

Since each AQ in RGQA is complemented by its answer, the dataset enables measuring the accuracy of both AQ/UQ detection and VQA accuracy.
For this, we propose the ACC-FPR curve~\cite{dhamija2018agnostophobia}, a joint measure of VQA accuracy for AQs and UQ rejection performance.
This is complemented by introducing 3 new unsupervised RVQA methods that establish a set of baselines for future RVQA work.
These are classifiers with a binary output per class, which elicit a rejection when all class outputs are below a threshold. Three methods differ in training strategy and are shown capable of producing RVQA models that both reject UQs and answer AQs correctly, outperforming prior RVQA methods.

The first is to train the classifier with pseudo UQs, obtained by randomly pairing images and questions. This suffers from the fact that pseudo UQs are noisy and not always challenging. 
The second improves the sampling of image-question pairs, by using a RoI Mixup strategy to encourage the model to spot fine-grained mismatches between image and question during training. The third address the limitations of random sampling at the classifier output, by  ensembling multiple RVQA models.  
Experiments show that all strategies enhance RVQA performance and that they can be combined to achieve  best results. 
As shown in Fig.~\ref{fig:intro}(c), this combination (blue) significantly exceeds the performance of existing VQA models (yellow) under the joint objective of rejecting UQs and correctly answering AQs.


Overall, three contributions are made to VQA.  First, we introduce RGQA, a new challenging testing dataset for evaluating RVQA. It contains both fine- and coarse-grained image-question pairs which better align with real-world scenarios than previous datasets. Second, we propose an unsupervised training strategy that uses free pseudo UQs, combining random sampling, RoI Mixup, and model ensembling. Finally, extensive experiments demonstrate the effectiveness of the proposed methods over prior RVQA methods. We also show that the proposed models under-perform humans, which encourages future work in the RVQA problem. Code and dataset will be released upon publication.

\begin{figure*}
    \centering
    \begin{tabular}[t]{cccc}
      \includegraphics[width=.2\linewidth,valign=t]{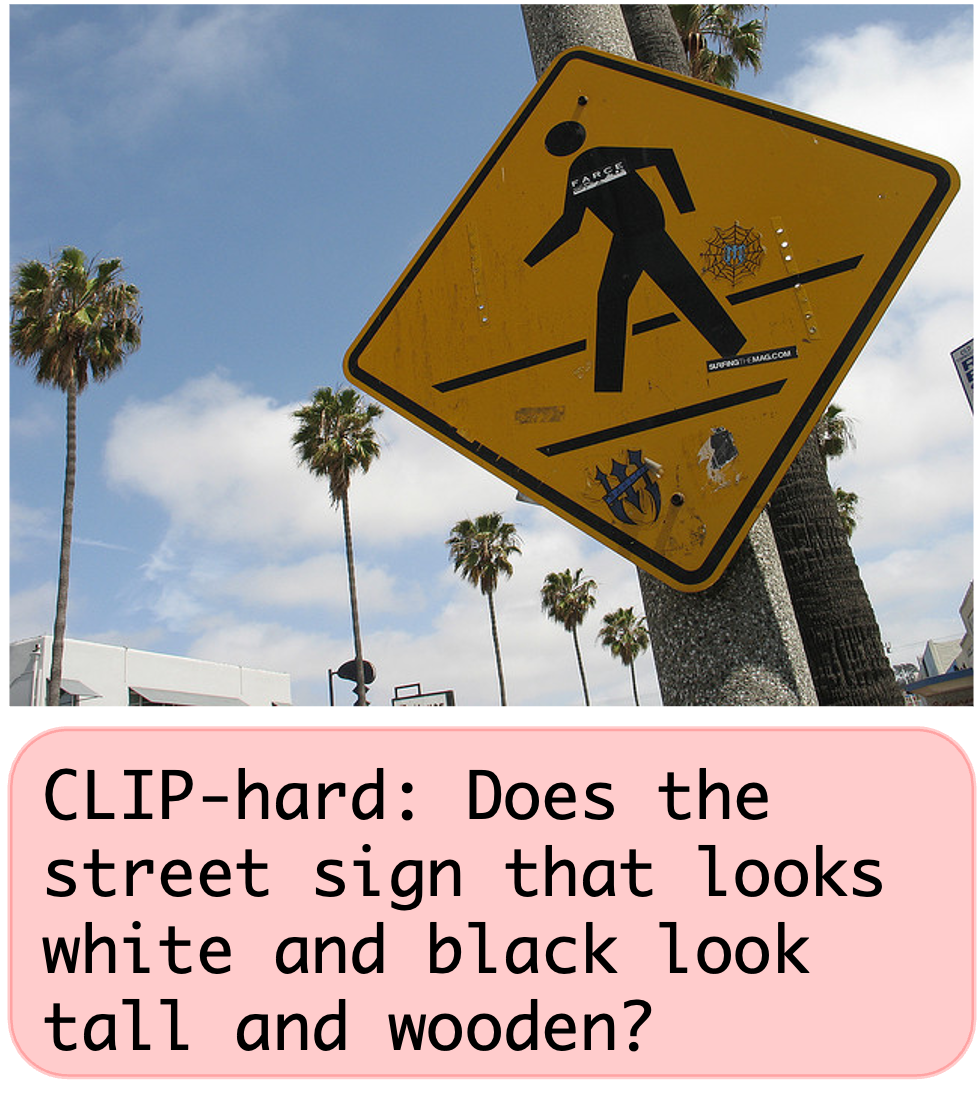}   &  
      \includegraphics[width=.2\linewidth,valign=t]{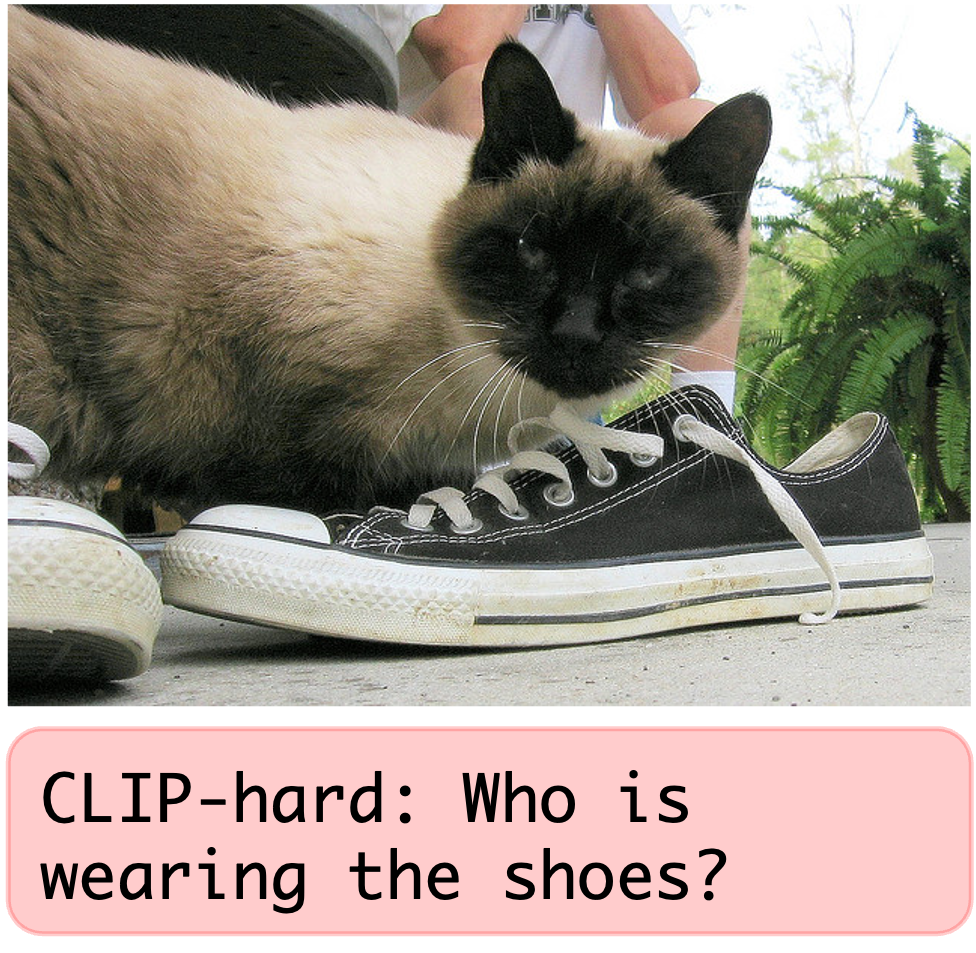}   &
      \includegraphics[width=.2\linewidth,valign=t]{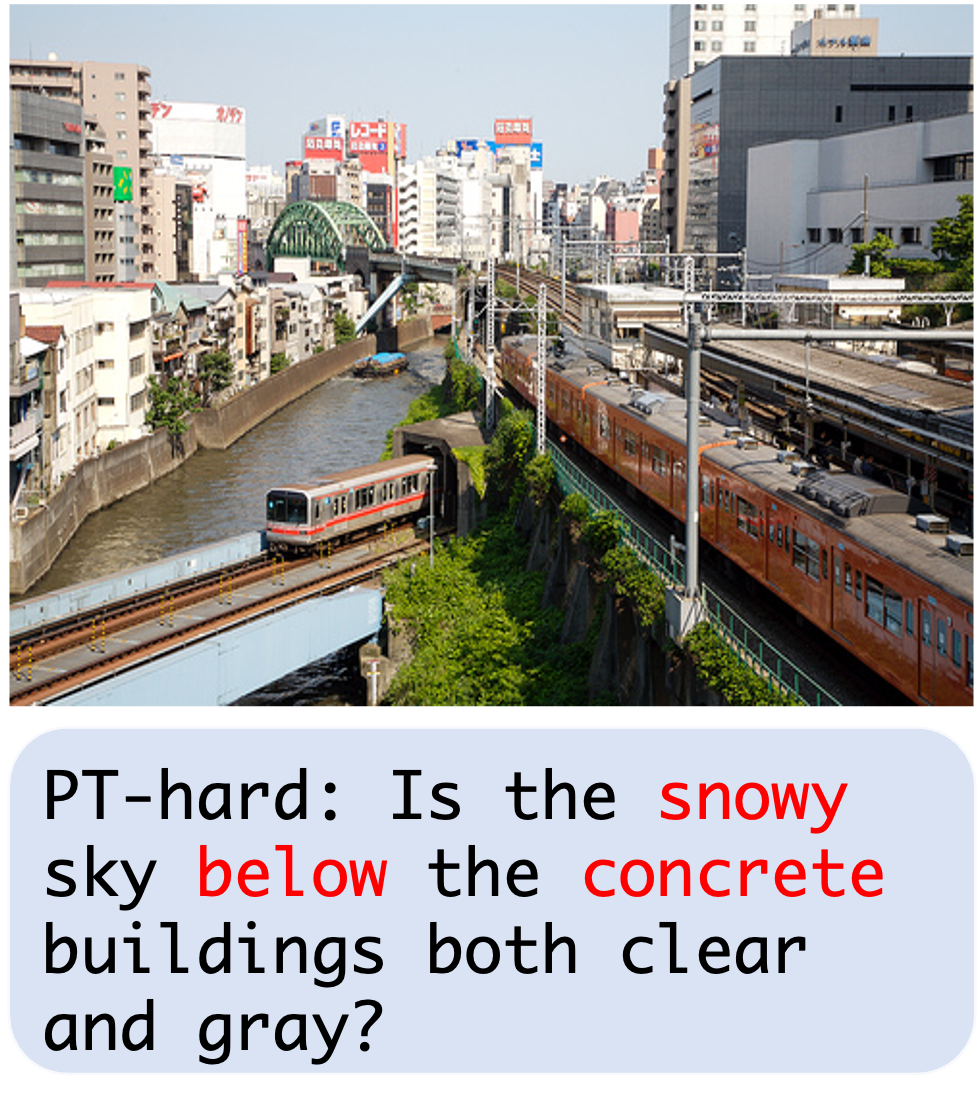} &
      \includegraphics[width=.2\linewidth,valign=t]{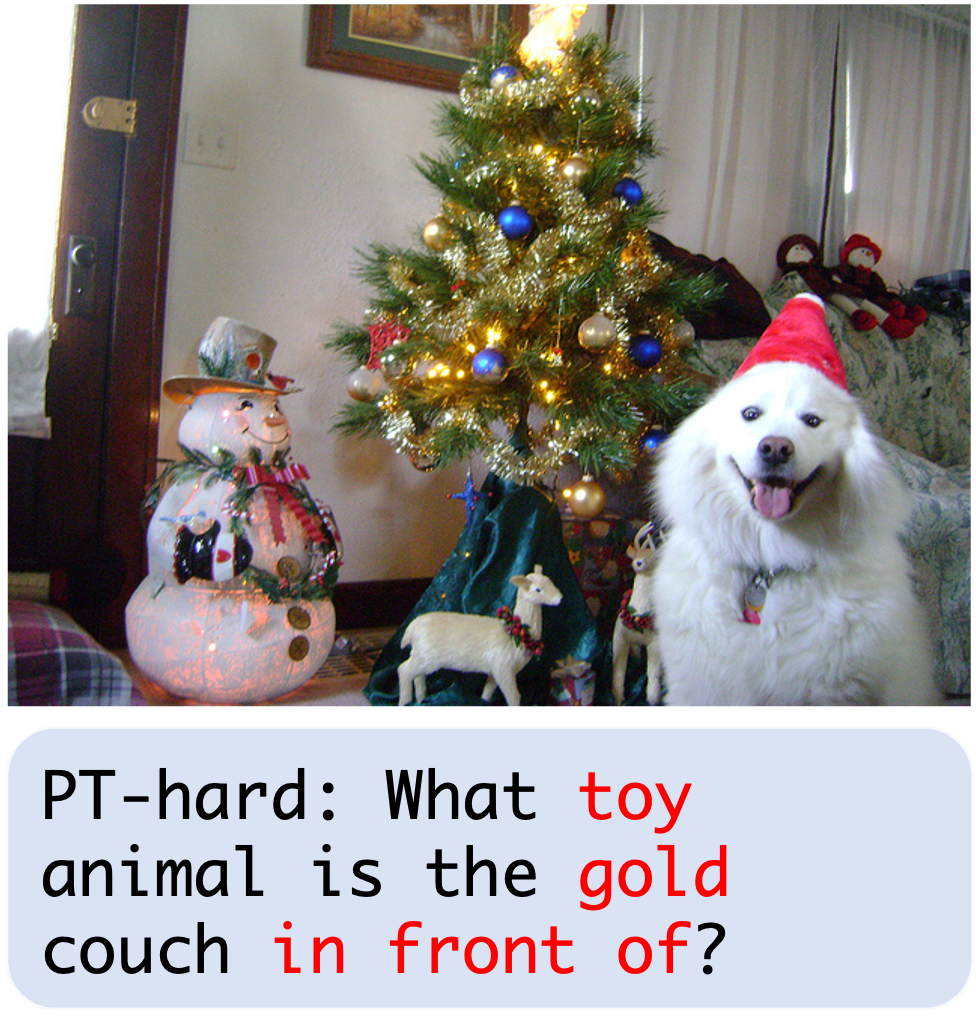}   
      \vspace{-5pt}
      \\
      \scriptsize{(a)} & \scriptsize{(b)} &  \scriptsize{(c)} & \scriptsize{(d)}\\
    \includegraphics[width=.2\linewidth,valign=t]{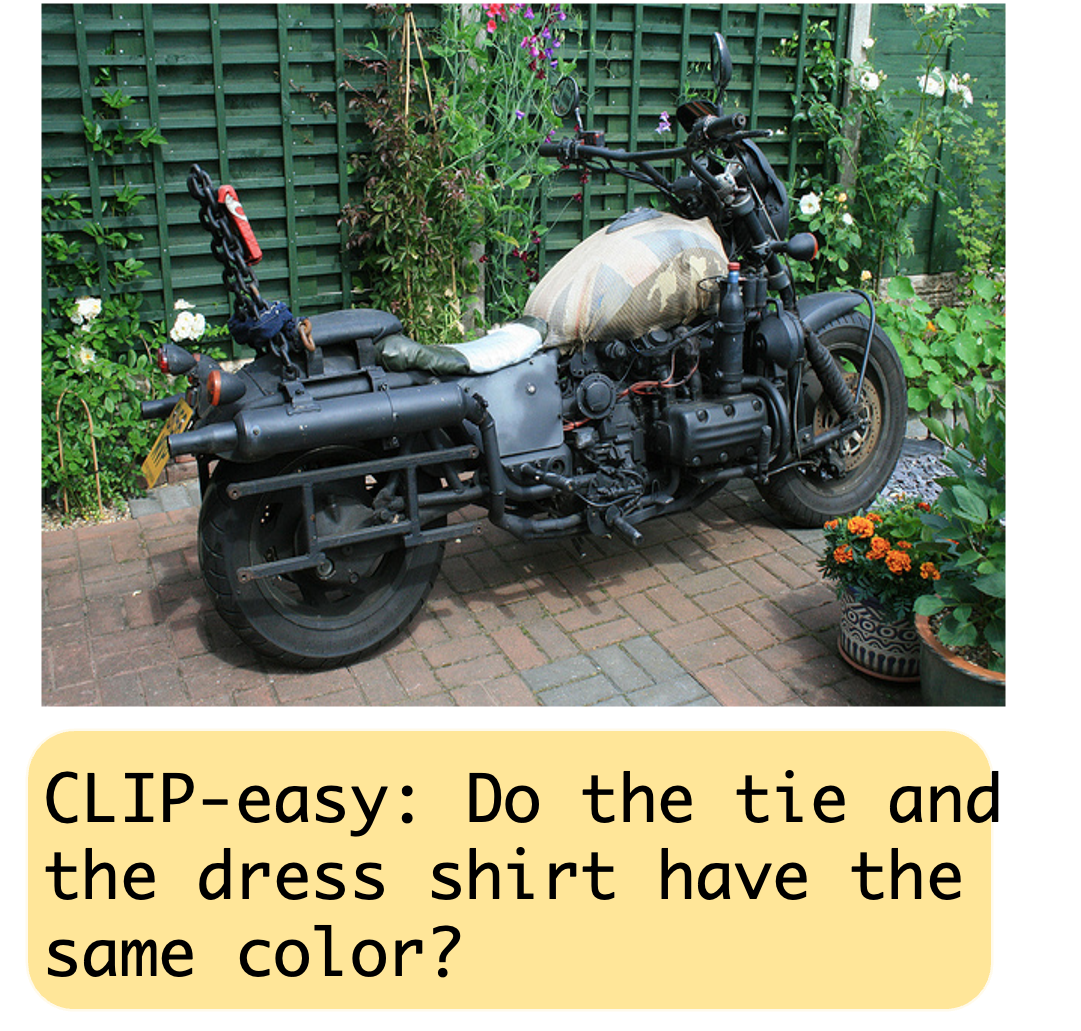}   &  
      \includegraphics[width=.2\linewidth,valign=t]{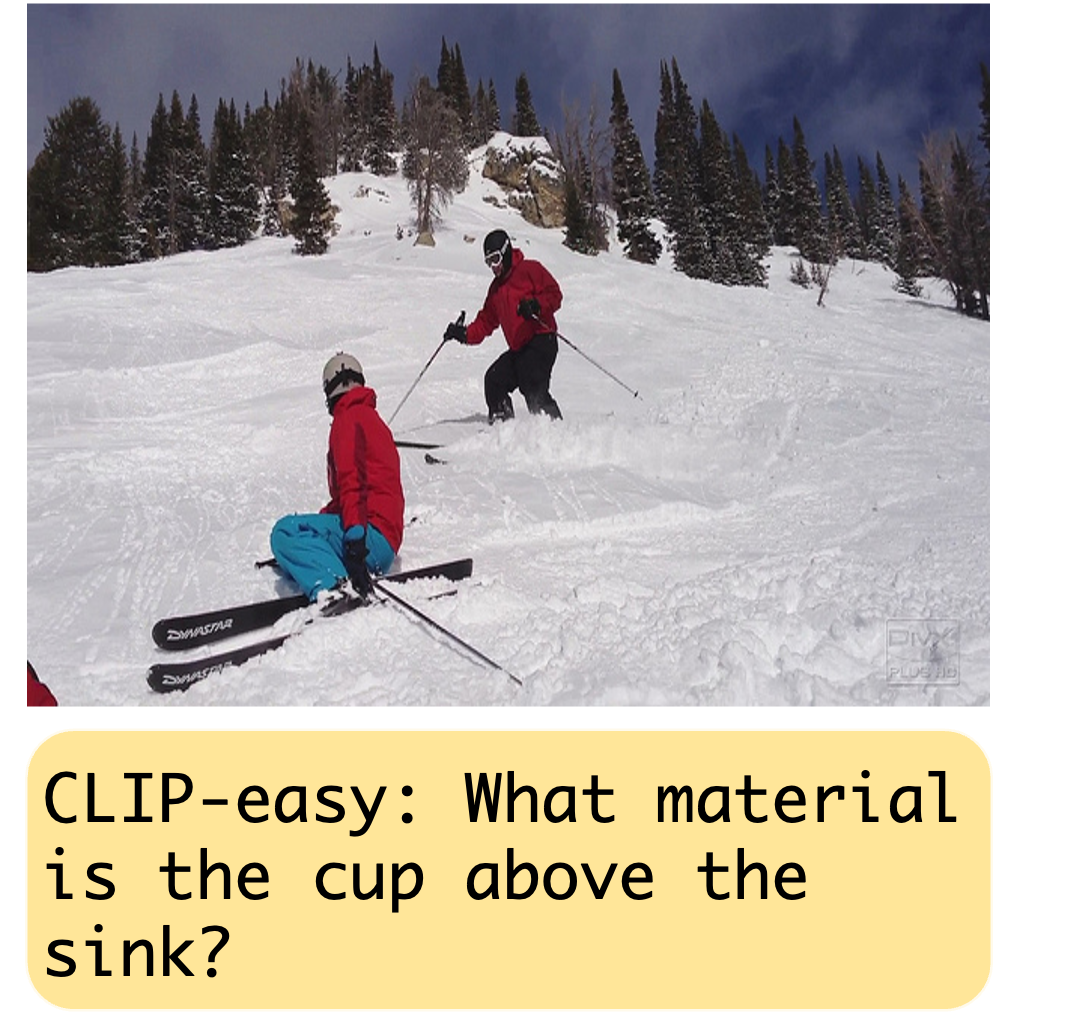}   &  
      \includegraphics[width=.2\linewidth,valign=t]{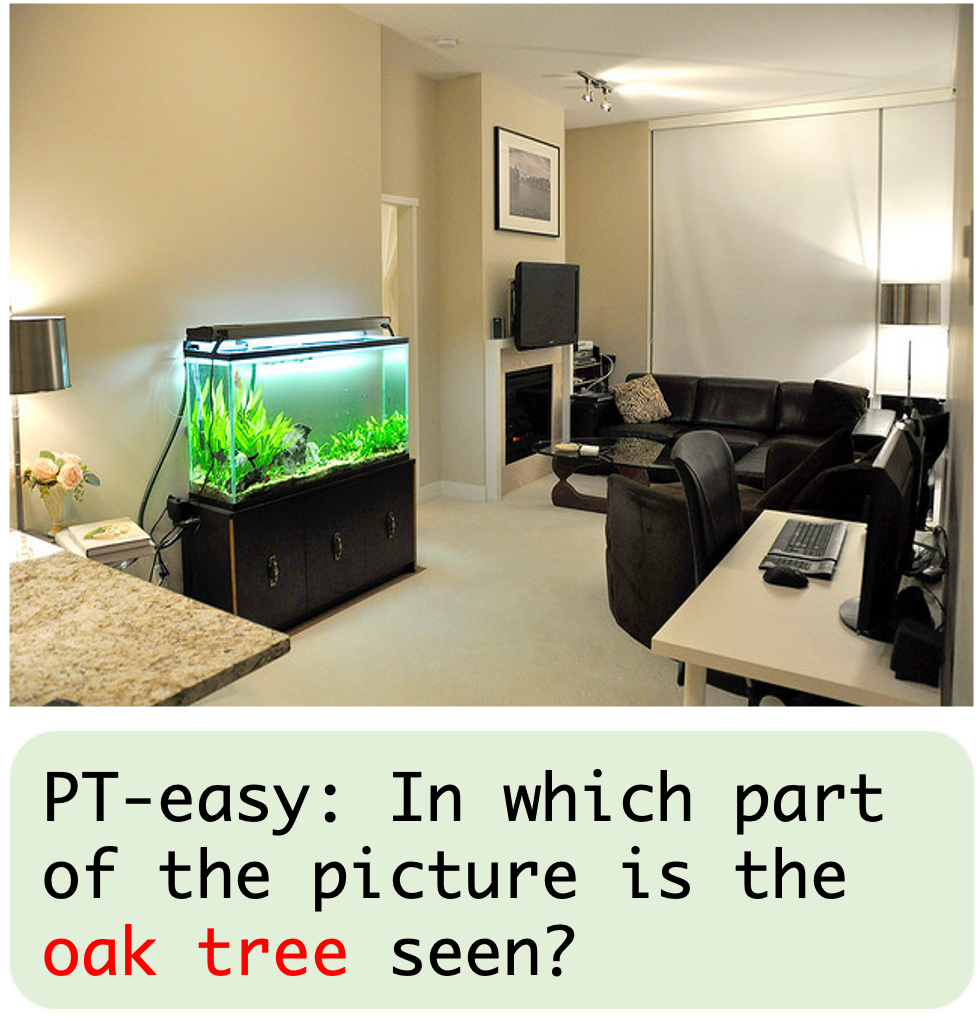}   &  
      \includegraphics[width=.2\linewidth,valign=t]{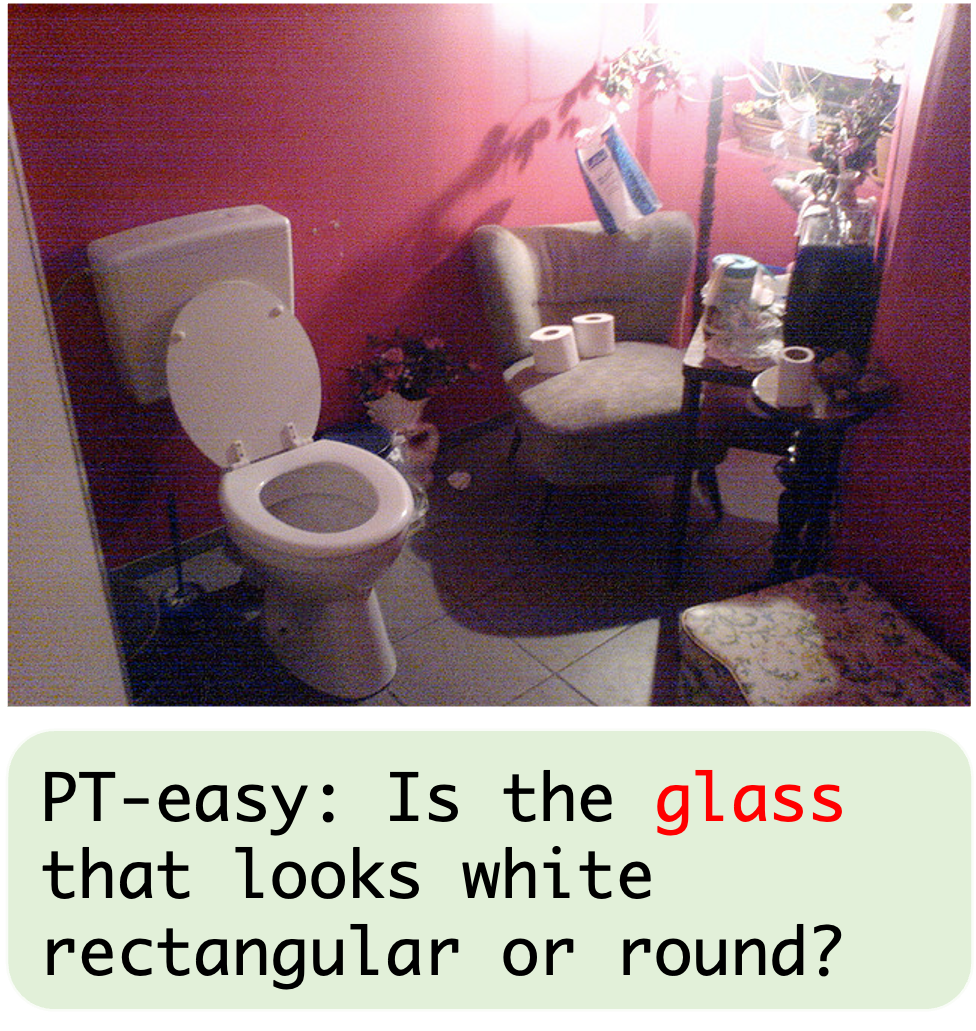}   
      \vspace{-5pt}
      \\ 
      \scriptsize{(e)} & \scriptsize{(f)} &  \scriptsize{(g)} & \scriptsize{(h)}\\
    \end{tabular}
    \caption{Examples of CLIP based  (a,b,e,f) and  Perturbation (PT) based UQs (c,d,g,h) in RGQA. For the PT-based UQs, the red words are modified from the original question. See appendix for more examples.}
    \label{fig:dataset_overview}
   
\end{figure*}

\section{Related Work}\vspace{-3pt}
 In this section, we review related works. See appendix for a broader discussion of the literature.

\noindent{\bf Realistic VQA (RVQA):}
The study of RVQA is still in its infancy.  A central question is how to assemble datasets of UQs, i.e. unrelated pairs of images and questions. Most methods start from a VQA dataset.
VTFQ~\cite{ray2016vtfq} collected a RVQA dataset by randomly pairing images and questions.
QRPE~\cite{mahendru2017qrpe} uses question-derived object/attribute premises. The associated image is then replaced by its Euclidean nearest neighbor in a set of images without the extracted premises.  These approaches are limited by the inability of random pairing or Euclidean similarity to guarantee a fine-grained semantic mismatch between image and UQ.

VizWiz~\cite{gurari2018vizwiz} is a VQA dataset
from the visually impaired setting, with UQs asked by people. 
However, its images are of poor quality and contain one or a few objects, which prevents complex interaction between objects, scenes, and language.
TDIUC~\cite{kafle2017tdiuc} and C2VQA~\cite{toor2018c2vqa} are created by checking if objects mentioned in questions also appear in images. 
While UQ cardinality can be easily scaled up~\cite{kafle2017tdiuc}  by randomly paring images and questions without common objects, this assumes that the only reason for a UQ is object mismatch.
In comparison, the proposed RGQA dataset considers both coarse- and fine-grained mismatches, based on stronger measures of image-question similarity. 
No constraints of image content are also imposed on UQ generation, producing a more challenging and diverse dataset.

All previous works address supervised RVQA, using annotated UQs, which is expensive and limits dataset sizes. For instance, \cite{ray2016vtfq} generates a caption per image with NeuralTalk2~\cite{karpathy2015neuraltalk} and measures question-caption similarity with a binary LSTM classifier. \cite{mahendru2017qrpe} further extracts the question premise and uses the concatenated question-premise-caption triplet as classifier input. \cite{li2020neural} uses this architecture to reject UQs in VQA. \cite{lee2021regularizing} uses the maximum attention score between objects and text tokens for rejection and regularizes attentions by training on UQs. In this work, we explore an unsupervised training strategy that is model-agnostic and does not rely on annotated UQs.

\noindent{\bf Out of Distribution Detection (OOD)}
RVQA is closely related to OOD in classification~\cite{hendrycks17baseline,Balaji17,liang2018enhancing,hendrycks2019oe,9201929,hendrycks2017a,lee2018maha} which aims to detect samples on which a classifier has not been trained. This has been addressed by temperature scaling of classifier logits~\cite{liang2018enhancing}, using Mahalanobis distance~\cite{lee2018maha} or energy scores~\cite{liu2020energy} to measure the distance to the training distribution, ensembling predictions from multiple models~\cite{Balaji17,Vyas2018OutofDistributionDU}, or  regularizing in-distribution (ID) features~\cite{9578373}.  It is also possible to use a background dataset, with different distribution from the training dataset, during training~\cite{Dhamija18,hendrycks2019oe,9201929,li2020background}. While background datasets can significantly improve OOD, prior works in RVQA~\cite{li2020neural,lee2021regularizing} show a performance degradation for AQs. We devise sampling strategies that address this problem.

The classification and OOD performance are usually reported by combining Area Under ROC curve (AUROC) and  accuracy on ID samples~\cite{oza2019c2ae,bao2021evidential,zhou2021placeholders,yue2021counterfactual}. However, separate metrics increase the difficulty to compare models. We introduce a unified metric for the RVQA problem.

\vspace{-10pt}
\section{RGQA Dataset}\vspace{-3pt}
In this section, we introduce the RGQA dataset for evaluating RVQA systems. It is a human-annotated dataset with  $\sim29K$ UQs and built upon the \textit{testdev} set of GQA~\cite{hudson2018gqa}. We purposely choose GQA because of its size, complex question structure, and high quality of images and annotations. 

\subsection{Dataset Curation} \vspace{-3pt}



RGQA has a balanced coverage of AQs and UQs. AQs are image-question pairs with answers from the GQA \textit{testdev} set. For UQs, we first generate a {\it candidate set\/} using two different approaches, \textit{CLIP-based} and \textit{Perturbation-based}, to mitigate potential UQ generation biases. Human annotators then decide which candidates are true UQs.

\begin{table*}
\caption{Comparison to previous datasets. The proposed RGQA dataset has longer and more fine-grain UQs and requires a multi-task classifier to solve the RVQA problem. RGQA is only for evaluation purposes.
}
    \centering
    \scalebox{0.75}{\begin{tabular}{c|c|c|c|c|c|c|c|c}
    \hline
        Dataset & Supervised UQ & Type & UQ Annotation & Image Source & Question Source & UQ(\%) & \# Test Pair & Avg. Length\\
        \hline
        VTFQ~\cite{ray2016vtfq} & \cmark & UQ det. & human & MSCOCO & VQAv1 & 89.24 & 31464 & 7.53\\
        QRPE~\cite{mahendru2017qrpe} & \cmark & UQ det. & generated & MSCOCO & VQAv1 & 50.87 & 35476 & 7.76\\
        C2VQA~\cite{toor2017c2vqa} & \cmark & UQ det. & generated & Visual Genome & Visual Genome & 50.00 & 29106 & 7.10\\
        TDIUC~\cite{kafle2017tdiuc} & \cmark & VQA+UQ det. & generated & MSCOCO+Visual Genome & VQAv1+Visual Genome & 22.17 & 538868 & 7.92\\
        VizWiz~\cite{gurari2018vizwiz} & \cmark & VQA+UQ det. & human & VizWiz & VizWiz & 27.84 & 8000 & 8.10\\
        \hline
        RGQA & \xmark & VQA+UQ det. & human & GQA \textit{testdev} & GQA \textit{testdev} & 52.22 & 55637 & 10.33\\
        \hline
    \end{tabular}
    }
    \label{tab:compare}
\end{table*}

\noindent\textbf{CLIP-based Candidate UQs:} 
Leveraging recent advances in image-text pre-training, we use
 CLIP
 ~\cite{radford2021clip} to measure similarity between images and questions. Given an image $I$, we consider the set of questions ${\cal Q}(I)$ in  the \textit{testdev} dataset, excluding 1) existence questions (e.g. ``Are there any ...?''), which can never be UQs, and 2) the questions originally paired with $I$. We then feed all pairs $(I, Q), Q \in {\cal Q}(I)$ to  the CLIP model and 
rank the questions by similarity score.
To cover the spectrum from simple to hard UQs,
$85$ questions sampled from the top $2,500$ are used as candidate UQs for CLIP-Hard,
while the last $50$ questions are used as candidate UQs for CLIP-Easy. Fig.~\ref{fig:dataset_overview} shows images from each set. The pairs of CLIP-Hard (Fig.~\ref{fig:dataset_overview} (a,b)) have more subtle mismatches than those of CLIP-Easy (Fig.~\ref{fig:dataset_overview} (e,f)).


\noindent\textbf{Perturbation-based Candidate UQs:} 
Given an AQ in GQA \textit{testdev}, a candidate UQ counterpart is generated by perturbing its objects and adjectives. This is implemented by first collecting a set of candidate objects and their attributes from the scene graphs of GQA \textit{train} and \textit{valid} set. 
For each AQ, objects and adjectives are extracted by POS 
tagging. Similar to the CLIP-based approach, both easy and hard UQs are generated by the perturbation-based approach, resulting in the subsets PT-Easy and PT-Hard. For PT-Easy, each object in the AQ is replaced by a random but different object sampled from the candidate object set. For PT-Hard, the objects in AQ are kept but their attributes are replaced by different candidate attributes of the same object. Finally, the spatial relation terms in PT-Hard are replaced by antonyms, such as ``left/right'' and  ``top/bottom''. Conflicting questions, like ``What color are the black shoes?'' are then eliminated.  Fig.~\ref{fig:dataset_overview}(g,h) and Fig.~\ref{fig:dataset_overview}(c,d) show examples from PT-Easy and PT-Hard, with the perturbed text in red.


\noindent\textbf{Human Annotation:}
Human annotators analyze all image-question candidates and decide which are true UQs. Following ~\cite{winoground, Hasan2018OverviewOI,qed,Chen_2022_CVPR}, we use 8 expert annotators with experience in visual language research. The annotator is shown an image and two questions (see interface in appendix), and asked to choose from ``\textit{valid}'' (corresponding to AQs) and ``\textit{invalid}'' (UQs) options for each question. We instruct the annotator to choose ``\textit{valid}'' if the decision is ambiguous, due to unclear images, confusing wording, or any other reason. These annotations are discarded.


This process produced $11,264$ UQs for CLIP-Hard, $5,689$ for CLIP-Easy, $6,130$ for PT-Easy and $5,963$ for PT-Hard. The next step aimed to sample a similar number of AQs, to balance the dataset. For CLIP-Hard and CLIP-Easy, we randomly sample AQs to pair with UQs. For each UQ, we consider the associated image and retrieve the AQs originally paired with this image in GQA. We then randomly sample one of these AQs. This produced  $11,158$ questions for CLIP-Easy and $20,325$ for CLIP-Hard. For PT-Easy and PT-Hard, we pair with the original AQs for each perturbed UQ which results in $12,241$ questions in total for PT-Easy and $11,913$ for PT-Hard. See appendix for more details.

\subsection{Dataset Analysis} \vspace{-3pt}











\textbf{UQ Categories:}
RGQA covers a wide spectrum of UQs, including questions without valid answers (\textit{e.g.} Fig.~\ref{fig:dataset_overview} (b)), with false premise at object (\textit{e.g.} Fig.~\ref{fig:dataset_overview} (e)) or attribute level (\textit{e.g.} Fig.~\ref{fig:dataset_overview} (d)),
and underspecified questions (\textit{e.g.} ``Do the snowpants look black and long?'' for Fig.~\ref{fig:dataset_overview} (f)).
Many UQs also have subtle mismatches with the image, which can only be spotted via high-level understanding of image semantics. For instance, in  Fig.~\ref{fig:dataset_overview}(b), both the predicate ``wearing'' and the object ``shoes'' exist in the image, so the model needs to understand the semantics of ``wearing'' and search for their subject. Hence, beyond evaluating robustness, RGQA also measures how strongly VQA models learn semantics.

\textbf{Dataset Comparison:}
Table~\ref{tab:compare} compares RGQA to previous VQA datasets with UQs~\cite{ray2016vtfq,mahendru2017qrpe,gurari2018vizwiz,toor2017c2vqa,kafle2017tdiuc}. Several of these only address UQ detection. RGQA combines this with VQA, which better matches real-world applications. It also contains higher-quality human annotations, a better balance between AQs and UQs, and longer and more complex questions (last column) than previous datasets. Overall, it poses a greater challenge to model reasoning skills.

\definecolor{babyblue}{rgb}{0.52, 0.76, 0.91}
\definecolor{brightube}{rgb}{0.73, 0.56, 0.81}
\definecolor{celadon}{rgb}{0.51, 0.88, 0.67}
\begin{figure*}
\begin{minipage}{0.71\linewidth}
\begin{tabular}{cccccccccccc}
    \multicolumn{4}{c}{\includegraphics[width=0.31\linewidth]{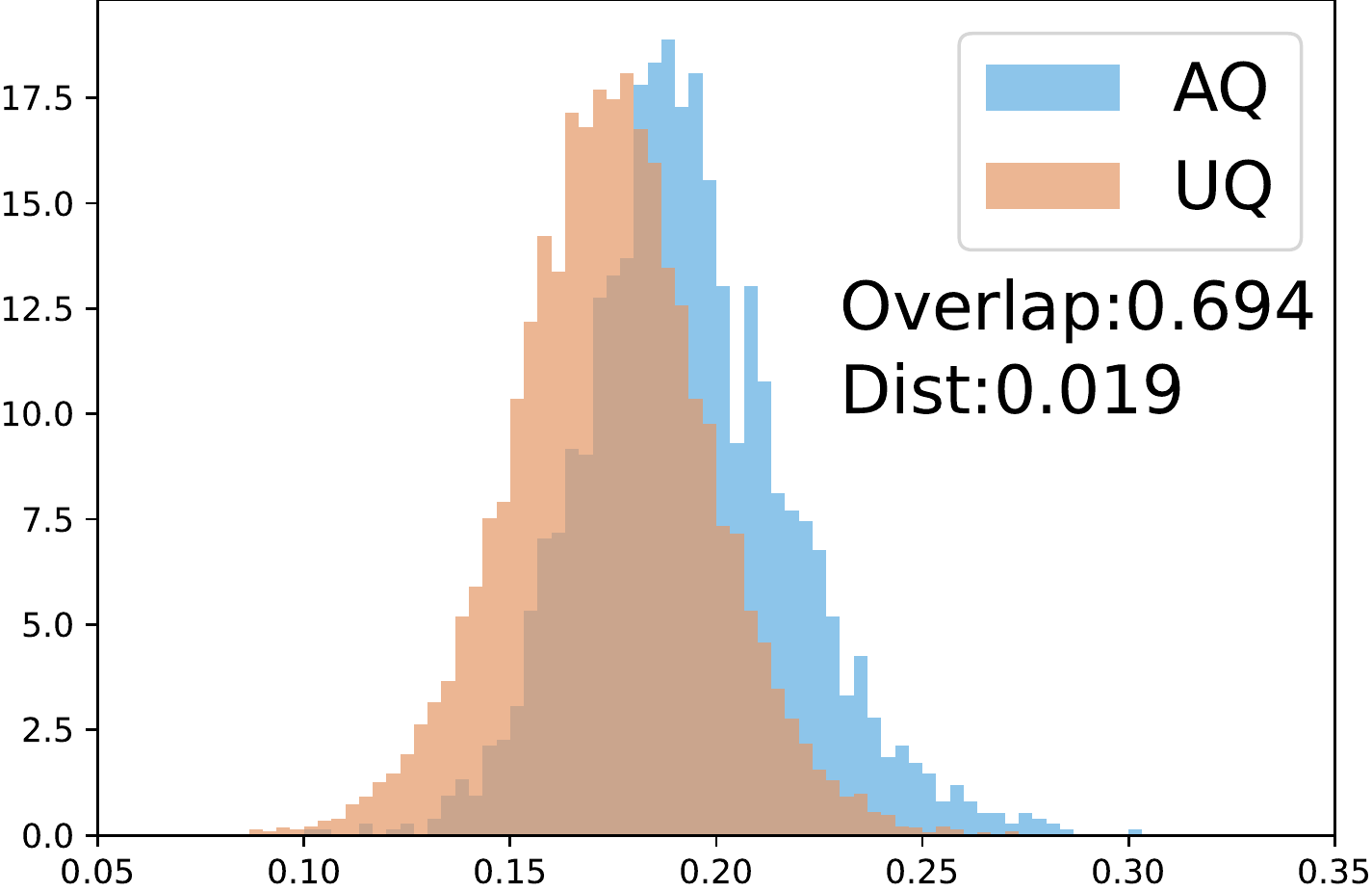}} & \multicolumn{4}{c}{\includegraphics[width=0.31\linewidth]{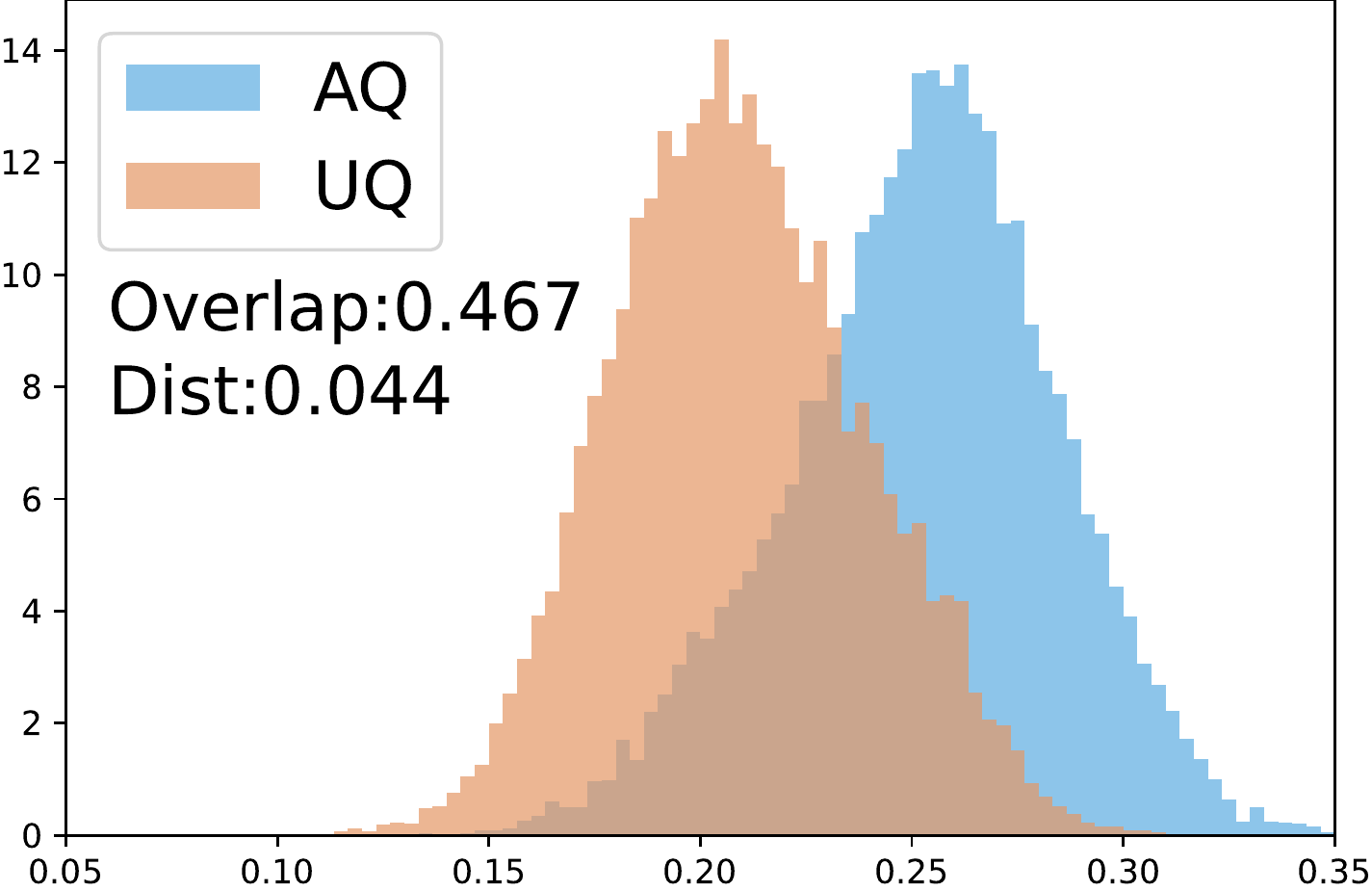}} & 
    \multicolumn{4}{c}{\includegraphics[width=0.31\linewidth]{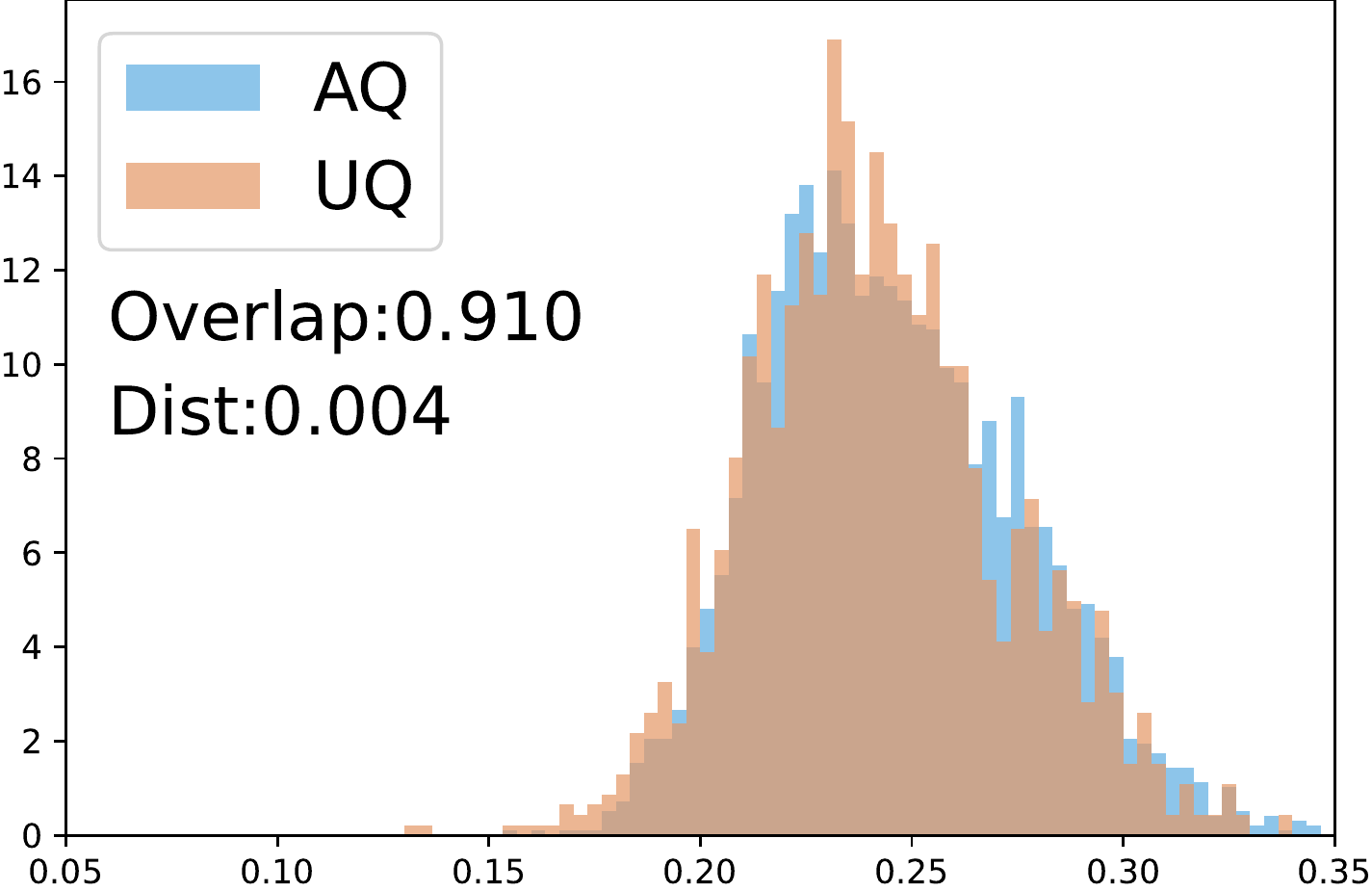}}
    \vspace{-5pt}
    \\
    \multicolumn{4}{c}{\scriptsize{VTFQ~\cite{ray2016vtfq}}}
    & \multicolumn{4}{c}{\scriptsize{QRPE~\cite{mahendru2017qrpe}}} & 
    \multicolumn{4}{c}{\scriptsize{VizWiz~\cite{gurari2018vizwiz}}} \\
    \multicolumn{3}{c}
    {\includegraphics[width=0.22\linewidth]{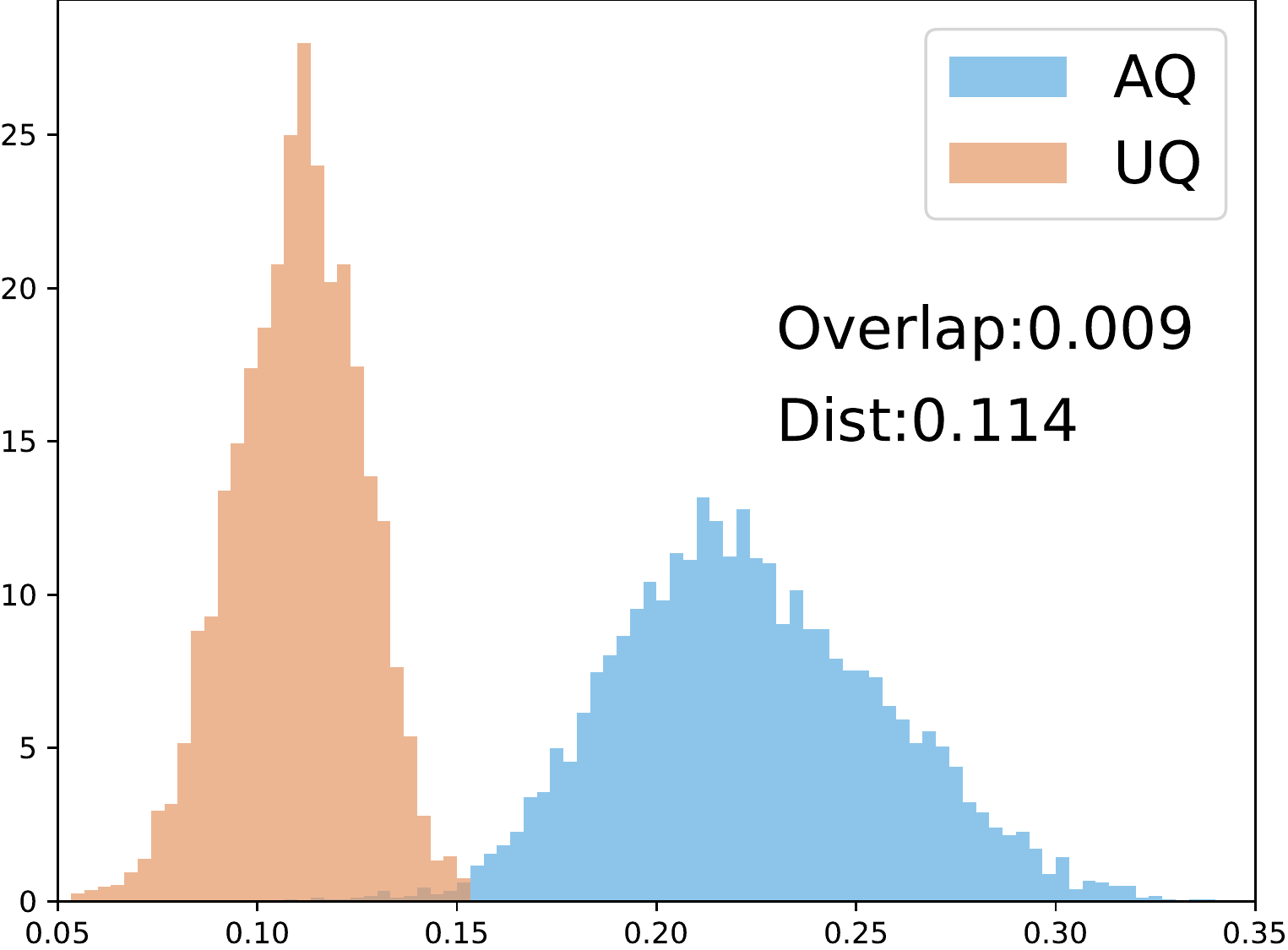}}
    & \multicolumn{3}{c}
    {\includegraphics[width=0.22\linewidth]{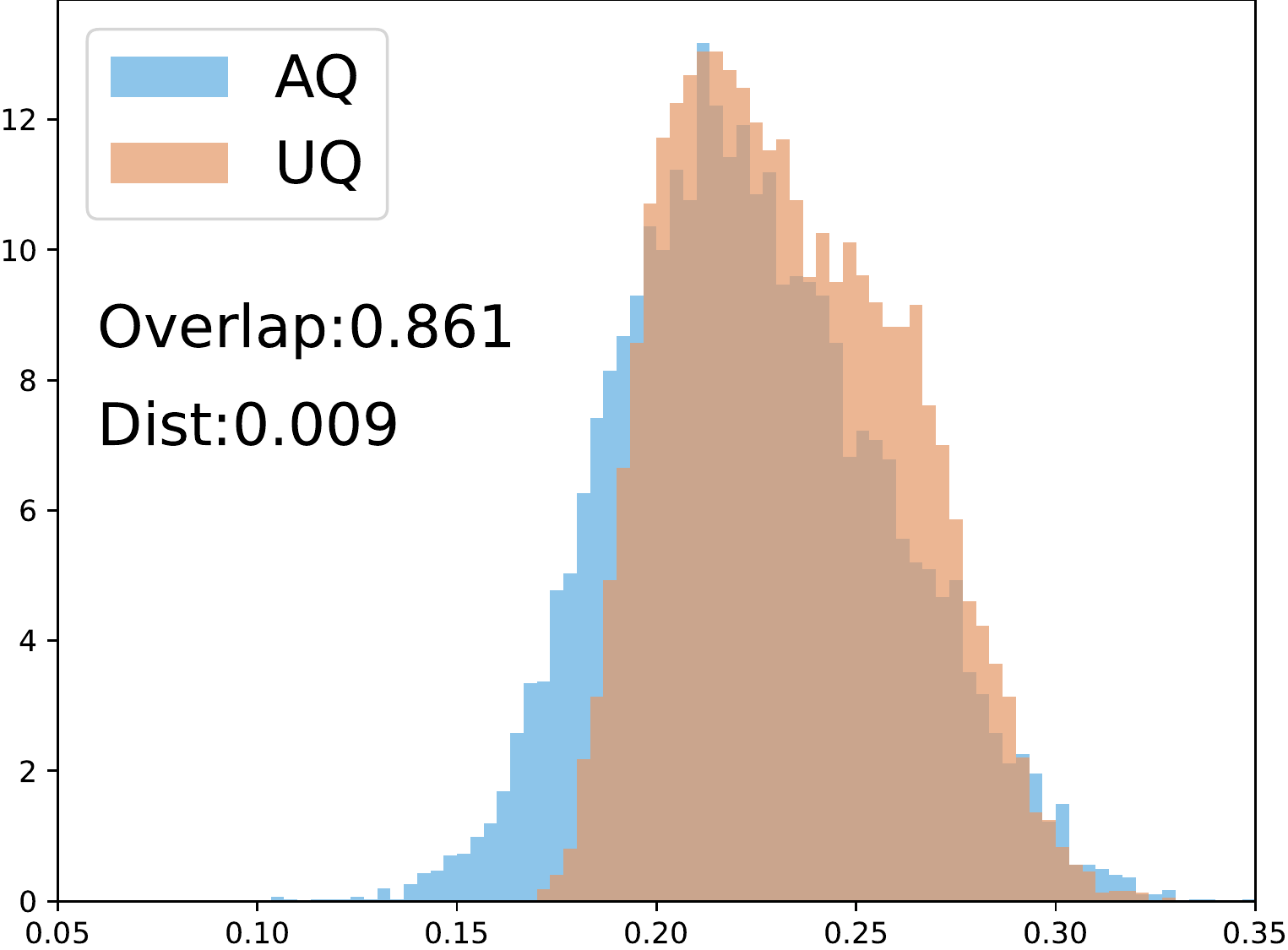}}
    & 
    \multicolumn{3}{c}
    {\includegraphics[width=0.22\linewidth]{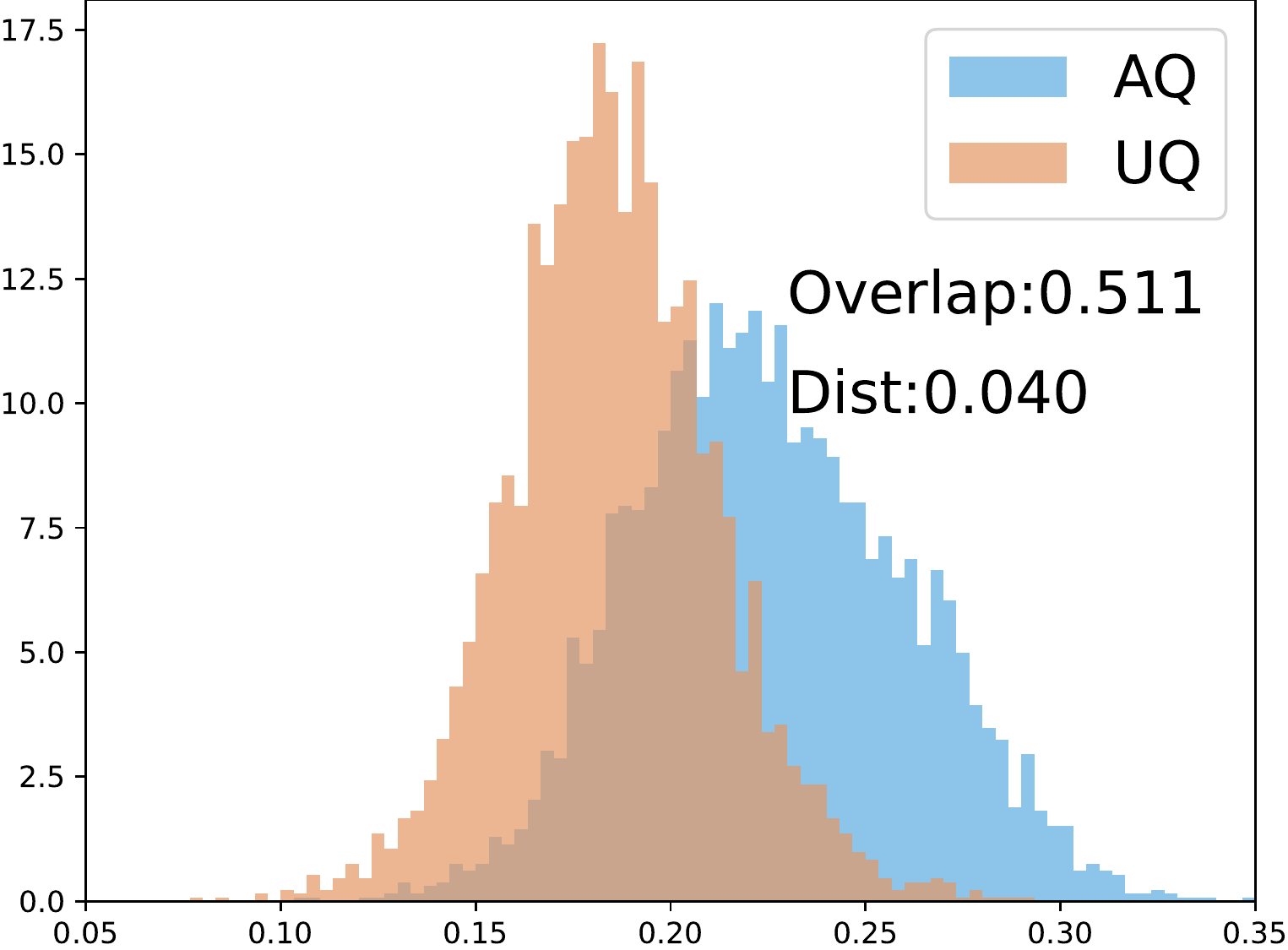}}
    & 
    \multicolumn{3}{c}
    {\includegraphics[width=0.22\linewidth]{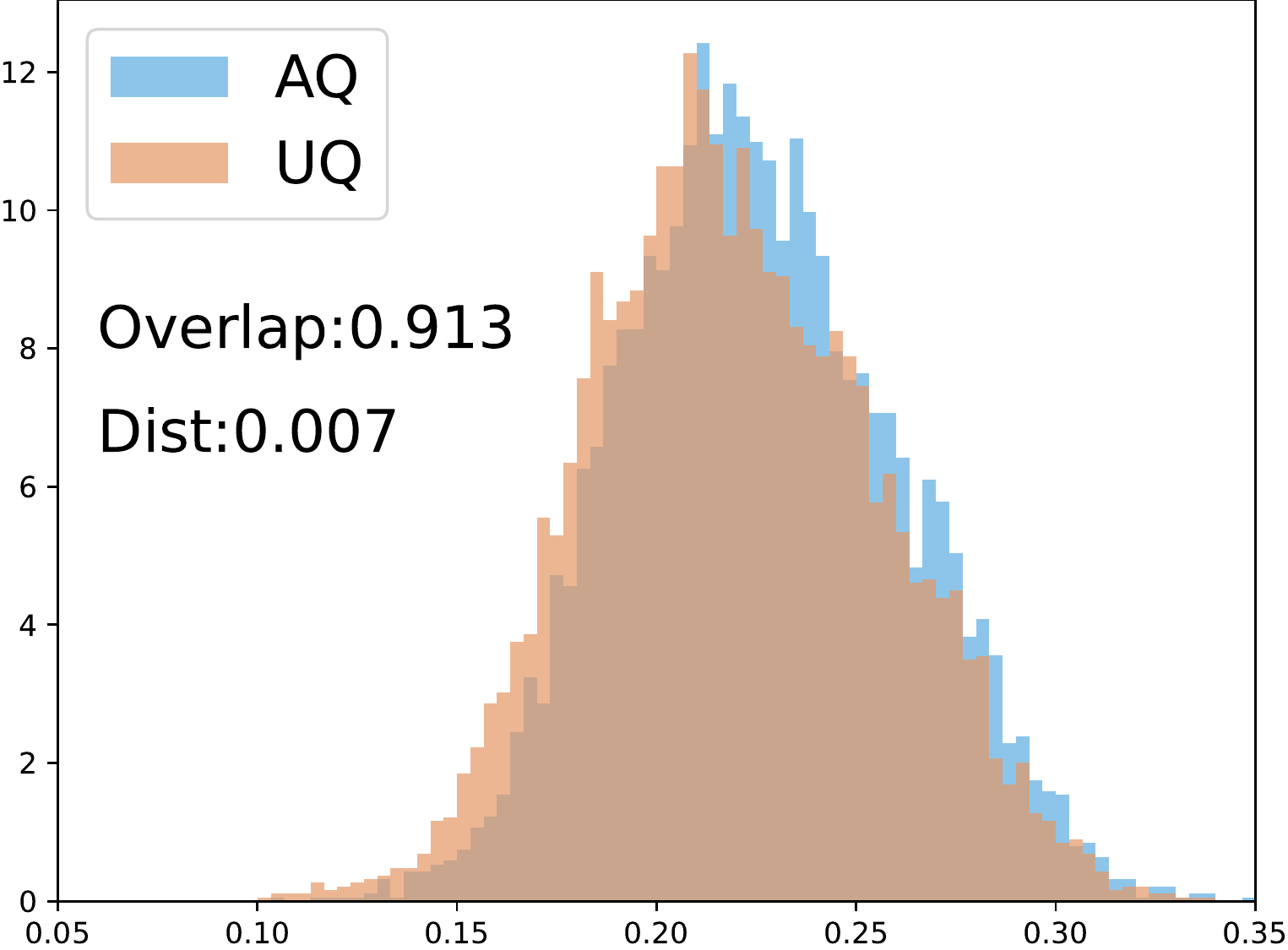}}
    \vspace{-5pt}
    \\
    \multicolumn{3}{c}{\scriptsize{CLIP-Easy (Ours)}}
    & \multicolumn{3}{c}{\scriptsize{CLIP-Hard (Ours)}} & 
    \multicolumn{3}{c}{\scriptsize{PT-Easy (Ours)}}& 
    \multicolumn{3}{c}{\scriptsize{PT-Hard (Ours)}} \\
\end{tabular}
\captionof{figure}{CLIP image-question similarity distribution of both AQs and UQs. The overlap area between 2 normalized histograms (sum of overall area=1) and the distance between the means are computed.
}
    \label{fig:clip score}
\end{minipage}%
\hspace{5pt}
\begin{minipage}{0.28\linewidth}
    \begin{tabular}{c}
        \includegraphics[width=\linewidth]{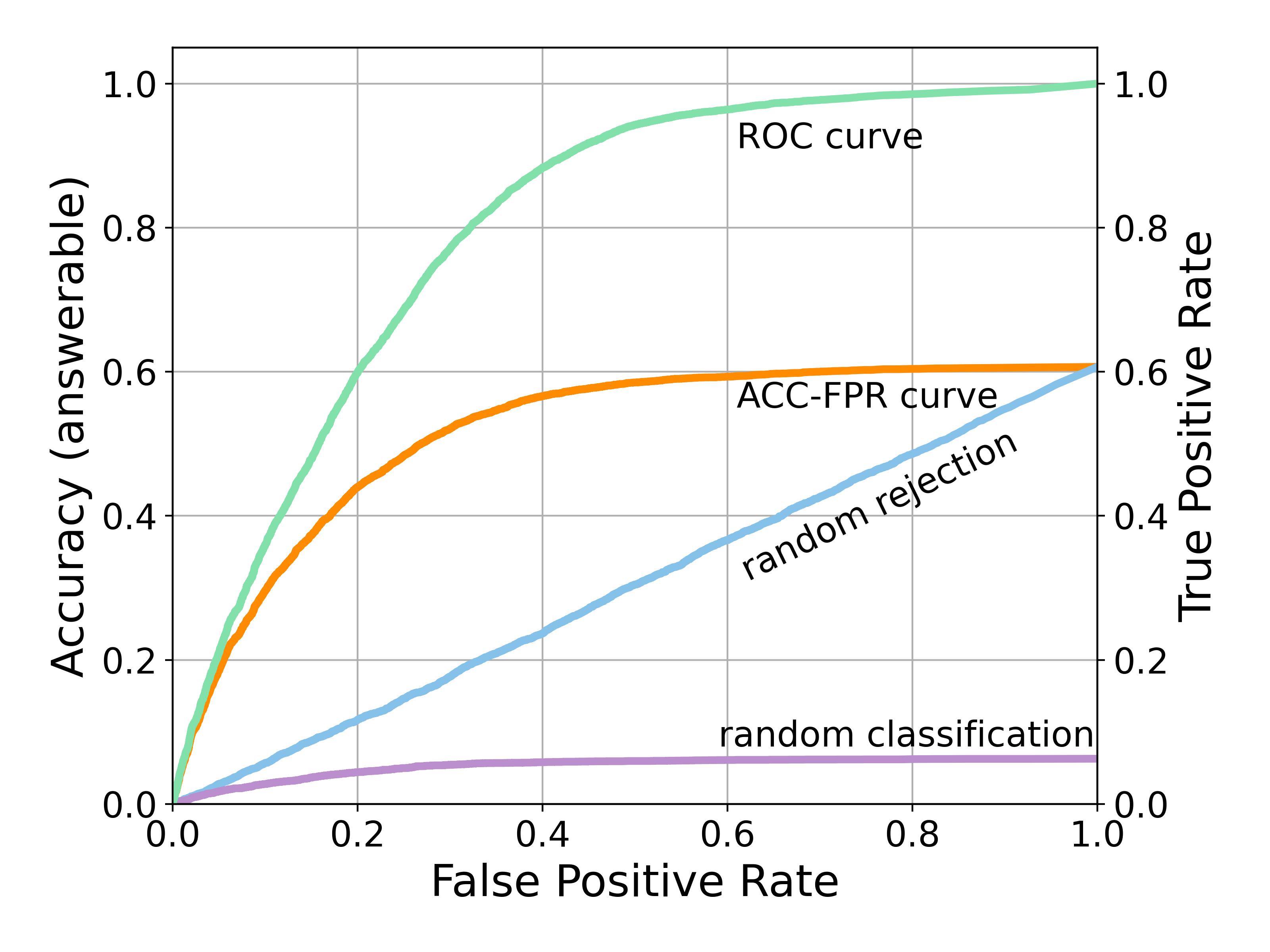}
    \end{tabular}
    \captionof{figure}{
    Comparison between ROC curve (\textcolor{celadon}{green}; right axis) and ACC-FPR curve (\textcolor{orange}{orange}; left axis). See text for details.
    }
    \label{fig:auaf_example}
\end{minipage}
\end{figure*}

\textbf{AQs vs UQs:}
To gain insight on the differences between AQs and UQs, we performed an analysis from two aspects. The first is to plot the distributions of image-question CLIP similarity scores, as shown in Fig.~\ref{fig:clip score}. Clearly, for VTFQ~\cite{ray2016vtfq} and QRPE~\cite{mahendru2017qrpe} the scores are smaller, indicating simpler questions, and the AQ/UQ distributions have less overlap, showing that they can be easily separated. VizWiz~\cite{gurari2018vizwiz}, CLIP-Hard, and PT-Hard have larger scores and stronger overlap between the two distributions, indicating that their UQs have finer-grained mismatch between image and question.
However, while the CLIP score measures semantic similarity, it does not capture the answerability of UQs. The second strategy addresses this limitation, by plotting the distribution of questions by the first three words (See appendix). Other than a different order for the three most popular words (``Are'', ``Who'' and ``Which'') and a few changes on the proportions, there are no major differences between the AQ and UQ distributions. This shows that AQs/UQs cannot be easily separated by question structure.



\subsection{Evaluation Metrics} \vspace{-3pt}
\definecolor{babyblue}{rgb}{0.52, 0.76, 0.91}
\definecolor{brightube}{rgb}{0.73, 0.56, 0.81}
\definecolor{celadon}{rgb}{0.51, 0.88, 0.67}

Since UQ detection is an OOD problem, we leverage well-established OOD practices for evaluation.
However, because RVQA requires jointly solving UQ detection and VQA, the common OOD practice of reporting close-set accuracy and AUROC is not satisfying. We instead proposed to use the ACC-FPR curve, introduced as CCR-FPR curve in~\cite{dhamija2018agnostophobia}, which measures the joint performance.
Given a VQA classifier $f$ and a UQ detector $g$, ACC is the proportion of AQs with correct VQA prediction and accepted as AQ, i.e.
\begin{equation}
    \text{ACC}=\frac{|\{x_i|f(x_i)=a_i,g(x_i)=\text{AQ},(x_i,a_i)\in \mathcal{D}^{aq}\}|}{|\mathcal{D}^{aq}|},
\end{equation}
where $x_i=(v_i,q_i)$ denotes  image-question pair, $a_i$ is the corresponding VQA answer and  $\mathcal{D}^{aq}$ is the dataset of AQs.
FPR is the proportion of UQs falsely accepted as AQ, i.e.
\begin{equation}
    \text{FPR}=\frac{|\{x_i|g(x_i)=\text{AQ},x_i\in \mathcal{D}^{uq}\}|}{|\mathcal{D}^{uq}|},
\end{equation}
where $\mathcal{D}^{uq}$ is the dataset of UQs.
The ACC-FPR curve is drawn by connecting ACCs (y-axis) at different FPRs (x-axis) as in Fig.~\ref{fig:auaf_example}.
We define the maximum value of the curve on the y axis (best accuracy the model can achieve on AQs) as full accuracy (FACC).

A RVQA model with a strong VQA classifier $f$ and a UQ detector $g$ (\textcolor{orange}{orange line}) has higher FACC than a model with the same $g$ but random $f$ (\textcolor{brightube}{purple line}).
On the other hand, a model with the same $f$ but random $g$ (\textcolor{babyblue}{blue line}) has the same FACC but underperforms the RVQA model for all FPRs less than 1.
Note that the ROC curve (\textcolor{celadon}{green line}) is the special case of ACC-FPR curve with FACC$=1$.
As a single evaluation metric, we use \textit{Area Under} ACC-FPR curve (AUAF), for joint performance, FPR at $95\%$ FACC (FF95) for rejection, and FACC for classification.


\vspace{-3pt}
\section{Unsupervised RVQA Learning}
\vspace{-3pt}

In this section, we introduce unsupervised RVQA and three model-agnostic methods for unsupervised training.

\vspace{-2pt}
\subsection{Unsupervised RVQA}\vspace{-3pt}
\label{sec:problem}
Unsupervised RVQA learns a model, VQA classifier $f$ and UQ detector $g$, from a dataset of AQs $\mathcal{D}^{aq}_{tr}=\{(x_i,a_i)\}$, {\it without\/} annotated UQs. At testing, $g(x)$ decides whether a pair $x$ is accepted. If so, $f(x)$ then predicts an answer. 
\begin{figure*}
    \centering
    \includegraphics[width=0.9\linewidth,height=4cm]{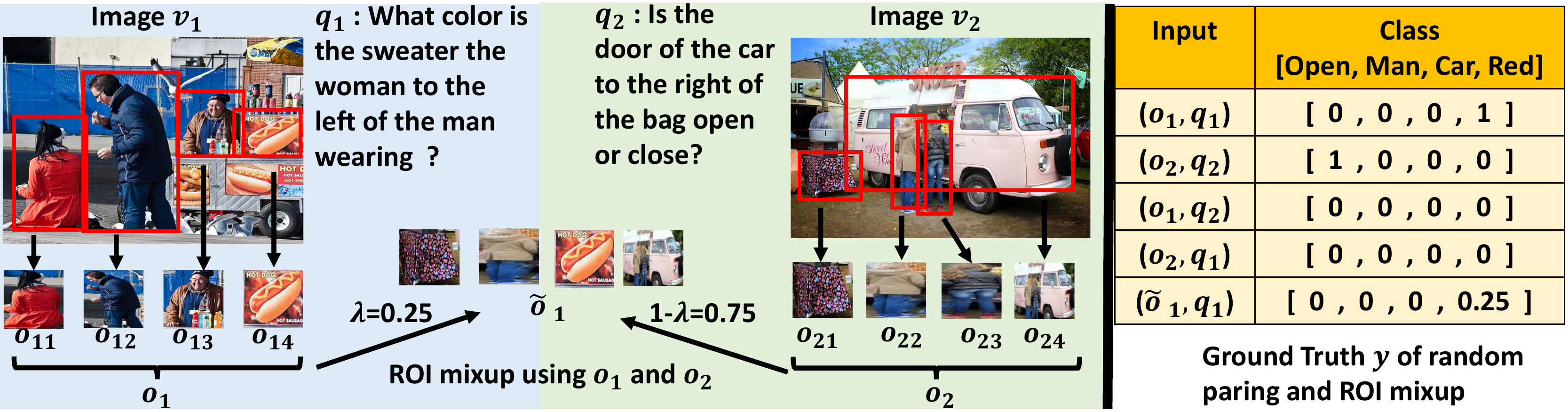}
    \caption{Illustration of the pseudo UQ and RoI Mixup. The right table shows the label for different visual question inputs. }
    \label{fig:method}
\end{figure*}

\vspace{-3pt}
\subsection{Training with Pseudo UQ} \vspace{-3pt}
\label{sec:background}
Inspired by recent OOD works using an auxiliary background dataset~\cite{dhamija2018agnostophobia,hendrycks2019oe,li2020background,ming2022poem} for training, we investigate training the RVQA model with a background dataset. 
For image classification, choosing a background dataset of reasonable scale and effective performance is non-trivial~\cite{li2020background}. 
However, this is much simpler for RVQA: a simple and natural choice is to  randomly pair images and question $\{(v_i,q_i)\}$ already available in the VQA dataset. Given an image $v_i$, we randomly sample a question $q_k$ belonging to a different image $v_k\neq v_i$ to form a \textit{\textbf{pseudo}} unanswerable image-question pair $(v_i,q_k)$. Fig.~\ref{fig:method} illustrates an example of this random paring, where image $\nu_1$ is paired with question $q_2$. Like this example, most randomly sampled pairs are unanswerable\footnote{We inspected 100 pairs on  GQA \textit{train} and found 77\% to be UQs.}. The \textit{\textbf{pseudo UQs}} are used to construct the unsupervised background dataset $\hat{\mathcal{D}}^{uq}_{tr}$.

With $\mathcal{D}^{aq}_{tr}$ and  $\hat{\mathcal{D}}^{uq}_{tr}$, the VQA classifier $f$ and binary UQ detector $g$ can be trained to minimize the risk
{\small 
\begin{eqnarray}
    {\cal R} &=& E_{(x_i,a_i)\in\mathcal{D}^{aq}_{tr}} \mathbb{I}(f(x_i)\neq a_i) \label{eq:objective} \\
    &+& E_{x_i\in \mathcal{D}^{aq}_{tr}} \mathbb{I}(g(x_i)\neq\text{AQ}) 
    + E_{x_i\in \hat{\mathcal{D}}^{uq}_{tr}} \mathbb{I}(g(x_i)\neq\text{UQ}), \nonumber
\end{eqnarray}
}
where the first term is the classification error and the last two are the detection error.
Different from most OOD methods, which use softmax outputs~\cite{hendrycks2019oe}, VQA models are usually trained as multi-label models. Let ${\cal Y} = \{1, \ldots, K\}$ be the set of possible answers. Then, the ground truth for $i^{th}$ example $x_i = (v_i,q_i)$ and $k^{th}$ answer is a binary variable, $y_{i,k} \in \{0,1\}$, with $y_{i,k} = 1$ if the answer holds for $x_i$ and $y_{i,k} = 0$ otherwise. The VQA model $f$ has $K$ binary outputs, where $f_k(x)$ is the predicted probability for $k^{th}$ answer, implemented with sigmoid functions and trained with the \textit{binary cross entropy} (BCE) loss
{\small 
\begin{equation}
    l_i=\sum_{k=1}^K y_{i,k}\log f_k(x_i) + (1 - y_{i,k})\log(1-f_k(x_i)).
    \label{eq:vqa uqd1}
\end{equation}
}
In Sec.~\ref{sec:pilot}, several  configurations of models $f$ and $g$ are ablated.
Best results were obtained with an {\it integrated\/} model, where both $f$ and $g$ 
share the network according to
{\small 
\begin{eqnarray}
    g(x) = \mathbb{I}(\max_k f_k(x) > \theta) \rightarrow  y^* = \arg \max_k f_k(x), 
    \label{eq:model}
\end{eqnarray}
}
where $\rightarrow$ means that the second equation is only implemented if $g(x)=1$. The rejection step first checks that there is at least one $f_k$ above threshold $\theta$. If so, VQA is performed. Otherwise, the example $x$ is identified as a UQ and rejected. This model minimizes (\ref{eq:objective}) by simply assigning labels $y_{i,k}=0, \forall k \in {\cal Y}$ to each UQ $x_i$, leading to
{\small 
\begin{equation}
    \mathcal{L}^{rvqa} = \frac{1}{N^{aq}_{tr}+N^{uq}_{tr}}\sum_{i=1}^{N^{aq}_{tr}+N^{uq}_{tr}} l_i,
    \label{eq:vqa uqd2}
    \vspace{-2pt}
\end{equation}
}
where $N^{aq}_{tr}$, $N^{uq}_{tr}$ is the size of $\mathcal{D}^{aq}_{tr}$ and $\hat{\mathcal{D}}^{uq}_{tr}$, respectively. 
\vspace{-2pt}
\subsection{RoI Mixup}
\vspace{-2pt}
While random pairing is effective for constructing a background dataset of UQs, it tends to produce coarse-grained UQs, where (see Fig.~\ref{fig:method}) image and question are weakly related. To increase the coverage of fine-grained mismatches, we propose an additional sampling strategy denoted as {\it RoI Mixup\/}, motivated by mixup data augmentation~\cite{zhang2018mixup,yun2019cutmix,chen2022transmix}.
Most VQA models have an object-based architecture~\cite{tan2019lxmert,uniter,villa,oscar,vinvl}, where image $v_i$ is represented as a set of $M$ (usually fixed) objects features $o_i=\{o_{i,m}\}_{m=1}^{M}$ detected by a pre-trained object detector~\cite{ren2015faster}. 
In training, RoI Mixup randomly replaces a portion $1-\lambda$, where $\lambda \in(0,1)$, of the objects in image $v_i$ with objects from another image $v_j \neq v_i$. This leads to a new and mixed set of objects $\tilde{o}_i$ 
{\small 
\begin{equation}
    \tilde{o}_i=\{o_{i,m}\}_{m=1}^{\lambda M}\bigcup \{o_{j,n}\}_{n=1}^{(1-\lambda) M} 
    \vspace{-2pt}
\end{equation}
}
with a new target one-hot vector $\tilde{y}_i=\lambda y_i$.
Note that $y_i$ can either be a correct answer, for AQs, or a zero vector, for UQs. Intuitively, by reducing the percentage $\lambda$ of original objects, the probability of the question being AQ should also shrink by $\lambda$. Fig.~\ref{fig:method} illustrates the mixing of two sets of visual features $o_1$ and $o_2$ with $\lambda=0.25$ to synthesize the object set $\tilde{o}$. 
Following ~\cite{zhang2018mixup}, $\lambda$ is sampled as $\lambda\sim\text{Beta}(1,\beta)$ where $\beta$ is a tunable hyper-parameter.

\vspace{-2pt}
\subsection{Model Ensembling}
\vspace{-2pt}
Random pairing and RoI Mixup are sampling strategies to create a background UQ dataset with a mix of coarse- and fine-grained UQs. It is also possible to improve the performance by regularizing the model output. As in the calibration literature~\cite{Balaji17,Vyas2018OutofDistributionDU}, we achieve this with model ensembles. 
Given $C$ models $\{f^c\}_{c=1}^C$, model $f^c$ predicts the probability of answer $y_k$ as $p^c(y_{k}=1|x)=f^c_{k}(x)$. Assuming  the predictions of different models are independent, the probability predicted by the ensemble is $p^E(y_k=1|x)= f^E_c(x) = \prod_{c=1}^C f_k^c(x)$. Model ensembling is then implemented by replacing $f$ with $f^E$ in (\ref{eq:model}), which produces more conservative predictions and rejects more UQs.

\definecolor{LightCyan}{rgb}{0.88,1,1}
\begin{table*}[ht!]
\begin{minipage}{0.68\linewidth}


\begin{subfloat}
\centering
\captionof{table}{RVQA comparison of recent VQA models, using MSP for the UQ detector $g$. * indicates that the model is not finetuned on GQA dataset. Larger AUAF and smaller FF95 are better.}

\adjustbox{max width=0.95\linewidth}{
\begin{tabular}{|c|ccc|ccc|ccc|ccc|c|}
\hline
\multicolumn{1}{|c|}{}&\multicolumn{3}{c|}{CLIP-Easy}&\multicolumn{3}{c|}{CLIP-Hard}&\multicolumn{3}{c|}{PT-Easy}&\multicolumn{3}{c|}{PT-Hard}&Avg.\\
Classifiers & AUAF & FF95$\downarrow$ & FACC & AUAF & FF95$\downarrow$ & FACC & AUAF & FF95$\downarrow$ & FACC & AUAF & FF95$\downarrow$ & FACC & AUAF\\
        \hline
        BUTD~\cite{uniter} & 38.45 & 64.75 & 53.50 & 36.13 & 79.14 & 53.08 & 37.83 & 66.05 & 53.02 & 33.60 & 83.11 & 51.31 & 36.50 \\
        Uniter~\cite{uniter}  & 40.03 & 73.15 & 57.08 & 39.42 & 80.48 & 57.10 & 41.45 & 61.76 & 56.82 & 35.17 & 83.52 & 55.08 & 39.01 \\
        LXMERT~\cite{tan2019lxmert}  &   42.39 & 76.25 & 0.87 & 42.60 & 78.92 & 60.49 & 47.30 & 61.79 & 59.94 & 38.12 & 85.14 & 58.76 & 42.60 \\
        SwapMix~\cite{SwapMix} &  46.31 & 71.98 & 61.05 & 42.44 & 78.41 & 60.10 & 46.19 & 62.27 & 59.77 & 37.78 & 82.73 & 58.37 & 43.18  \\
        Vilt~\cite{vilt} &  46.17 & 69.62 & 58.91 & 40.66 & 79.21 & 57.39 & 48.06 & 60.54 & 60.64 & 37.93 & 82.40 & 57.63 & 43.21   \\
        Oscar~\cite{oscar} &  45.51 & 72.14 & 62.09 & 41.76 & 80.04 & 61.72 & 46.38 & 64.27 & \bf63.44 & 39.16 & 83.15 & 60.20 & 43.2  \\
        VinVL~\cite{vinvl} &  49.86 & 69.87 & \bf64.89 & 46.36 & 78.16 & \bf64.61 & 41.68 & 84.27 & 63.38 &  41.67 & 84.26 & \bf63.37 & 44.89
  \\
        MDETR~\cite{Kamath2021MDETRM} &  47.81 & 70.32 & 62.91 & 43.86 & 78.94 & 62.05 & 47.14 & 70.04 & 62.93 & 39.04 & 84.11 & 60.30 & 44.46   \\
        BLIP-VQAv2*~\cite{li2022blip} &   35.93 & 69.39 & 51.67 & 34.94 & 82.10 & 51.13 & 37.44 & 69.33 & 52.49 & 32.62 & 86.91 & 49.79 & 35.23   \\
        \hline
    \end{tabular}
    }
    \label{tab:strong_vl}

\end{subfloat}
\begin{subfloat}
\centering
\captionof{table}{Comparison between different RVQA approaches on AUAF. Cells with light cyan background denote training with pseudo UQs. See appendix for full table with FF95 and FACC. 
}

\adjustbox{max width=0.95\linewidth}{
\begin{tabular}{|c||ccccc||ccccc||ccccc|}
\hline
\multicolumn{1}{|c||}{ }& \multicolumn{5}{|c||}{BUTD~\cite{butd}} & \multicolumn{5}{|c||}{UNITER~\cite{uniter}} & \multicolumn{5}{|c|}{LXMERT~\cite{tan2019lxmert}}\\
\multirow{ 2}{*}{RVQA Approaches} & CLIP & CLIP & PT & PT & \multirow{ 2}{*}{Avg.} & CLIP & CLIP & PT & PT & \multirow{ 2}{*}{Avg.} & CLIP & CLIP & PT & PT & \multirow{ 2}{*}{Avg.}\\
 & easy & hard & easy & hard & & easy & hard & easy & hard & & easy & hard & easy & hard &\\
\hline
\hline
\hline
FRCNN&   33.58 & 30.73 & 31.43 & 26.94 & 30.67 
&   35.81 & 33.09 & 33.67 & 28.82 & 32.84
&   38.43 & 35.22 & 35.73 & 31.00 & 35.09\\
MSP&   38.45 & 36.13 & 37.83 & 33.60 & 36.50  
&    40.03 & 39.42 & 41.45 & 35.17 & 39.01
&   42.39 & 42.60 & 47.30 & 38.12 & 42.60\\
ODIN&   38.47 & 36.14 & 37.80 & 33.60 & 36.50  
&   40.04 & 39.43 & 41.45 & 35.16 & 39.02
&   42.41 & 42.59 & 47.33 & 38.12 & 42.61\\
Maha&   30.05 & 25.75 & 25.34 & 23.93 & 26.26  
&   37.52 & 33.74 & 35.87 & 31.68 & 34.70
&   57.68 & 44.96 & 49.44 & 39.25 & 47.83\\
Energy &   38.47 & 36.19 & 37.77 & 33.67 & 36.52  
&   40.10 & 39.42 & 41.41 & 35.19 & 39.03
&   38.76 & 42.11 & 47.00 & 37.90 & 41.44\\

\cellcolor{LightCyan}Q-C & \cellcolor{LightCyan} 53.04 &  \cellcolor{LightCyan} 36.20 &  \cellcolor{LightCyan} 47.14 &  \cellcolor{LightCyan} 29.06 &  \cellcolor{LightCyan} 41.36
& \cellcolor{LightCyan}  56.61
& \cellcolor{LightCyan}  38.67
& \cellcolor{LightCyan}   50.12
& \cellcolor{LightCyan}  30.93
& \cellcolor{LightCyan}  44.08
& \cellcolor{LightCyan}  60.39
& \cellcolor{LightCyan}41.31
& \cellcolor{LightCyan}53.11
& \cellcolor{LightCyan}33.18
& \cellcolor{LightCyan}46.99
\\
\cellcolor{LightCyan}Resample & \cellcolor{LightCyan}  40.25 & \cellcolor{LightCyan} 37.73 & \cellcolor{LightCyan} 39.54 & \cellcolor{LightCyan} 34.78 & \cellcolor{LightCyan} 38.07
& \cellcolor{LightCyan}  58.66
& \cellcolor{LightCyan}48.08
& \cellcolor{LightCyan}53.65
& \cellcolor{LightCyan}39.84
& \cellcolor{LightCyan}50.05
& \cellcolor{LightCyan}  60.47
& \cellcolor{LightCyan}50.80
& \cellcolor{LightCyan}55.74
& \cellcolor{LightCyan}42.18
& \cellcolor{LightCyan}52.29
\\
\hline
\cellcolor{LightCyan} RP w/ hard UQ & \cellcolor{LightCyan}  43.74 & \cellcolor{LightCyan} 43.27 & \cellcolor{LightCyan} 37.62 & \cellcolor{LightCyan} 36.17 & \cellcolor{LightCyan} 40.2
& \cellcolor{LightCyan}  44.92
& \cellcolor{LightCyan} 47.14
& \cellcolor{LightCyan} 41.89
& \cellcolor{LightCyan} 37.92
& \cellcolor{LightCyan} 42.96
& \cellcolor{LightCyan}  53.60
& \cellcolor{LightCyan}  51.39
& \cellcolor{LightCyan}  46.95
& \cellcolor{LightCyan}  42.96
& \cellcolor{LightCyan}  48.72
\\

\cellcolor{LightCyan}RP(Ours) & \cellcolor{LightCyan}  56.31 & \cellcolor{LightCyan} 44.09 & \cellcolor{LightCyan} 50.51 & \cellcolor{LightCyan} 37.18 & \cellcolor{LightCyan} 47.02
& \cellcolor{LightCyan}  58.35
& \cellcolor{LightCyan} 48.37
& \cellcolor{LightCyan} 54.42
& \cellcolor{LightCyan} 40.27
& \cellcolor{LightCyan} 50.35
& \cellcolor{LightCyan}  60.51
& \cellcolor{LightCyan}  51.49
& \cellcolor{LightCyan}  56.08
& \cellcolor{LightCyan}  42.53
& \cellcolor{LightCyan}  52.65
\\

\cellcolor{LightCyan}Mix(Ours)& \cellcolor{LightCyan}  56.85 & \cellcolor{LightCyan} 44.70 & \cellcolor{LightCyan} 51.27 & \cellcolor{LightCyan}
37.59 & \cellcolor{LightCyan} 47.60
& \cellcolor{LightCyan}  59.08
& \cellcolor{LightCyan} 49.00
& \cellcolor{LightCyan} 54.63
& \cellcolor{LightCyan} 41.50
& \cellcolor{LightCyan} 51.05
& \cellcolor{LightCyan}  60.79
& \cellcolor{LightCyan}51.91
& \cellcolor{LightCyan}56.83
& \cellcolor{LightCyan}43.56
& \cellcolor{LightCyan}53.27
\\

\cellcolor{LightCyan}Ens(Ours) & \cellcolor{LightCyan}  \bf 57.25 & \cellcolor{LightCyan}
\bf 45.46 & \cellcolor{LightCyan}
\bf 51.95 & \cellcolor{LightCyan}
\bf 38.46 & \cellcolor{LightCyan}
\bf 48.28 
& \cellcolor{LightCyan}  \bf 59.62
& \cellcolor{LightCyan}\bf 49.65
& \cellcolor{LightCyan}\bf 55.79
& \cellcolor{LightCyan}\bf 42.14
& \cellcolor{LightCyan}\bf 51.8
& \cellcolor{LightCyan}  \bf 61.03 & \cellcolor{LightCyan} 
\bf 52.42 & \cellcolor{LightCyan} 
\bf 56.90 & \cellcolor{LightCyan} 
\bf 43.75 & \cellcolor{LightCyan} 
\bf 53.52
\\
\hline
\end{tabular}
    }
    \label{tab:main}

\end{subfloat}

\end{minipage}%
\hspace{1pt}
\begin{minipage}{0.32\linewidth}
\centering
\begin{tabular}{cc}
    \includegraphics[width=0.43\linewidth]{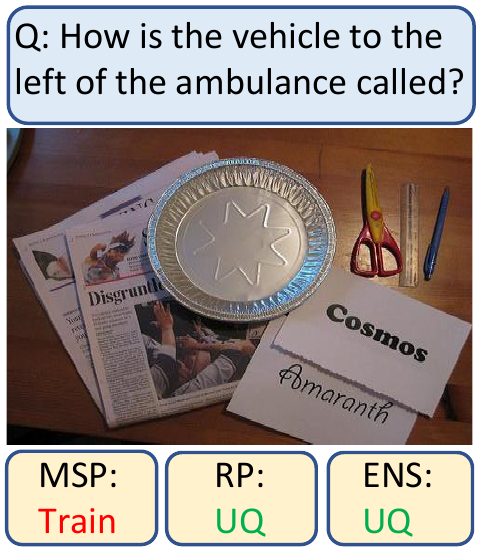} & \includegraphics[width=0.43\linewidth]{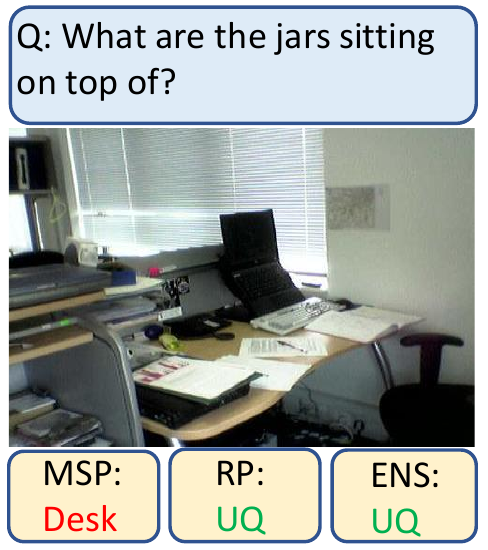} 
    \vspace{-2pt}
    \\
    \scriptsize{CLIP-Easy} & \scriptsize{CLIP-Hard}\\
    \includegraphics[width=0.43\linewidth]{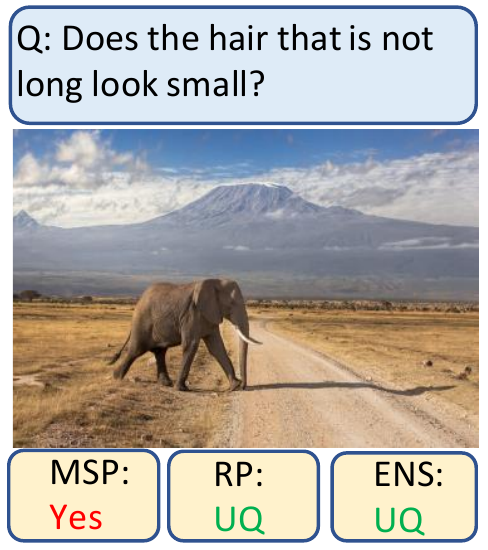} & \includegraphics[width=0.43\linewidth]{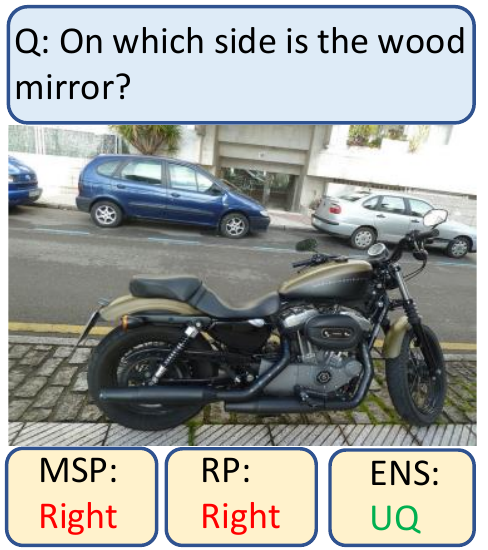}
    \vspace{-2pt}
    \\
    \scriptsize{PT-Easy} & \scriptsize{PT-Hard} \\
\end{tabular}

    \captionof{figure}{Qualitative examples for a threshold such that all models achieve $55\%$ accuracy. 
    }
    \label{fig:qualitative}
\end{minipage}

\end{table*}

\vspace{-2pt}
\section{Experiments} \label{sec:experiment}
\vspace{-2pt}


In this section, we discuss a set of experiments that leverage the proposed RGQA dataset and metrics to evaluate the RVQA performance of both existing VQA models and proposed unsupervised RVQA training techniques.
In what follows, ``RP'' means the model is trained with pseudo UQs,``Mix'' means RoI Mixup examples are also used, and ``Ens'' is the ensembling of RP and Mix. 

\subsection{Experimental Set-up}
An RVQA model consists of a VQA model $f$ and a UQ detector $g$. RVQA methods vary along three dimensions: VQA model $f$, RVQA architecture, which determines how $f$ and $g$ are combined, and RVQA approach, which uses the architecture to implement the RVQA method. We consider several models, architectures, and approaches.   

\vspace{-15pt}
\paragraph{VQA models:}
We consider the nine VQA models~\cite{butd,uniter,tan2019lxmert,vilt,Kamath2021MDETRM,oscar,vinvl,li2022blip,SwapMix}  listed in Table~\ref{tab:strong_vl}. These represent a sampling of the literature, ranging from smaller models like BUTD~\cite{butd} to recent large scale models, like VinVL~\cite{vinvl}. All models are finetuned on GQA~\cite{hudson2018gqa}, except BLIP~\cite{li2022blip} whose finetuning requirements exceed our resources. BUTD/UNITER/LXMERT were trained for 1/7/7 epochs, respectively, with the original hyperparameters. For MDETR/OSCAR/VinVL/SwapMix, we used VQA checkpoints fine-tuned on GQA from the authors' githubs. Since Vilt~\cite{vilt} does not have a GQA checkpoint, it was fine-tuned on GQA using its pre-trained weights and fine-tuning procedure from prior works~\cite{Kamath2021MDETRM,oscar}. See appendix for details.

\vspace{-15pt}
\paragraph{RVQA approaches:}
We group RVQA approaches into two categories. {\it Post-hoc, training free methods\/}  use the finetuned VQA model $f$ directly, implementing $g$ with post-hoc operations. These frequently involve thresholding a confidence score derived from the model predictions, a popular approach in the OOD literature
. {\it Training based methods\/} retrain the VQA model, using unlabeled data (pseudo-UQs), to learn $g$. The proposed RP, Mix, and Ens methods are of this type. 
We considered the following approaches.

\noindent\textit{\textbf{Post-hoc, training free methods.}} 

\textbf{MSP~\cite{hendrycks17baseline}:} Confidence score is the largest probability output by VQA model; 
\textbf{ODIN~\cite{liang2018enhancing}.} Extension of MSP that uses temperature scaling and input processing to boost performance. For RVQA, input processing is only applied to visual features. The temperature is $1e5$ and the noise $1e-4$ for all datasets; \textbf{Maha~\cite{lee2018maha}.} Estimate class-conditional Gaussian distribution of the VQA model features and use the Mahalanobis distance with respect to the closest class as confidence score. \textbf{Energy~\cite{liu2020energy,wang2021canmulti}.} Energy scoring method, initially proposed for Softmax based models~\cite{liu2020energy} and recently adapted to multi-label models~\cite{wang2021canmulti}. We find that only considering the top-$2$ classes improves performance. \textbf{FRCNN.} A rule-based method, which compares object names detected by Faster-RCNN~\cite{ren2015faster} with the nouns in the question. All object names and nouns are converted into word stems. If there exist nouns that are not in the object names, the question is declared as UQ.

\noindent\textit{\textbf{Training based methods.}}

\textbf{Resample~\cite{li2020background}.} An OOD method that performs iterative adversarial weighting of background examples (i.e. pseudo UQs),  assigns higher weights to harder examples and the reweighted dataset is trained. 
\textbf{Q-C~\cite{ray2016vtfq}.} A caption is generated per image and its  similarity to the question is measured. 
While \cite{ray2016vtfq} adopts NeuralTalk2~\cite{karpathy2015neuraltalk}, we use BLIP~\cite{li2022blip} captions. 
To measure similarity, we finetune a BERT model that takes the concatenation of a caption and a question and predicts whether the two match, with a binary score.

\vspace{-15pt}
\paragraph{RVQA architectures:} 
Several configurations of model $f$ and detector $g$ were considered. \textbf{Integrated:} sequential implementation of $g$ and $f$ as in (\ref{eq:model}); \textbf{Branched:} a common backbone with decoupled classifier heads for $f$ and $g$; \textbf{Multi-branched:} generalizes Branched by taking features from multiple layers; \textbf{Separated:} trains $g$ and $f$ separately, with different models~\cite{li2020neural}. $\bf{K+1:}$~\cite{zhou2021placeholders} defines UQs as an additional $(K+1)^{th}$ VQA class and trains $f$ as a $K+1$-class classifier. The integrated approach is applicable to all methods discussed above. The remaining architectures are only possible for training-based methods since they require pseudo-UQs to train separate $g$ heads, models, or classes. 

\subsection{Quantitative Results}
The combinatorial space of RVQA methods, VQA models, and RVQA architectures makes a comparison of all possibilities infeasible. We instead use a factorial experiment: start by ablating the architecture
given a model and method, then compare models given the best architecture, and finally compare different methods for a few models.

\vspace{-8pt}
\subsubsection{RVQA Architecture}\label{sec:pilot}\vspace{-4pt}

We started by investigating if the multiple architectures possible for trained models have any benefit over the integrated architecture of~(\ref{eq:model}), which can be universally used. These experiments used the LXMERT VQA model and RP training.  Fig.~\ref{fig:arch_human} left compares the averaged AUAF of the different architectures on RGQA. The {\it integrated\/}  architecture has top performance, followed by Separated that, besides not being universal, doubles parameter sizes and inference time. We use the integrated architecture in the following experiments.

\vspace{-8pt}
\subsubsection{VQA Model} \vspace{-2pt}
We next compared the UQ robustness of the different VQA models, using the MSP RVQA approach. Table~\ref{tab:strong_vl} shows that all models are quite vulnerable to UQs, with average AUAF across datasets below $45$. This shows that there is significant room for improvement. Interestingly, larger and more recent models do not fare significantly better than smaller models. Despite their superior AQ performance (FACC), they have similar FF95 and AUAF to the smaller models at the top of the table. Since the smaller models are much easier to train, we use them in the remaining experiments.

\vspace{-8pt}
\subsubsection{RVQA Approach} \vspace{-2pt}
We finally compared the proposed RP, Mix, and Ens to all prior RVQA approaches discussed above. In these experiments, all approaches use BUTD, UNITER or LXMERT models. Non-trainable approaches use models learned from AQs alone, trainable methods leverage a background dataset of pseudo UQs. For Mix, we empirically find the best $\beta$ value per model and use it for all subsets. See appendix for more details.
Table~\ref{tab:main} (see appendix for full table with FACC and FF95) summarizes the performance of all models on the 4 RGQA subsets. The last column is the averaged AUAF across subsets. The table allows several conclusions.

\textbf{Post-hoc approaches do not help.}
While MSP outperforms FRCNN,  post-hoc approaches like ODIN, Maha, and Energy, which do not leverage pseudo-UQ,  fail to improve on MSP. Surprisingly, these approaches have similar performance for CLIP-Easy and CLIP-Hard, even though CLIP-Easy has much coarser-grained image-question pairs.

\textbf{Pseudo UQs are effective.}
The cyan cells of Table~\ref{tab:main} show that training based methods, which leverage pseudo UQs, have significantly better RVQA performance (AUAF) than methods that do not. This is mainly due to a decrease of FF95 without sacrificie of FACC (see all metrics in appendix). Q-C consistently improves upon MSP by $5-10$ pts. Resample further improves performance for most models. However, the proposed RP improves on both, outperforming Q-C by $\sim$5.9 pts and Resample by $\sim$3.4 pts on average. This is somewhat surprising, since Resample is a more sophisticated sampling strategy. We hypothesize that Resample is unsuitable for the noisy background data generated by random pairing, likely applying larger weights to noisy examples (AQs) and hurting RVQA performance. The proposed Mix and Ens approaches have additional gains, producing the best results across VQA models. Finally, unlike prior RVQA works~\cite{li2020neural,lee2021regularizing}, RP, Mix, and Ens do not harm VQA performance, even improving FACC. See appendix for GQA test set performance.


\textbf{Impact of VQA model.}
Comparing the 3 models of Table~\ref{tab:main}, shows that RVQA approaches are more beneficial for models of higher VQA accuracy (FACC). For instance, for MSP on CLIP-Hard, from BUTD to LXMERT a FACC increase from 53.08 to 60.49 (shown in appendix)  is accompanied by an AUAF increase from around 36 to 42. This shows that better VQA reasoning skills help the model detect UQs. 
However, note that these gains saturate quickly, as shown in Table~\ref{tab:strong_vl}. Together, the two tables show that RVQA benefits more from pseudo-UQ than from large models.

\begin{figure}[t!]
    \begin{subfigure}[t]{.49\linewidth}
        \centering
        \includegraphics[width=0.99\textwidth,height=2.7cm]{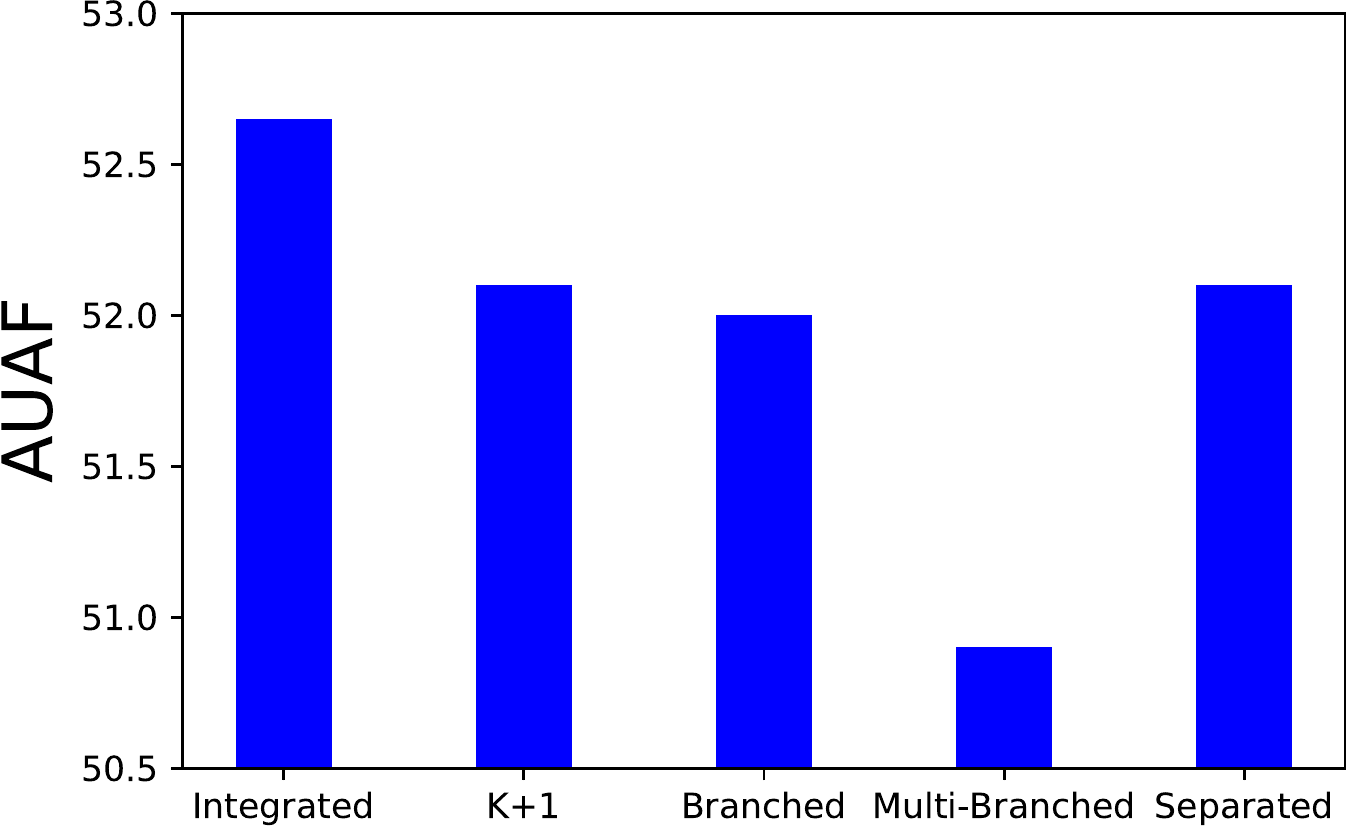}
    \end{subfigure}
    \begin{subfigure}[t]{.49\linewidth}
        \centering
        \includegraphics[width=0.99\textwidth,height=2.7cm]{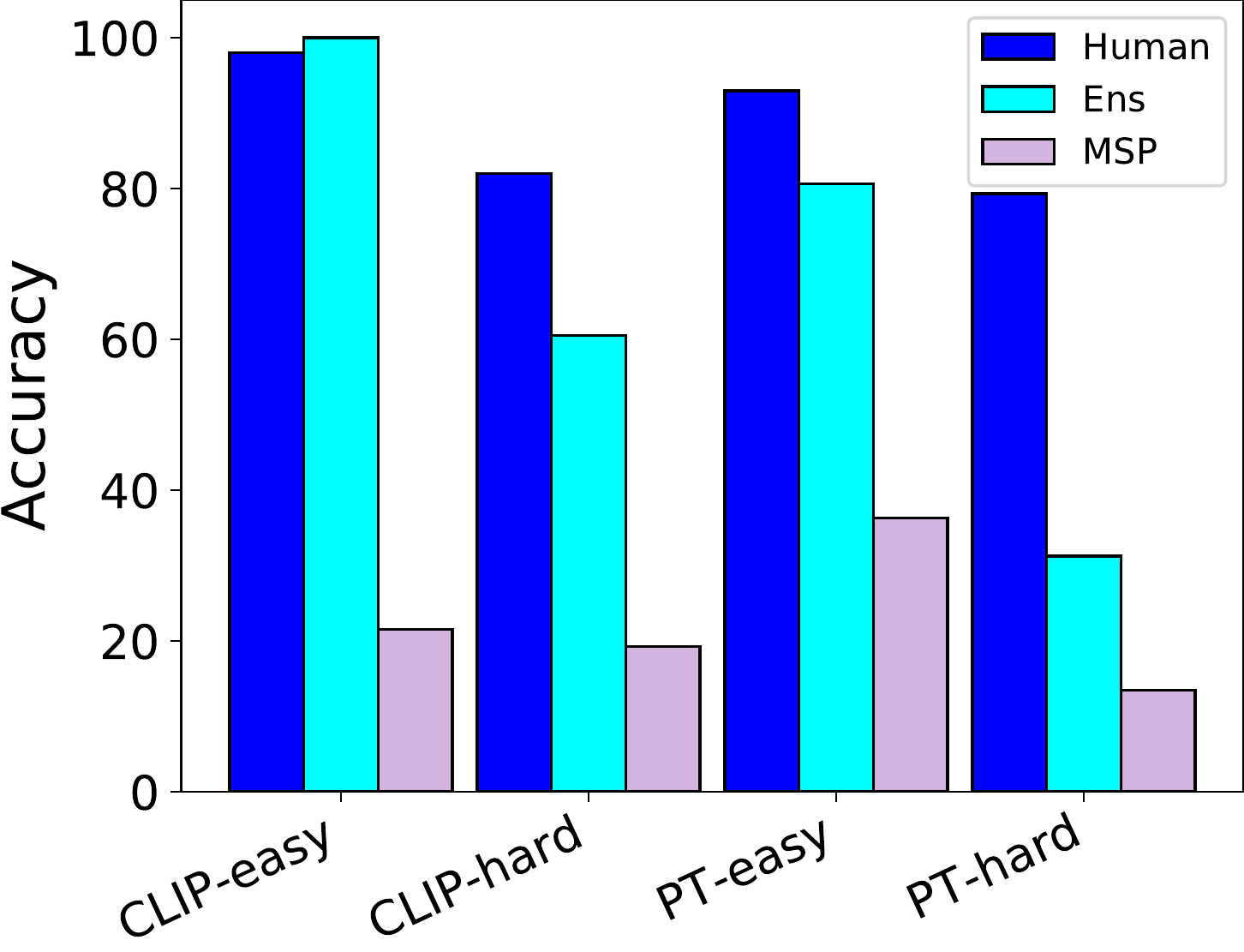}
    \end{subfigure}
    \caption{Left: RVQA architecture ablation. Right: Human evaluation.}
    \label{fig:arch_human}
\end{figure}

\textbf{UQ Diversity.} 
Most approaches achieve higher AUAF on CLIP-Easy and PT-Easy, because these 2 subsets have either low CLIP score or object level mismatch between image and question. Conversely, most approaches underperform on CLIP-Hard and PT-Hard, where UQs have subtle mismatches at attribute or relation level. This trend holds across VQA models and subsets. We also consider RP training only on hard pseudo UQs, selected by CLIP score, (RP w/ hard UQs in Table~\ref{tab:main}), which produced a weaker AUAF than  standard RP, especially on CLIP-Easy and PT-Easy. These results show the importance of UQ diversity.




\vspace{-3pt}
\subsection{Qualitative results} 
\begin{figure}
    \centering
\includegraphics[width=\linewidth,height=3.5cm]{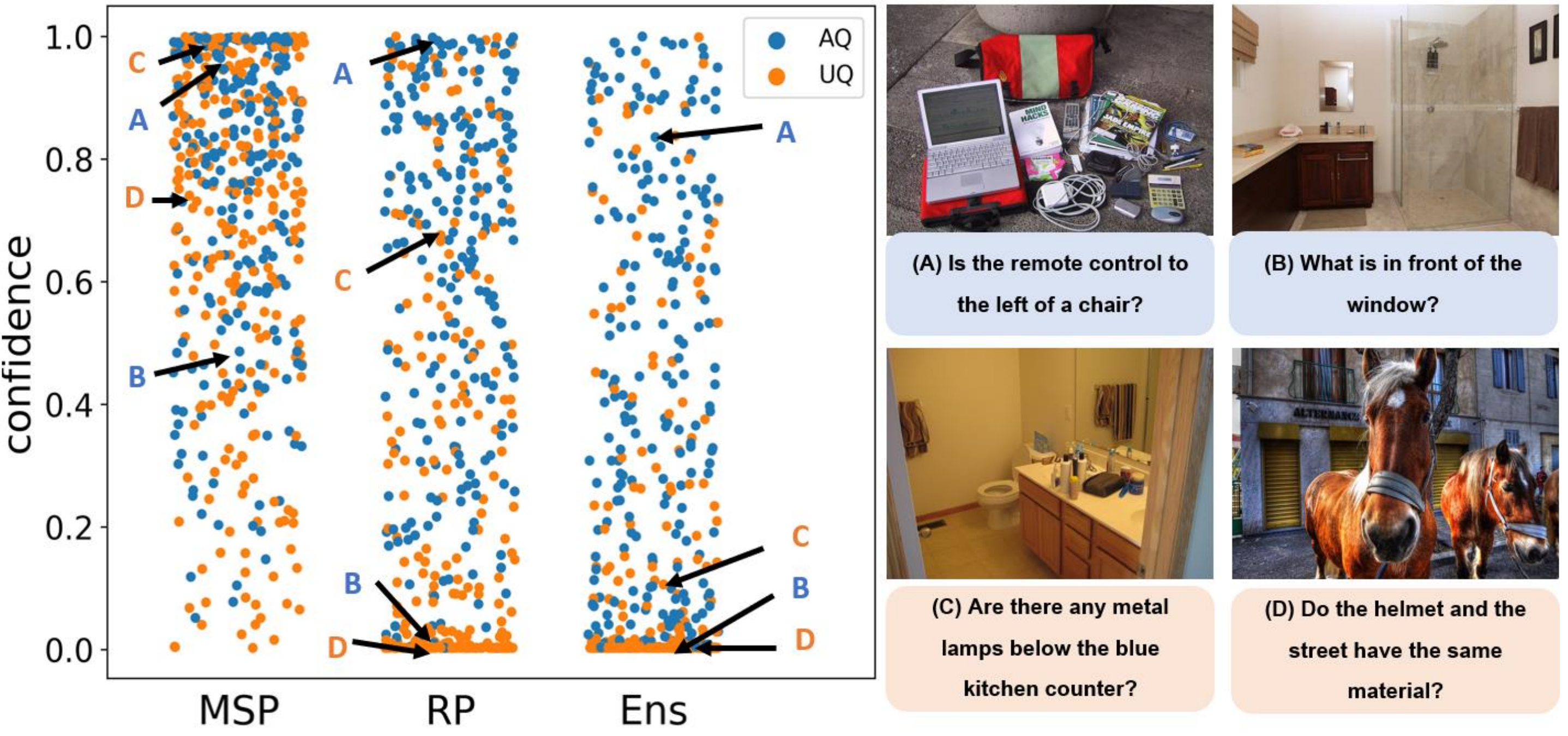}
    \caption{Left: confidence scores of MSP, RP, and Ens methods for $500$ random samples. Right: qualitative examples. AQs/UQs are shown in blue/orange. B is an annotation error in the original GQA dataset.}
    \label{fig:conf}
\end{figure}

\vspace{-2pt}
\textbf{Confidence score distribution:}
Fig.~\ref{fig:conf} compares the confidence score distribution of the post-hoc MSP approach to the proposed RP and Ens training-based methods. It shows that MSP tends to be over-confident for both AQs (blue) and UQs (orange), while RP and Ens have higher (lower) scores for AQs (UQs). MSP is also not able to capture fine-grained mismatches. For instance,  it assigns to UQ C a higher score than to AQ A. Finally, the confidence scores of AQ B show that RP and Ens can even detect incorrect annotations in the original GQA dataset.

\textbf{Model prediction:}
Fig.~\ref{fig:qualitative} shows some qualitative examples from the four subsets of RGQA. The rejection threshold is set such that all models have accuracy of $55\%$. Ens correctly rejects all UQs, and RP three of the four, while MSP fails in all cases. Note that, for the fine-grained mismatches of the hard subsets, the VQA system tends to respond by statistical association -the missing jars are ``sitting on the desk" and the nonexistent wood mirror is on the ``right," which is the side of the bike closest to the camera. 

\vspace{-3pt}
\subsection{Human Evaluation}
\label{sec:human} \vspace{-3pt}
To assess the challenge posed by the UQs in RGQA dataset, we conducted a human evaluation on MTurk. Workers were asked to perform the binary rejection on 50 AQs and 50 UQs for each subset. Fig.~\ref{fig:arch_human} right shows the rejection accuracy on UQs, comparing to models thresholded so as to achieve the same true positive rate on AQs. As expected, annotators found CLIP-Hard and PT-Hard more challenging. While Ens approaches human performance on the easier subsets, the gap on harder subsets is large. This suggests that more research is needed on the rejection of fine-grained image-question pairs.  




\vspace{-5pt}
\section{Conclusion}\vspace{-2pt}

We studied the problem of realistic VQA (RVQA) that aims to both reject UQs and answer AQs. Prior RVQA methods assume labeled UQs for training. It was argued that prior datasets are insufficient because they contain poor-quality images or lack of UQ diversity. To address this, we assembled the RGQA dataset, using 2 approaches to generate candidate UQs for human annotation. This allowed RGQA to cover broader granularities in  image-question mismatch. A combination of pseudo UQs, RoI Mixup, and model ensembles was then proposed for unsupervised training of RVQA models. Experiments show that the resulting models outperform RVQA baselines for both easy and hard UQs. Comparison to human performance shows that more research is needed in RVQA. 

\section{Acknowledgments}
This work was partially funded by NSF awards IIS-1924937 and IIS-2041009, a gift from Amazon, a gift from Qualcomm, and NVIDIA GPU donations. We also acknowledge and thank the use of the Nautilus platform for some of the experiments discussed above.

{\small
\bibliographystyle{ieee_fullname}
\bibliography{egbib}
}
\clearpage

\appendix
\counterwithin{figure}{section}
\counterwithin{table}{section}
\renewcommand\thefigure{\thesection\arabic{figure}}
\renewcommand\thetable{\thesection\arabic{table}}

\section*{Appendix}

\section{Real World VQA System}
\begin{figure}
    \centering
    \begin{tabular}{c}
       \includegraphics[width=0.95\linewidth]{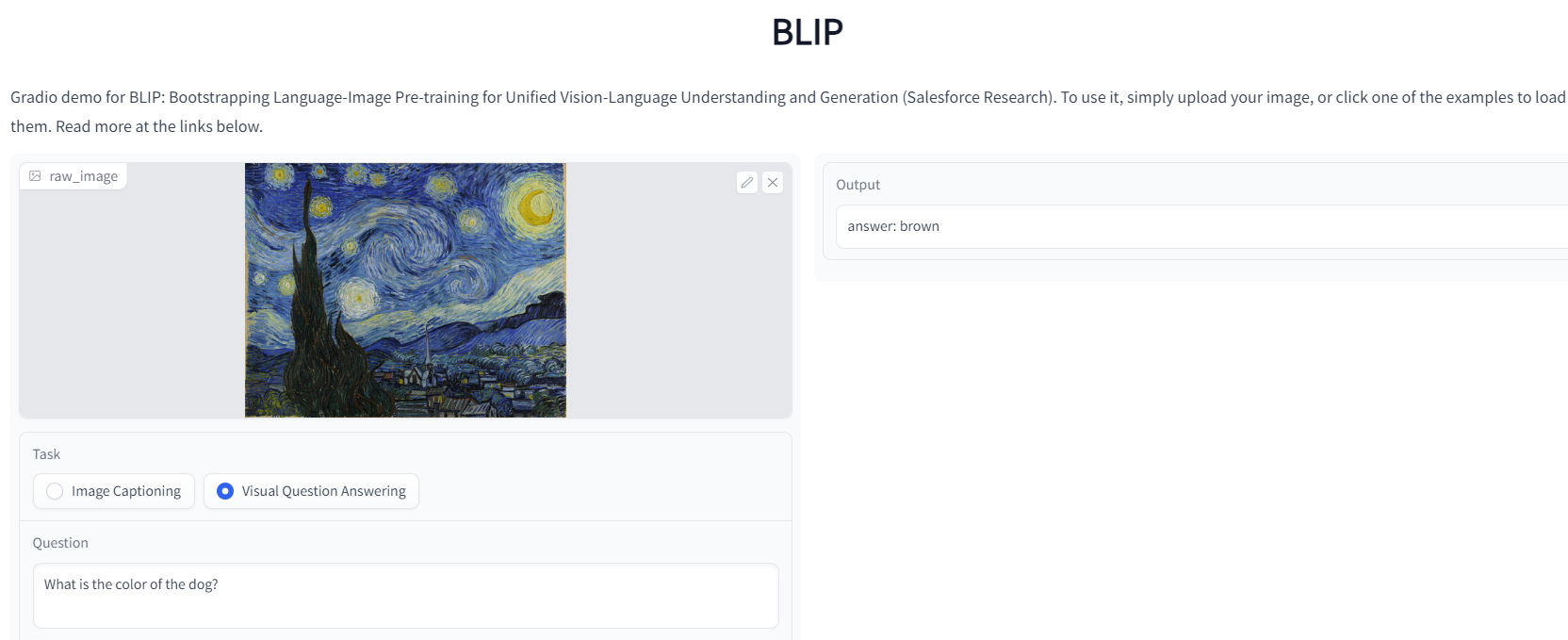}  \\  
       \includegraphics[width=0.95\linewidth]{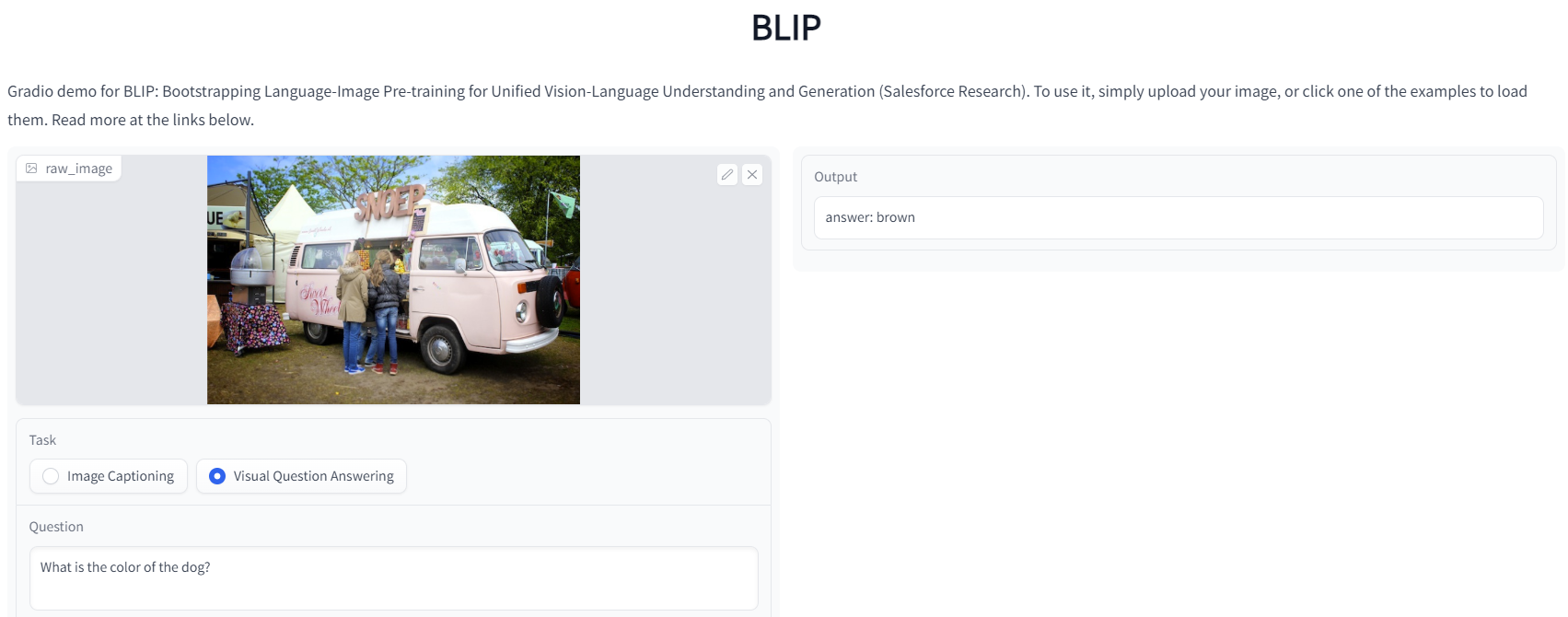} \\
        \includegraphics[width=0.95\linewidth]{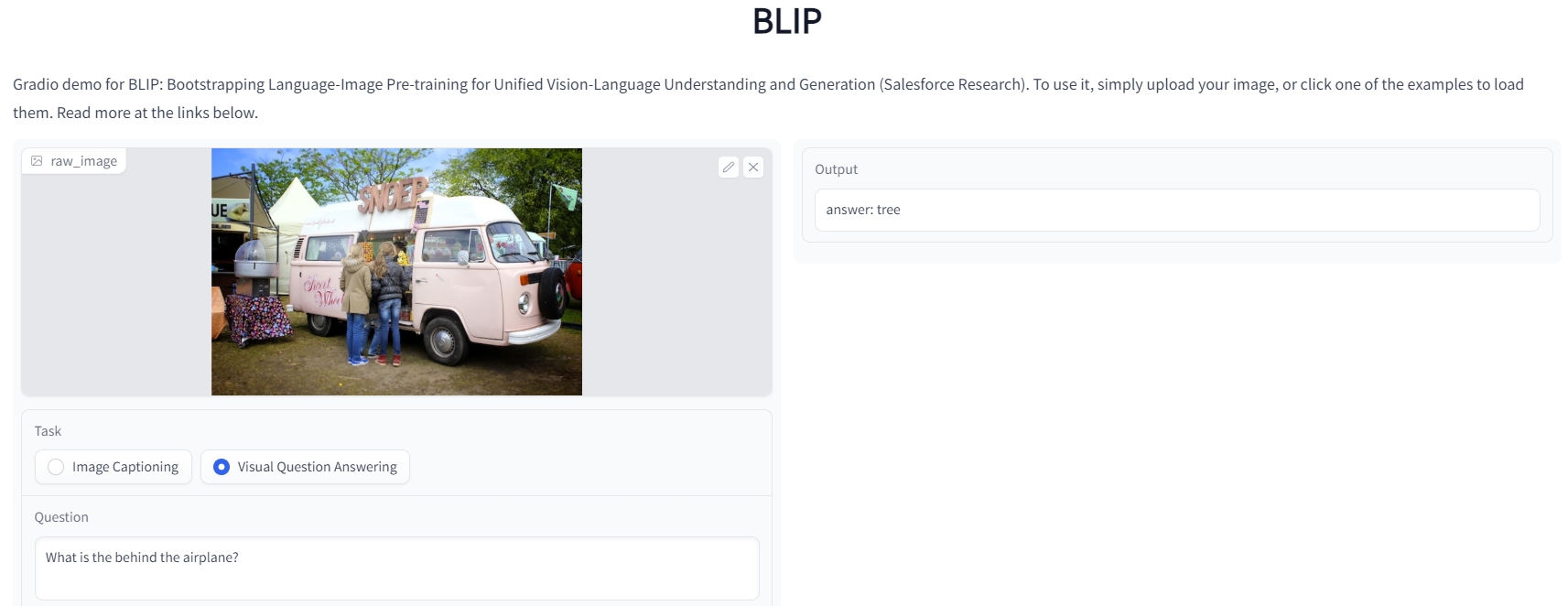}\\
        \includegraphics[width=0.95\linewidth]{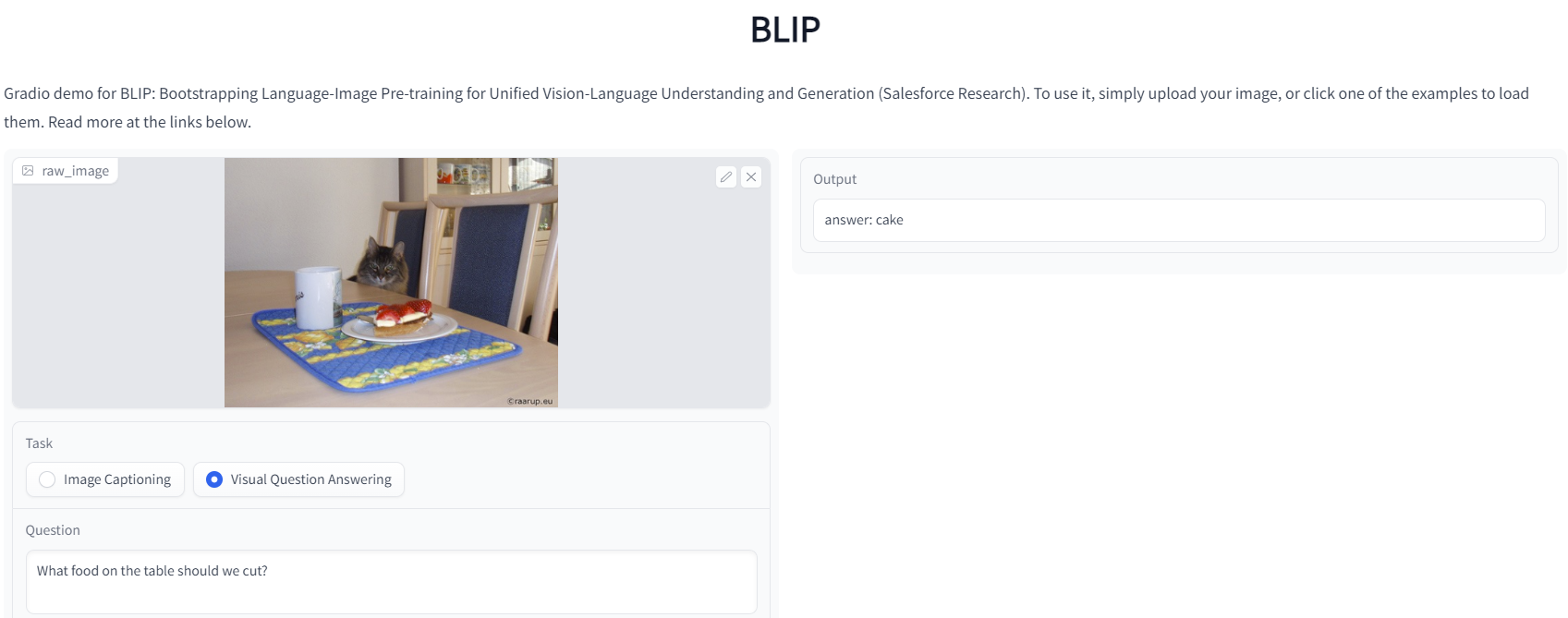}\\
        \includegraphics[width=0.95\linewidth]{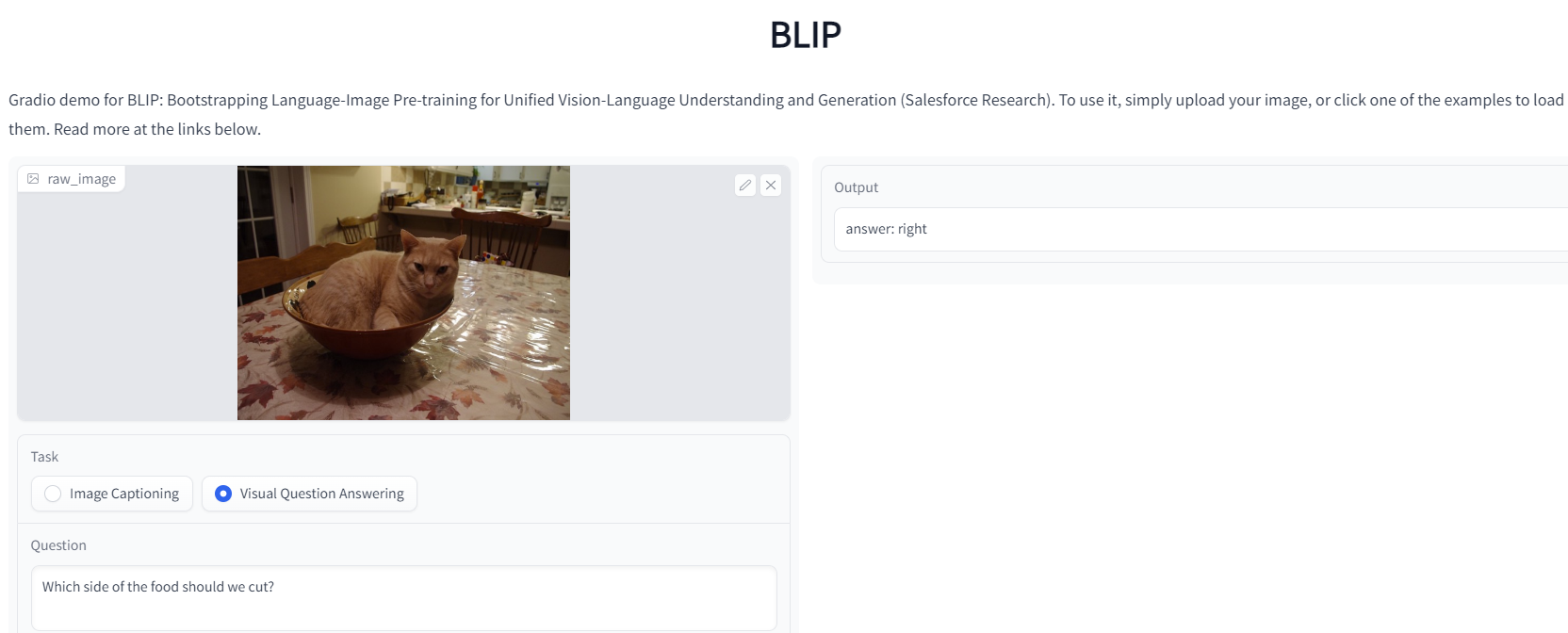}
    \end{tabular}
    
    \caption{Illustration of the RVQA problem in real-world VQA system using the BLIP~\cite{li2022blip} demo website. The top image is provided on BLIP's demo website and the rest are GQA images.  }
    \label{fig:blip_demo}
\end{figure}

As discussed in the main paper, despite that most recent VQA models have superior performance on AQs, they fail to detect UQs and underperform the proposed methods in terms of AUAF and FF95. To further evaluate these recent classifiers deployed in the real-world VQA system, we investigate the robustness of BLIP~\cite{li2022blip} under RVQA setting, using their ~\href{https://huggingface.co/spaces/Salesforce/BLIP}{online demo}. We use their provided example image and GQA image as visual inputs and ask several unanswerable questions. As shown in Fig.~\ref{fig:blip_demo}, when the user enters a question with objects that do not appear in the image, the model cannot reject or provide further instruction to the user. 
This shows that models optimizing for better AQ performance does not address the problem of RVQA, which hinders the application of real-world VQA system.

\section{Training with Hard Pseudo UQs}
Additional details for training the VQA classifiers on only hard pseudo UQs are provided. The hard pseudo UQs are the UQ pairs with higher CLIP similarity scores. We use CLIP to rank the questions for each image according to the similarity and select only top-$1,000$ questions to construct image-question pairs. As shown in Table~\ref{tab:main_full}, we observe that model that only trains on hard pseudo UQs performs similarly to our best model on CLIP-Hard and PT-Hard. However, the performance degrades significantly in terms of AUAF by around 7 and 10 points on CLIP-Easy and PT-Easy, respectively. This highlights the need for a dataset with broader coverage of UQ difficulty and indicates that overfitting VQA models on hard pseudo UQs will not address the problem of RVQA in general.

\section{Training and Evaluation Details}
In this section, the training and evaluation details of the experiments are discussed. The training is conducted using PyTorch~\cite{pytorch} for all the experiments. 
For evaluating different OOD methods, we adopt the VQA classifier of BUTD~\cite{butd} from \url{https://github.com/siddk/vqa-outliers}, LXMERT~\cite{tan2019lxmert} from \url{https://github.com/airsplay/lxmert} and Uniter~\cite{uniter} from \url{https://github.com/ChenRocks/UNITER} and \url{https://github.com/YIKUAN8/Transformers-VQA}. Both LXMERT and Uniter are initialized from pre-trained weights. For BUTD/LXMERT/Uniter, we used the optimizer of Adamax/Adam/Adam, respectively. The learning rate for BUTD/LXMERT/Uniter is set as $2e-3$/$1e-5$/$1e-5$, respectively. 
For RoI Mixup, we select $\beta$ as $0.7$/$5$/$3$ for BUTD/LXMERT/Uniter. Since VQA models use the BCE loss, methods adapted from OOD literature are based on the implementation of this multi-label OOD \href{https://github.com/deeplearning-wisc/multi-label-ood}{github}. We also use CLIP from \href{https://huggingface.co/}{huggingface} and POS tagger from \href{https://spacy.io/}{Spacy} to process the text.

For the comparison of  9 different VQA classifiers~\cite{SwapMix,tan2019lxmert,butd,uniter,vinvl,oscar,Kamath2021MDETRM,li2022blip,vilt}, we further adopt the pretrained checkpoint on GQA of SwapMix~\cite{SwapMix}, Oscar~\cite{oscar}, VinVL~\cite{vinvl} and  MDETR~\cite{Kamath2021MDETRM} from their official github links. We also finetune Vilt on GQA following the procedure in \cite{Kamath2021MDETRM,oscar}, because Vilt~\cite{vilt} only released the checkpoint from its pretraining stage and does not have checkpoint finetuned on GQA. For BLIP~\cite{li2022blip}, we directly downloaded its checkpoint from their \href{https://github.com/salesforce/LAVIS}{github link}, which is trained on Visual Genome~\cite{krishnavisualgenome} and VQA2.0~\cite{Goyal2017MakingTV} dataset. Due to the computation constraint of our GPU cluster, we are not able to finetune BLIP on GQA. However, since GQA is also built on Visual Genome, we measure the GQA performance of BLIP without fine-tuning its checkpoint. Note that BLIP supports open-ended VQA, so we follow its VQA setting and use its decoder to rank the GQA candidate answers (rank 1 is selected as prediction). The comparison between different VQA classifiers uses the maximum probability (MSP) as UQ/AQ criterion.

\section{Additional RGQA Details} \label{sec:rgqa_details}
\begin{figure*}
    \centering
    \begin{tabular}{cc}
    
    \multicolumn{2}{c}{\includegraphics[width=0.8\textwidth]{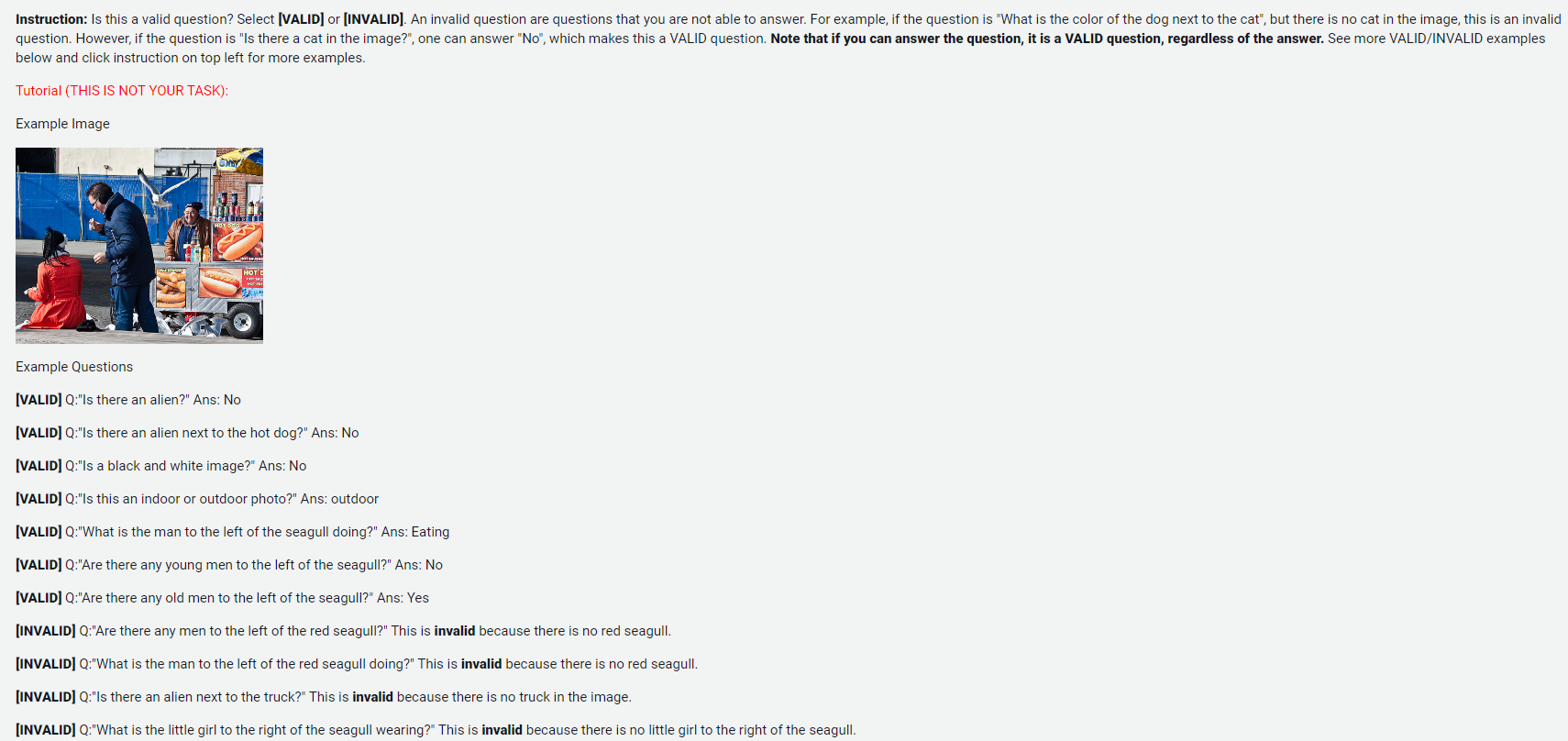}} \\ 
    \multicolumn{2}{c}{(a) Instruction to the annotators.}
    \\
    \includegraphics[width=0.25\textwidth, height=0.38\textwidth]{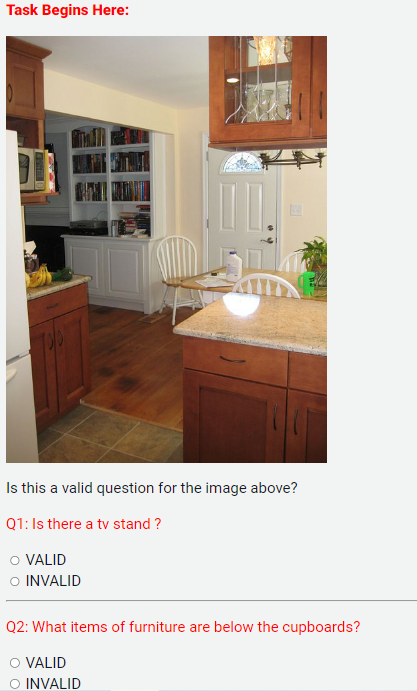} & 
    \includegraphics[width=0.25\textwidth, height=0.38\textwidth]{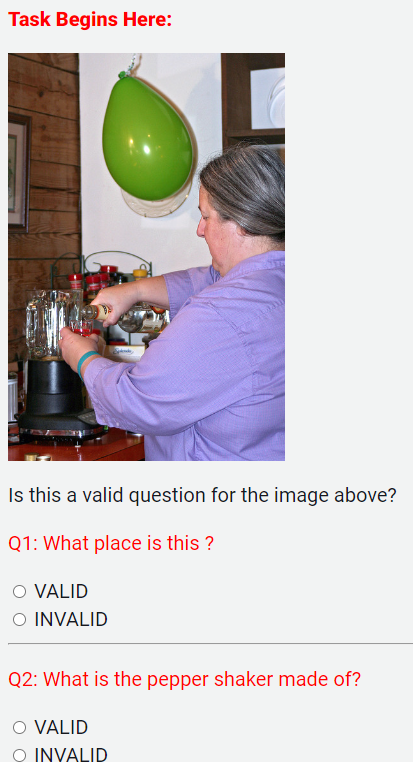} \\
    (b) Example 1 & (c) Example 2 \\
    \includegraphics[width=0.25\textwidth, height=0.38\textwidth]{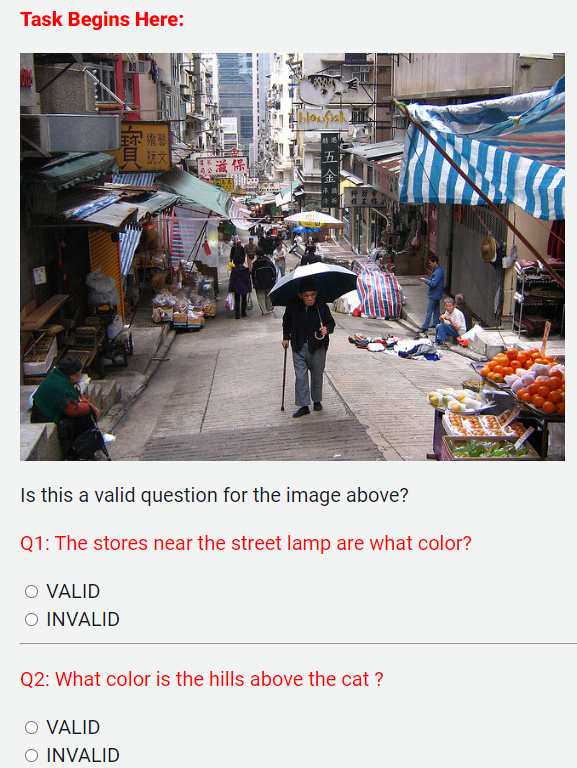} & 
    \includegraphics[width=0.28\textwidth, height=0.38\textwidth]{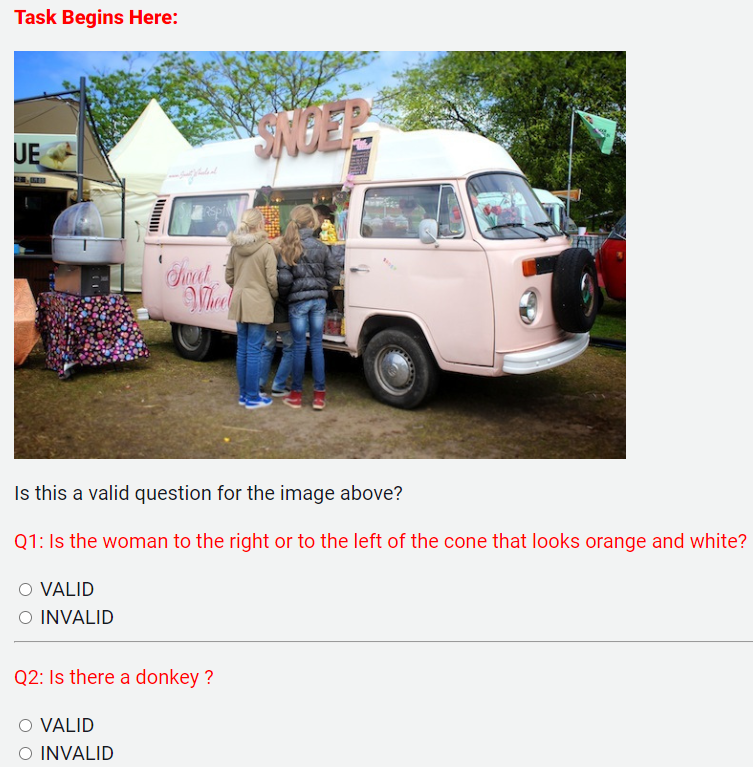}\\
    (d) Example 3 & (e) Example 4
    
    \end{tabular}
    \caption{The annotator is asked to read the instruction in (a). (b-e) are the task that are assigned to the annotator. See text in Sec.~\ref{sec:rgqa_details} for more details. }
    
    \label{fig:mturk}
\end{figure*}

\textbf{Dataset Annotation.}
As mentioned in Sec. 3.1, the candidate UQs are passed to the annotators. The annotators are asked to read the instruction with few AQs (i.e. Valid) and UQs (i.e. Invalid) examples, as shown in Fig.~\ref{fig:mturk}(a). After reading the instruction, the annotator is given the tasks, where 2 questions in random order and an image are given. 
One of the questions requires annotation, while the other question is the ``filter question". The filter question is used to ensure that the annotator fully comprehends the task and is paying attention during the process.  Some examples of the task are shown in Fig.~\ref{fig:mturk}(b-e). Take Fig.~\ref{fig:mturk}(b) for example. 2 questions are presented to the annotator and the first question ``Is there a tv stand?" is the filter question. The annotator is expected to answer ``valid" for the filter question since it is an answerable question with answer ``No". Fig.~\ref{fig:mturk}(d) is another example, where the filter question of ``What color is the hills above the cat?" is shown on the second question. Obviously, the annotator is expected to answer ``invalid", because there is no hill above the cat in the image.

More specifically, the filter question is guaranteed to be either answerable or unanswerable. To create filter questions automatically, we extract all the object names from the annotated scene graph in GQA~\cite{hudson2018gqa} and curate a set of object names. 
For the candidate set of answerable filter questions, the template of ``Are/Is there a $\langle$obj$\rangle$?" is used, where $\langle$obj$\rangle$ is a randomly selected object from the object set. Furthermore, the candidate set of answerable filter questions is augmented with ``"Is this indoor or outdoor?", ``Is this a color image?" and ``What place is this ?" to increase the diversity of answerable filter questions.

For the candidate set of unanswerable questions, we adopt the template of ``What color is the $\langle$obj0$\rangle$ $\langle$rel$\rangle$ the $\langle$obj1$\rangle$?", where $\langle$obj0$\rangle$ and $\langle$obj1$\rangle$ are 2 randomly selected objects from the object set and the $\langle$rel$\rangle$ is a randomly selected relation from a set of predefined relations (e.g. next to, around, under, on and above).

\textbf{AQ vs UQ Ratio.}
As mentioned in the main paper, each UQ is paired with an AQ. However, this could result in duplicated AQs, because the number of UQs could be larger than that of AQs for some images. As a result, the duplicated AQs are removed from the proposed dataset, which explains the reason that the proportion of UQ in the main paper is around 52\%.

\begin{figure*}[ht!]
    \begin{subfigure}{.49\linewidth}
    \centering
        \includegraphics[trim={3cm 0 0 0},scale=0.33]{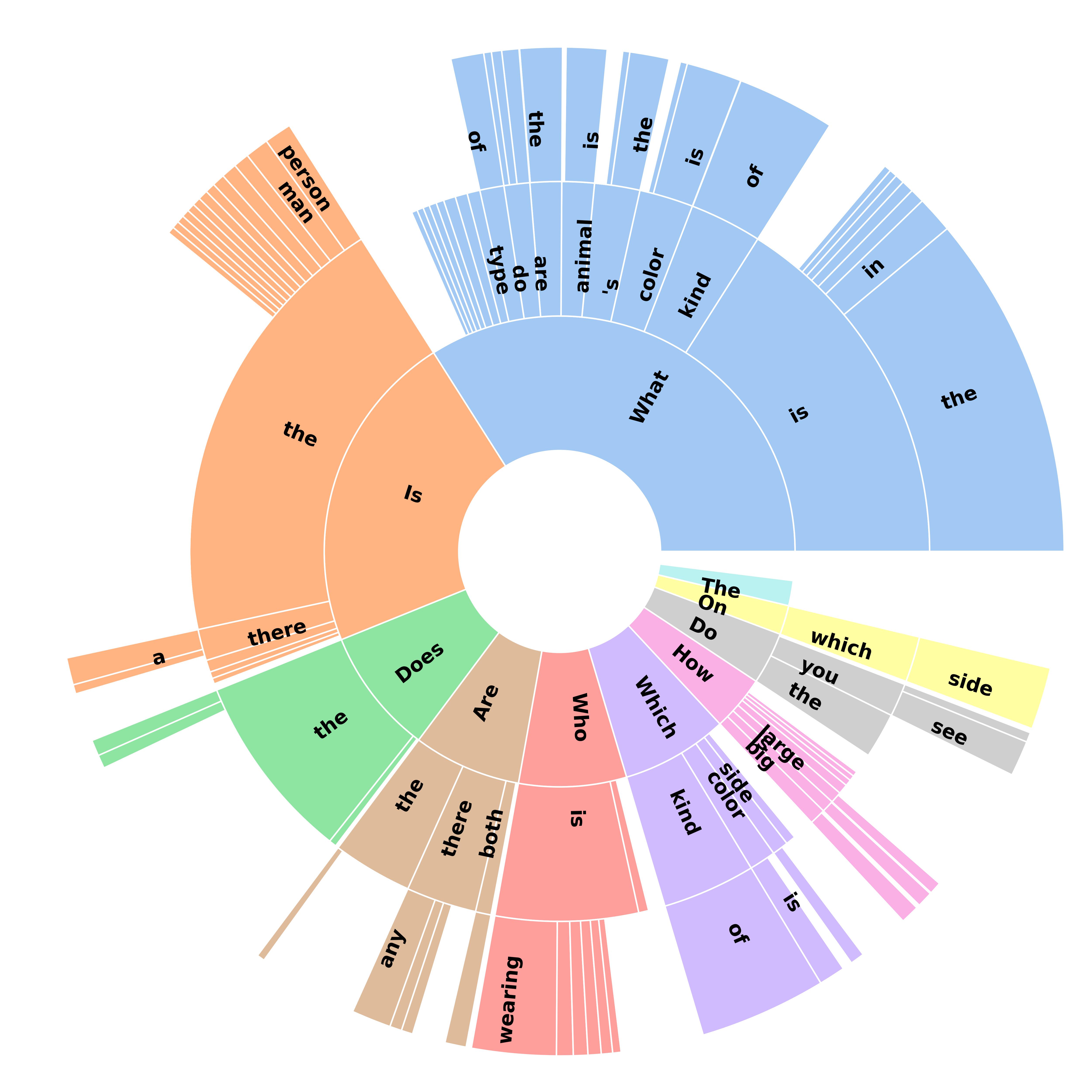}
        \caption{Answerable question (AQ)}
        \label{subfig:first words aq}
    \end{subfigure}
    \begin{subfigure}{.49\linewidth}
    \centering
        \includegraphics[trim={3cm 0 0 0},scale=0.33]{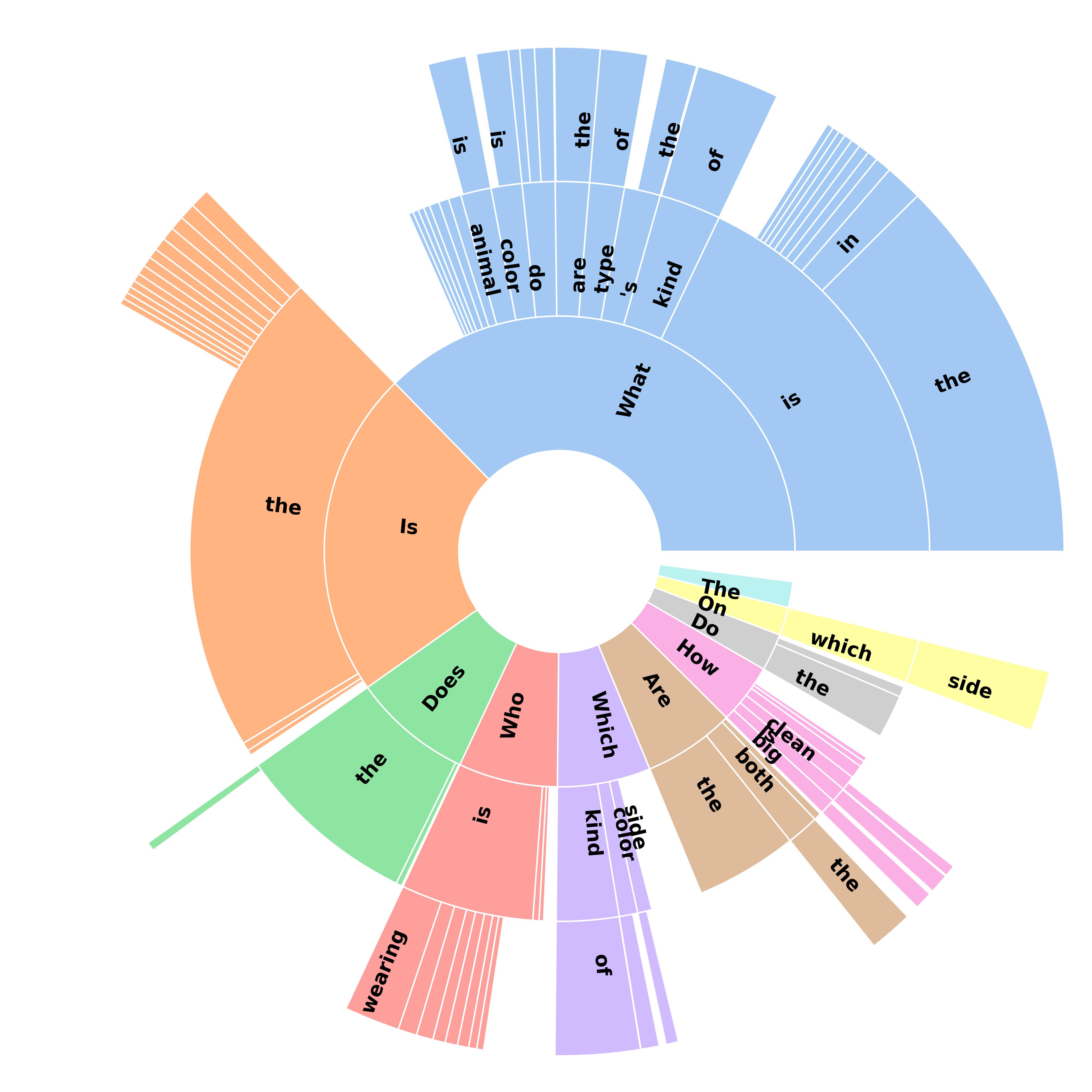}
        \caption{Unanswerable question (UQ)}
        \label{subfig:first words uq}
    \end{subfigure}
    \caption{Distribution of questions by first three words for all subsets in RGQA. 
    The white regions are marginal probabilities for those less populated words.}
    \label{fig:first words}
\end{figure*}

\textbf{AQ vs UQ Question Structure}
We further analyze the difference between AQ and UQ from its question structure. This is done by plotting the distribution of questions by the first three words, as shown in Fig.~\ref{fig:first words}, While the three most popular words (``Are'', ``Who'' and ``Which'') in AQs and UQs have minor difference in their order and proportions, there are no major differences between the question structure of AQ (Fig.~\ref{fig:first words}(a)) and UQ (Fig.~\ref{fig:first words}(b)). This indicates that AQs/UQs cannot be easily separated by word frequency and distribution.

\textbf{Conflicting Candidate UQs Removal:} Conflicting Candidate UQs like “What color are the black shoes?” are filtered using predefined rules. For example, for a question asking  about color, a program checks whether the answer (i.e. black) is in the text.

\textbf{CLIP Bias:}
We notice that CLIP might have a bias when producing UQs (e.g. confuses attributes of multiple objects in the same image~\cite{Yamada2022WhenAL}). We prevent these biases by introducing PT-based UQs and using human annotators to confirm the validity of UQs. Obviously, there could be other biases, e.g. a preponderance of certain types of objects in the dataset. The characterization of these is a project in itself and left for future work.

\section{Additional Related Work}
\textbf{Visual language pretraining (VLP)}~\cite{vinvl,vilt,oscar,vlp,UnicoderVLAU,visualbert,Lu202012in1MV,ViLBERT,villa,uniter,tan2019lxmert,li2022blip} has been a dominated manner to learn generalizable multi-modal feature for visual language task. During the pre-training stage, the models are usually trained on self-supervised tasks: (a) masked language modeling, (b) masked image modeling, and (c) image-text matching. For (a) and (b), the model predicts the masked words and masked patches using the rest of the unmasked text and image. For (c), the image and text are randomly paired and the model is asked whether the pair are matched. The universal feature from the pre-training stage are shown to be applicable to various downstream tasks, including image captioning~\cite{vilt,vlp,vinvl}, visual grounding~\cite{uniter, villa, ho2022yoro,Kamath2021MDETRM} and VQA~\cite{Lu202012in1MV,tan2019lxmert,ViLBERT}. For more detailed related work, please refer to recent surveys~\cite{Du2022ASO,RUAN20221}.

Despite that most of the recent VQA models~\cite{Lu202012in1MV,tan2019lxmert,ViLBERT,vlp,vinvl,oscar,visualbert,ViLBERT,villa,uniter} are fine-tuned after VLP, there is no evidence that these model are robust toward UQ. Our experiments analyze the SoTA VQA models~\cite{tan2019lxmert,uniter,oscar,vinvl,li2022blip} and found their vulnerability to the UQs.
This is surprising since the proxy task of image-text matching is optimized during the pretraining stage.
In this work, we proposed a new training scheme specifically tailored for the RVQA task, which does not require any annotated UQs during training and is robust to UQ during inference.

\textbf{Mixup}
Inspired by the mixup data augmentation~\cite{zhang2018mixup,yun2019cutmix,chen2022transmix}, we proposed RoI mixup to encourage the RVQA model to be aware of the fine-grain mismatch between image and text. 
While the proposed RoI mixup is similar to \cite{SwapMix} and \cite{simvqa}, the goal is entirely different.\cite{SwapMix} mitigates the reliance of the VQA model on irrelevant background, while \cite{simvqa} mixes object features of the same classes from different domains, to achieve domain invariance between synthetic and real VQA datasets. Our aim is to detect UQs, which these methods cannot. For example, although \cite{SwapMix} leverages scene graph annotation of objects "relevant" to the text, to swap object features, it fails to detect UQs, as shown in the experiment section. No additional annotation is needed for the proposed RoI Mixup.

\textbf{VQA-OOD}
While there are certain similarities between OOD and realistic VQA, they are different.
\cite{mutant,KV22, XGGM} address OOD where the distributions of training and test set are different. However, \textit{there is no answer} for UQs, so \cite{mutant,KV22, XGGM} are not applicable to the proposed RVQA task.

\section{GQA test set performance}
The goal of realistic VQA is to detect the UQs without sacrificing the VQA performance on AQs. To ensure that the proposed training strategy does not degrade the VQA performance, we report the accuracy of the GQA test set. As shown in Table~\ref{tab:gqa_test}, the original LXMERT~\cite{tan2019lxmert} achieves 77.8 accuracy, while the proposed RP and Mix have similar performance. We also report the accuracy for binary (e.g. yes/no) questions and open questions.

\begin{table}
    \centering
    \caption{GQA test set performance without rejections. The backbone is LXMERT.}
    \scalebox{0.99}{
    \begin{tabular}{|c|c|c|c|}
    \hline
    Methods & Accuracy & Binary & Open\\
    \hline
    \hline
    Original & 77.8 & 45.0 & 60.3 \\
    RP & 76.3 & 45.6 & 60.0 \\
    Mix & 77.1 & 45.8 & 60.5 \\
    Ens & 77.1 & 46.2 & 60.7 \\
    \hline
    \end{tabular}
    }
    \label{tab:gqa_test}
\end{table}

\section{Additional examples from RGQA}
\begin{figure*}
    \centering
    \begin{tabular}{cccc}
        \includegraphics[width=0.22\textwidth,valign=t]{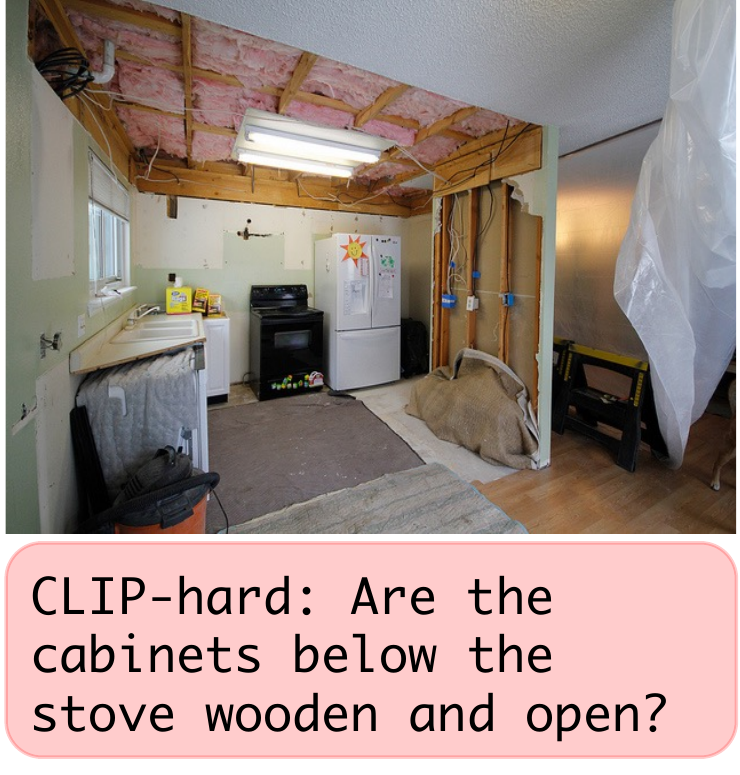} &
        \includegraphics[width=0.22\textwidth,valign=t]{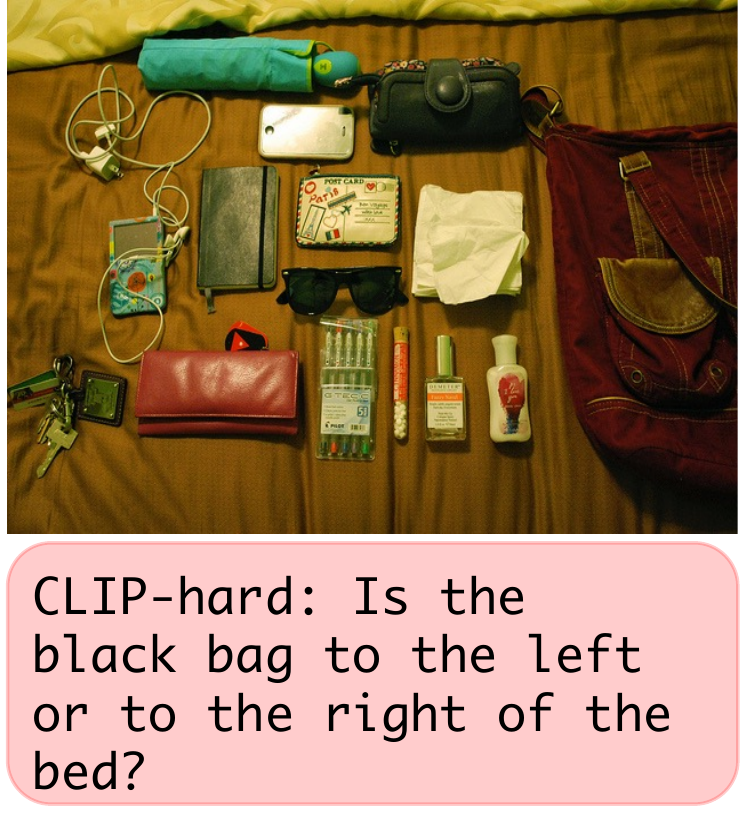} &
        \includegraphics[width=0.22\textwidth,valign=t]{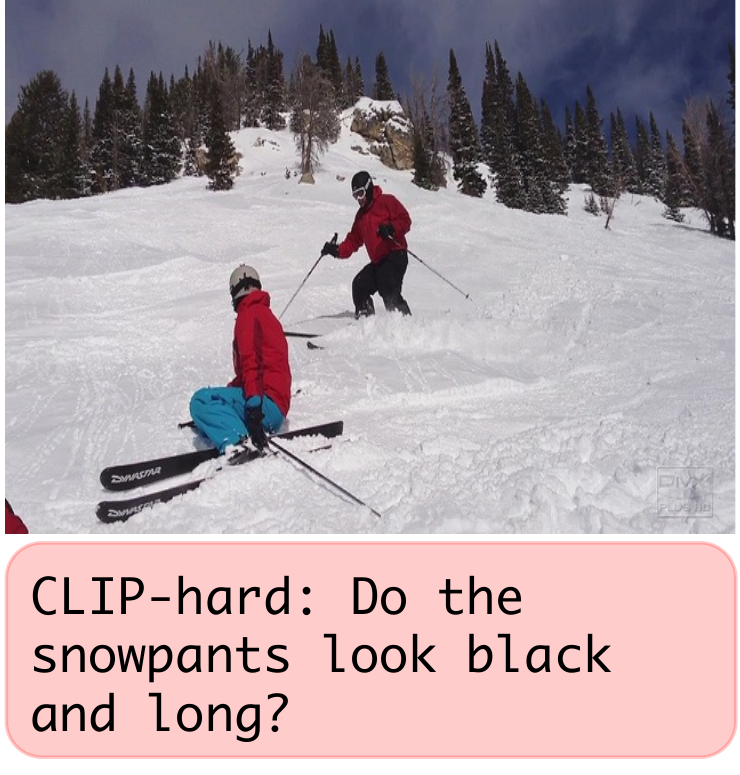} &
        \includegraphics[width=0.22\textwidth,valign=t]{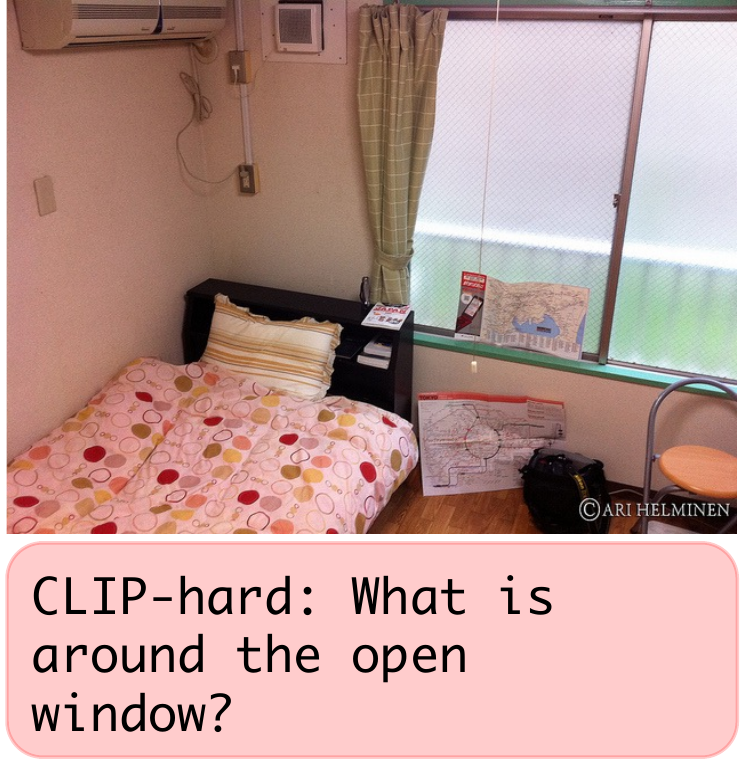} \\
        (a) & (b) & (c) & (d) \\
        \includegraphics[width=0.22\textwidth,valign=t]{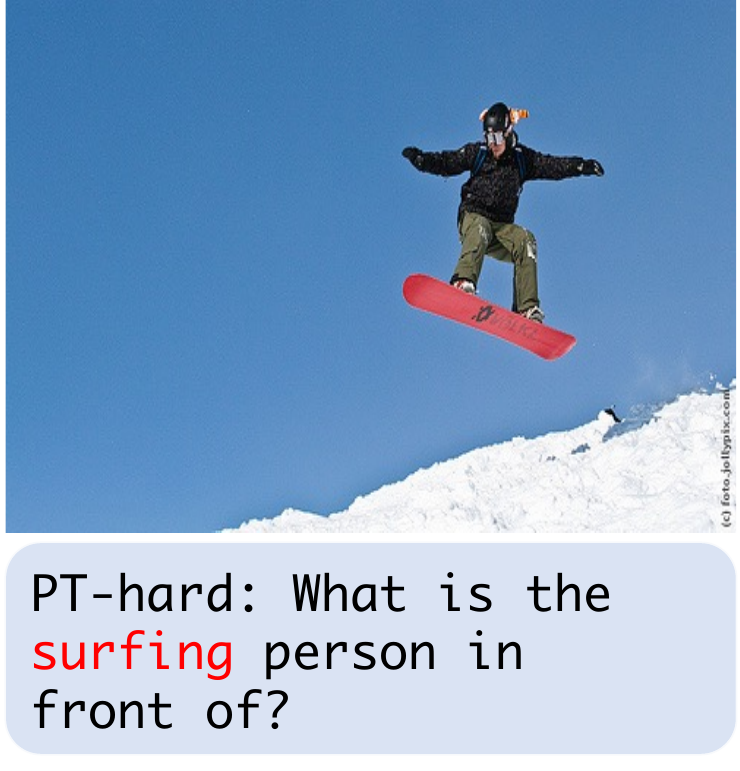} &
        \includegraphics[width=0.22\textwidth,valign=t]{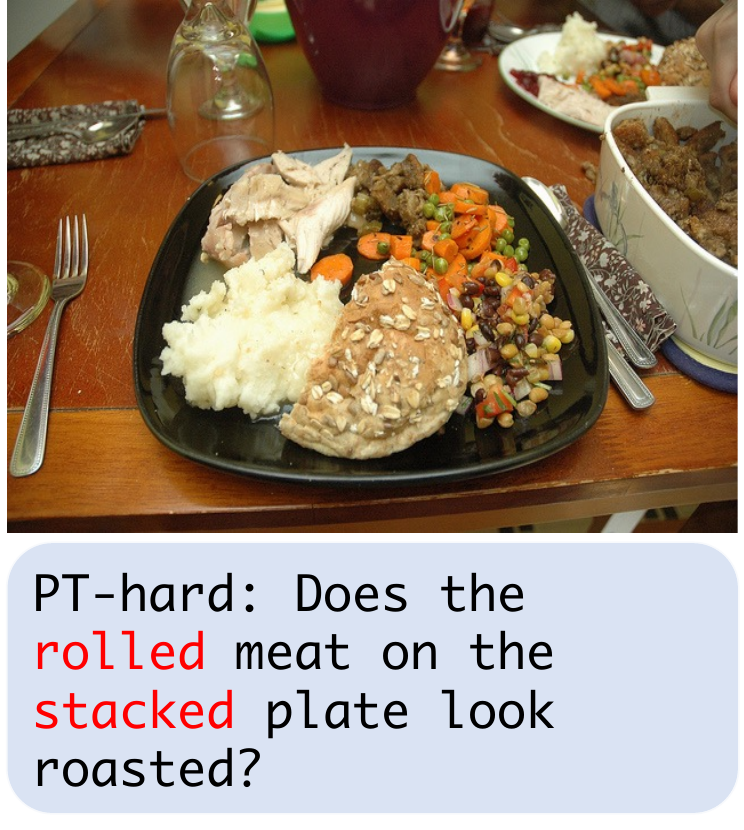} &
        \includegraphics[width=0.22\textwidth,valign=t]{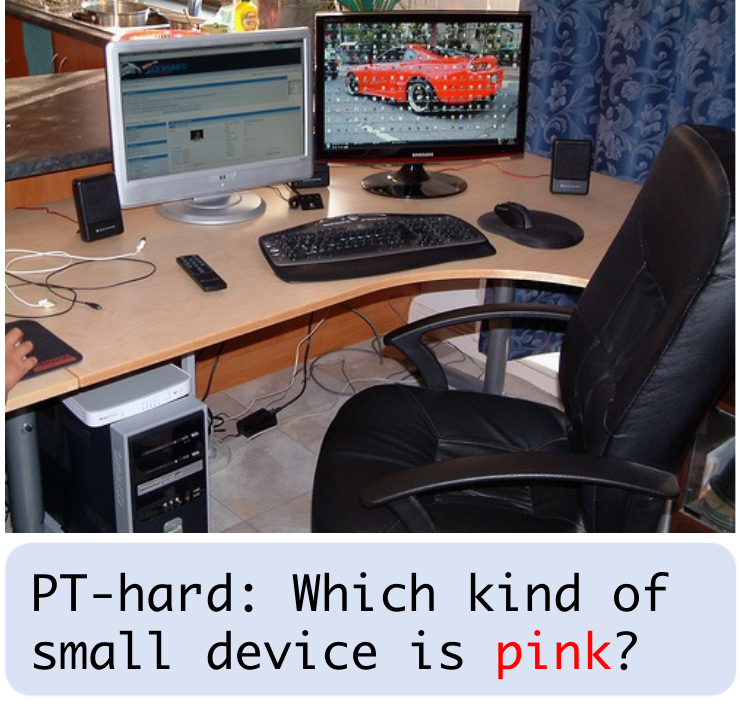} &
        \includegraphics[width=0.22\textwidth,valign=t]{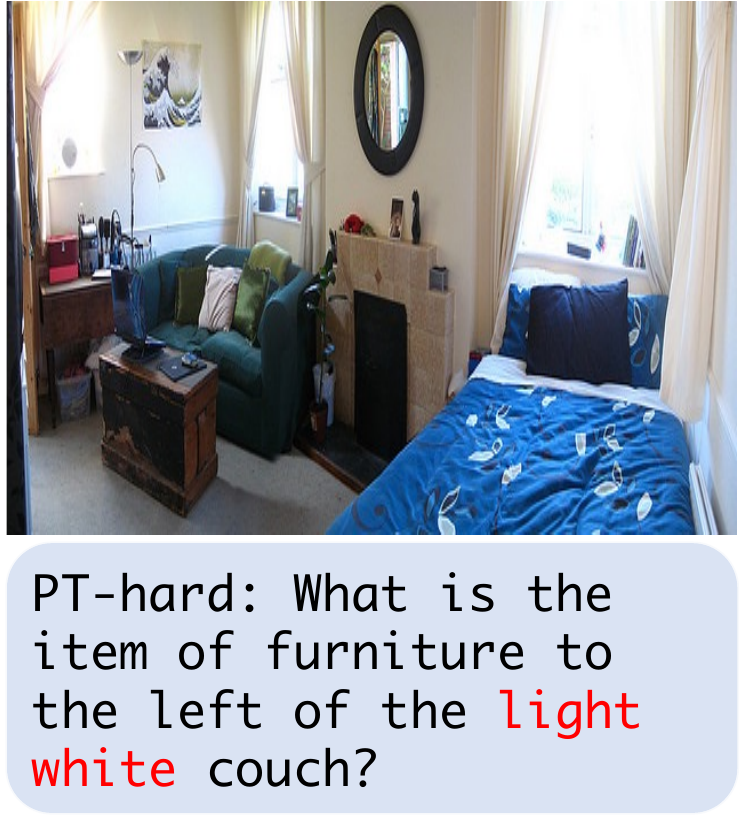} \\
        (e) & (f) & (g) & (h) \\
        \includegraphics[width=0.22\textwidth,valign=t]{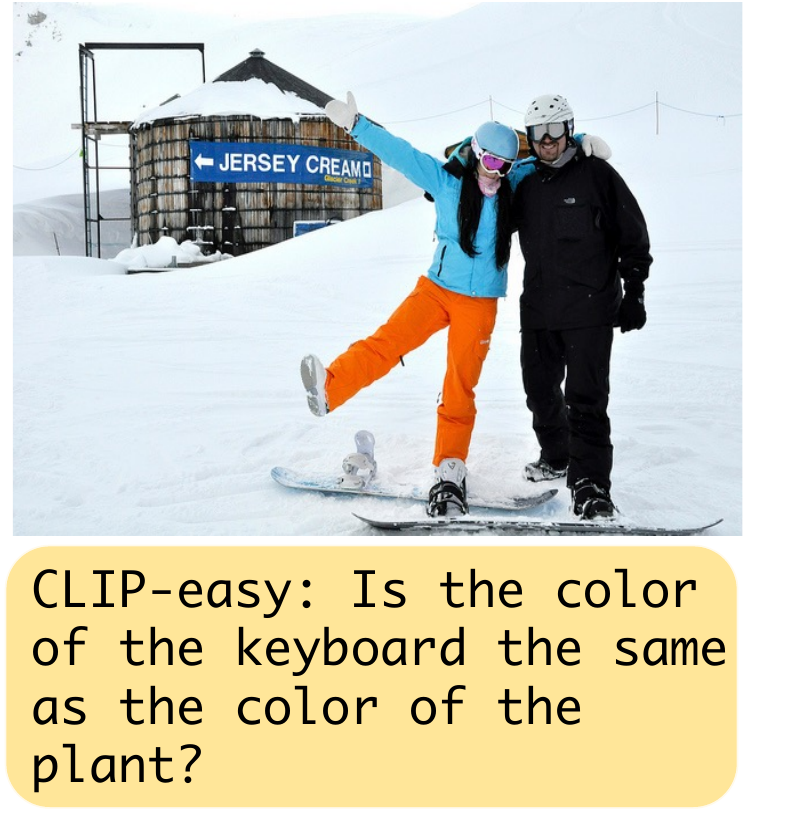} &
        \includegraphics[width=0.22\textwidth,valign=t]{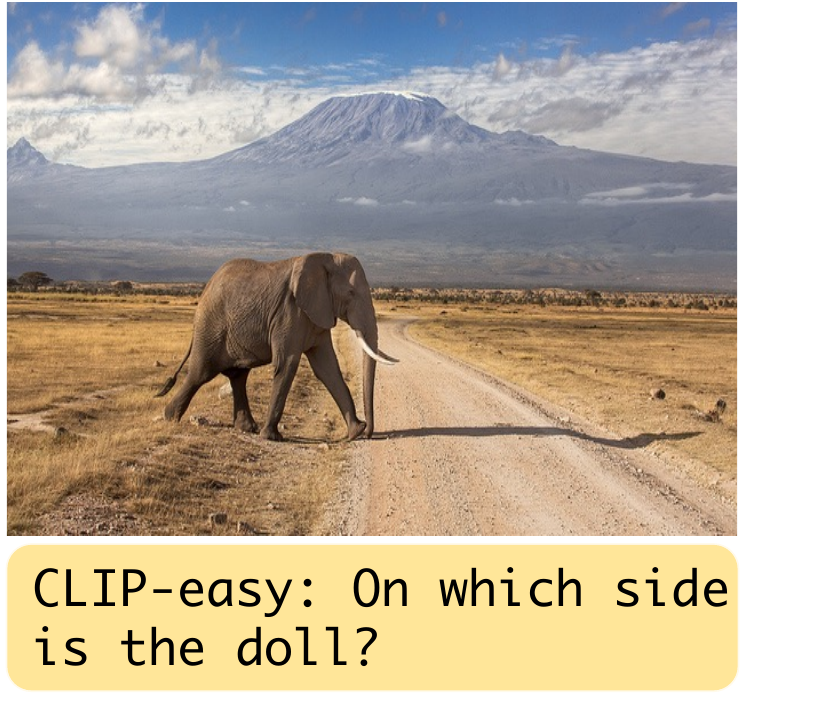} &
        \includegraphics[width=0.22\textwidth,valign=t]{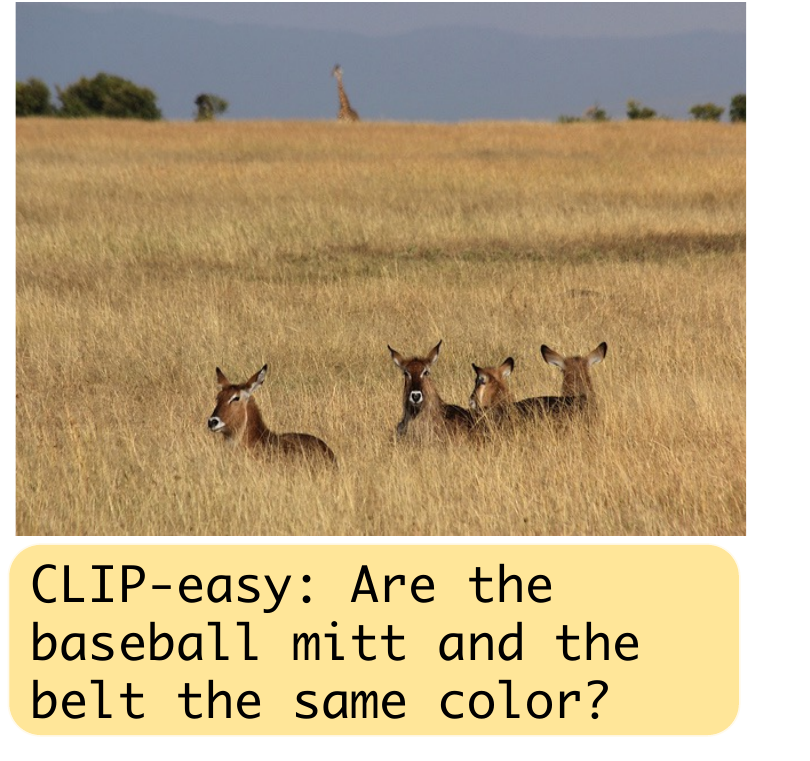} &
        \includegraphics[width=0.22\textwidth,valign=t]{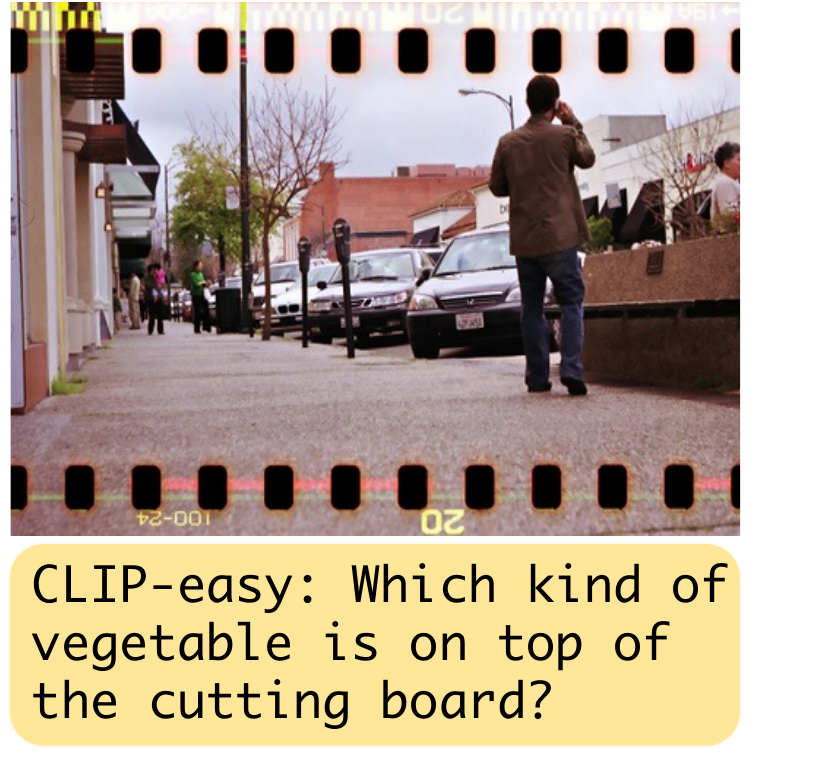} \\
        (i) & (j) & (k) & (l) \\
        \includegraphics[width=0.22\textwidth,valign=t]{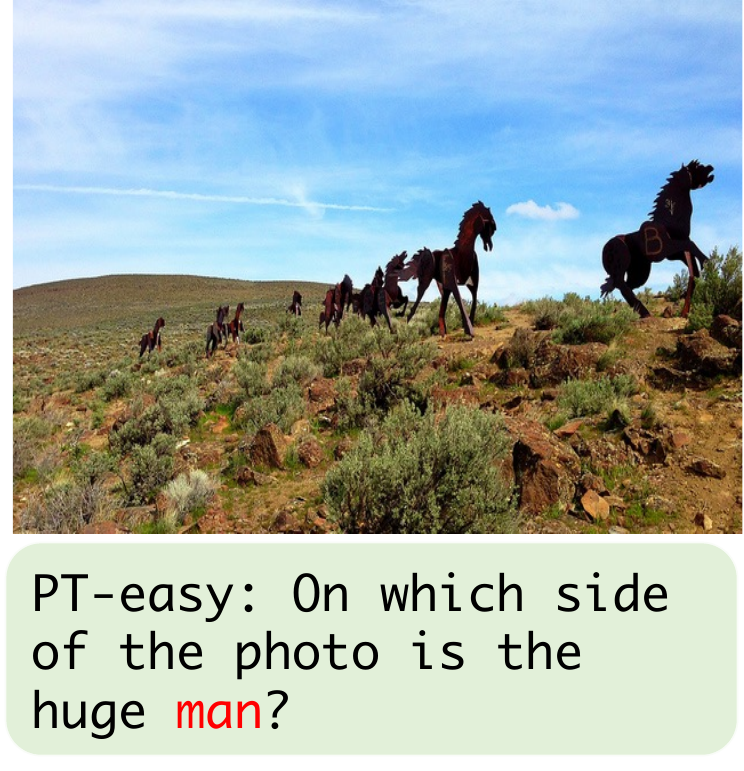} &
        \includegraphics[width=0.22\textwidth,valign=t]{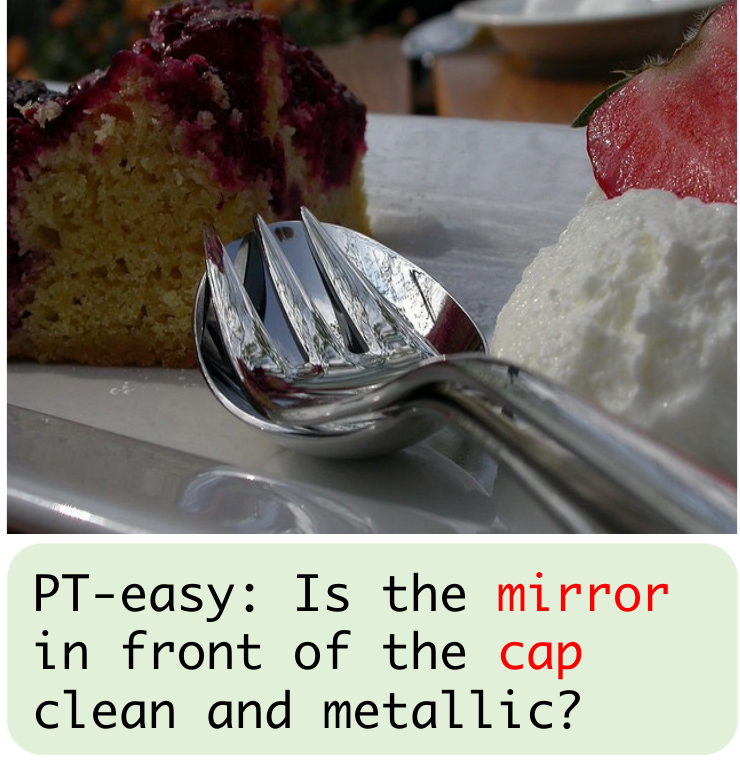} &
        \includegraphics[width=0.22\textwidth,valign=t]{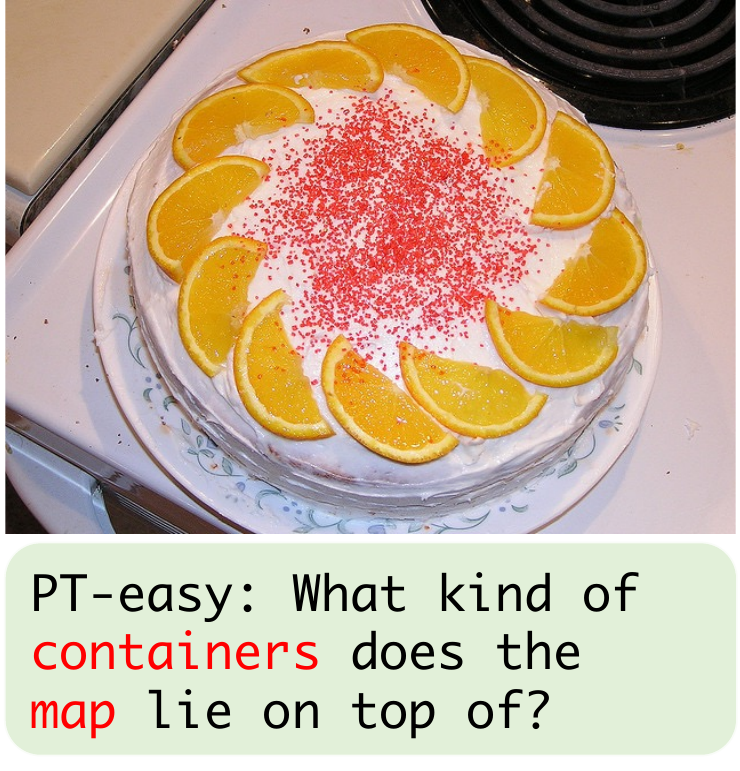} &
        \includegraphics[width=0.22\textwidth,valign=t]{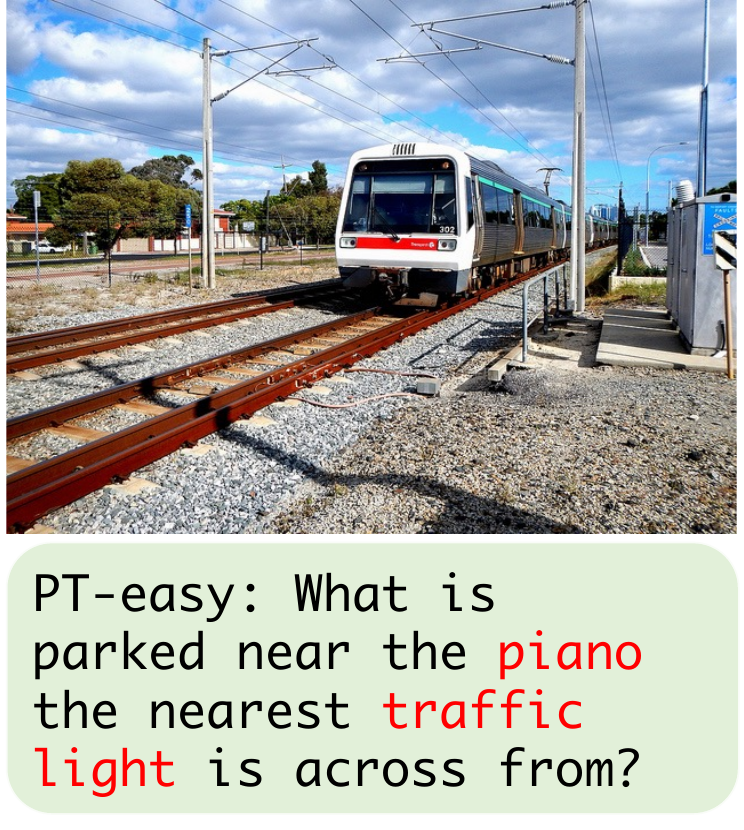} \\
        (m) & (n) & (o) & (p) \\
    \end{tabular}
    \caption{More examples from RGQA dataset across 4 different subsets.}
    \label{fig:more_examples}
\end{figure*}

We show more examples from RGQA in Figure~\ref{fig:more_examples}, where (a-d), (e-h), (i-l) and (m-p) are CLIP-Hard, PT-Hard, CLIP-Easy, and PT-Easy, respectively.

\vspace{-5pt}
\section{Social Impact and Future Work}
\noindent\textbf{Social Impact:} In this work, we study the problem of RVQA and proposed a new dataset containing UQs to evaluate the existing VQA classifiers. Note that the goal is not to falsify the VQA system, but to evaluate and improve the robustness of the system to UQs. We plan to expand the proposed dataset with more diverse types of UQ and include more annotations, such as the category of UQs. We hope this dataset will encourage more research on RVQA. 

\noindent\textbf{Future Work:}
While we have evaluated the open-ended VQA model (See BLIP experiment) on the proposed dataset, we would like to further explore open-ended VQA in the future, especially by allowing the model to explain why a question is a UQ. However, this might require further annotations. Hence, we see this as a direction for future work. Another direction is to compare the distribution between the proposed RGQA dataset and the unanswerable question collected from the real-world system.

\definecolor{LightCyan}{rgb}{0.88,1,1}
\begin{table*}[ht!]
\centering
\caption{Comparison between different RVQA approaches. Larger AUAF and smaller FPR@95 is better. Cells with light cyan background denote training with pseudo UQs. 
}
\scalebox{0.8}{
\begin{tabular}{|c|ccc|ccc|ccc|ccc|c|}
\hline
\multicolumn{1}{|c|}{}&\multicolumn{3}{c|}{CLIP-Easy}&\multicolumn{3}{c|}{CLIP-Hard}&\multicolumn{3}{c|}{PT-Easy}&\multicolumn{3}{c|}{PT-Hard}&Avg.\\
RVQA Approaches & AUAF & FF95$\downarrow$ & FACC & AUAF & FF95$\downarrow$ & FACC & AUAF & FF95$\downarrow$ & FACC & AUAF & FF95$\downarrow$ & FACC & AUAF\\
\hline
\hline
\multicolumn{1}{|c}{ }& \multicolumn{13}{c|}{BUTD~\cite{butd}}\\
\hline
FRCNN&  33.58 & 93.28 & 53.50 & 30.73 & 93.94 & 53.08 & 31.43 & 93.77 & 53.02 & 26.94 & 94.65 & 51.31 & 30.67\\
MSP&  38.45 & 64.75 & 53.50 & 36.13 & 79.14 & 53.08 & 37.83 & 66.05 & 53.02 & 33.60 & 83.11 & 51.31 & 36.50 \\
ODIN&  38.47 & 64.66 & 53.53 & 36.14 & 79.19 & 53.11 & 37.80 & 66.14 & 52.97 & 33.60 & 83.41 & 51.33 & 36.50 \\
Maha&  30.05 & 80.66 & 48.76 & 25.75 & 92.16 & 48.42 & 25.34 & 94.90 & 47.70 & 23.93 & 95.43 & 46.39 & 26.26 \\
Energy &  38.47 & 64.14 & 53.50 & 36.19 & 79.42 & 53.08 & 37.77 & 66.12 & 53.02 & 33.67 & 82.99 & 51.31 & 36.52 \\
\cellcolor{LightCyan}Q-C & \cellcolor{LightCyan}53.04 & \cellcolor{LightCyan} 3.48 & \cellcolor{LightCyan} 53.50 & \cellcolor{LightCyan} 36.20 & \cellcolor{LightCyan} 69.25 & \cellcolor{LightCyan} 53.08  & \cellcolor{LightCyan} 47.14 & \cellcolor{LightCyan} 42.18 & \cellcolor{LightCyan} 53.02 & \cellcolor{LightCyan} 29.06 & \cellcolor{LightCyan} 85.65 & \cellcolor{LightCyan} 51.31 & \cellcolor{LightCyan} 41.36
\\
\cellcolor{LightCyan}Resample & \cellcolor{LightCyan} 40.25 & \cellcolor{LightCyan} 65.23 & \cellcolor{LightCyan} 56.20 & \cellcolor{LightCyan} 37.73 & \cellcolor{LightCyan} 79.64 & \cellcolor{LightCyan} 55.45 & \cellcolor{LightCyan} 39.54 & \cellcolor{LightCyan} 66.43 & \cellcolor{LightCyan} 55.41 & \cellcolor{LightCyan} 34.78 & \cellcolor{LightCyan} 83.73 & \cellcolor{LightCyan} 53.79 & \cellcolor{LightCyan} 38.07
\\
\hline
\cellcolor{LightCyan}RP(w/ hard UQ)& \cellcolor{LightCyan}  43.74
& \cellcolor{LightCyan}  66.33
& \cellcolor{LightCyan} 56.04
& \cellcolor{LightCyan} 43.27
& \cellcolor{LightCyan} 70.38
& \cellcolor{LightCyan} 55.40
& \cellcolor{LightCyan} 37.62
& \cellcolor{LightCyan} 81.98
& \cellcolor{LightCyan} 55.21
& \cellcolor{LightCyan} 36.17
& \cellcolor{LightCyan} 84.97
& \cellcolor{LightCyan} 53.81
& \cellcolor{LightCyan} 40.2
\\
\cellcolor{LightCyan}RP(Ours) & \cellcolor{LightCyan} 56.31 & \cellcolor{LightCyan} 1.82 & \cellcolor{LightCyan} 56.64 & \cellcolor{LightCyan} 44.09 & \cellcolor{LightCyan} 56.57 & \cellcolor{LightCyan} 55.66 & \cellcolor{LightCyan} 50.51 & \cellcolor{LightCyan} 27.41 & \cellcolor{LightCyan} 55.03 & \cellcolor{LightCyan} 37.18 & \cellcolor{LightCyan} 80.38 & \cellcolor{LightCyan} 53.88 & \cellcolor{LightCyan} 47.02
\\

\cellcolor{LightCyan}Mix(Ours)& \cellcolor{LightCyan} 56.85 & \cellcolor{LightCyan} 1.65 & \cellcolor{LightCyan} 57.17 & \cellcolor{LightCyan} 44.70 & \cellcolor{LightCyan} 58.84 & \cellcolor{LightCyan} 56.59 & \cellcolor{LightCyan} 51.27 & \cellcolor{LightCyan} 29.28 & \cellcolor{LightCyan} 55.99 & \cellcolor{LightCyan} 37.59 & \cellcolor{LightCyan} 83.41 & \cellcolor{LightCyan} \bf{55.24} & \cellcolor{LightCyan} 47.60
\\

\cellcolor{LightCyan}Ens(Ours) & \cellcolor{LightCyan} \bf{57.25} & \cellcolor{LightCyan} \bf{1.31} & \cellcolor{LightCyan} \bf{57.50} & \cellcolor{LightCyan} \bf{45.46} & \cellcolor{LightCyan} \bf{56.04} & \cellcolor{LightCyan} \bf{56.90} & \cellcolor{LightCyan} \bf{51.95} & \cellcolor{LightCyan} \bf{24.69} & \cellcolor{LightCyan} \bf{56.02} & \cellcolor{LightCyan} \bf{38.46} & \cellcolor{LightCyan} \bf{80.08} & \cellcolor{LightCyan} 54.85 & \cellcolor{LightCyan} \bf{48.28}
\\
\hline
\multicolumn{1}{|c}{ }& \multicolumn{13}{c|}{UNITER~\cite{uniter}}\\
\hline
FRCNN&  35.81 & 93.28 & 57.08 & 33.09 & 93.93 & 57.10 & 33.67 & 93.77 & 56.82 & 28.82 & 94.68 & 55.08 & 32.84
\\
MSP&   40.03 & 73.15 & 57.08 & 39.42 & 80.48
& 57.10 & 41.45 & 61.76 & 56.82 & 35.17 & 83.52 & 55.08 & 39.01
\\
ODIN&  40.04 & 73.22 & 57.12 & 39.43 & 80.48 & 57.15 & 41.45 & 61.83 & 56.85 & 35.16 & 83.54 & 55.06 & 39.02
\\
Maha&  37.52 & 67.07 & 55.38 & 33.74 & 81.09 & 54.88 & 35.87 & 63.98 & 54.68 & 31.68 & 85.78 & 52.80 & 34.70
\\
Energy&  40.10 & 71.45 & 57.08 & 39.42 & 79.78 & 57.10 & 41.41 & 
61.31 & 56.82 & 35.19 & 83.63 & 55.08 & 39.03
\\
\cellcolor{LightCyan}Q-C& \cellcolor{LightCyan} 56.61 & \cellcolor{LightCyan} 3.53 & \cellcolor{LightCyan} 57.08 & \cellcolor{LightCyan} 38.67 & \cellcolor{LightCyan} 69.56 & \cellcolor{LightCyan} 57.10 & \cellcolor{LightCyan} 50.12 & \cellcolor{LightCyan} 45.64 & \cellcolor{LightCyan} 56.82 & \cellcolor{LightCyan} 30.93 & \cellcolor{LightCyan} 86.18 & \cellcolor{LightCyan} 55.08 & \cellcolor{LightCyan} 44.08
\\
\cellcolor{LightCyan}Resample& \cellcolor{LightCyan} 58.66 & \cellcolor{LightCyan} 0.755 & \cellcolor{LightCyan} 58.85 & \cellcolor{LightCyan} 48.08 & \cellcolor{LightCyan} 47.10 & \cellcolor{LightCyan} 57.60 & \cellcolor{LightCyan} 53.65 & \cellcolor{LightCyan} 22.42 & \cellcolor{LightCyan} 57.48 & \cellcolor{LightCyan} 39.84 & \cellcolor{LightCyan} 73.46 & \cellcolor{LightCyan} 55.33 & \cellcolor{LightCyan} 50.05
\\
\hline
\cellcolor{LightCyan}RP(w/ hard UQ)& \cellcolor{LightCyan}  44.92
& \cellcolor{LightCyan} 70.71
& \cellcolor{LightCyan} 59.02
& \cellcolor{LightCyan} 47.14
& \cellcolor{LightCyan} 59.81
& \cellcolor{LightCyan} 57.91
& \cellcolor{LightCyan} 41.89
& \cellcolor{LightCyan} 70.89
& \cellcolor{LightCyan} 58.36
& \cellcolor{LightCyan} 37.92
& \cellcolor{LightCyan} 80.19
& \cellcolor{LightCyan} 55.70
& \cellcolor{LightCyan} 42.96

\\

\cellcolor{LightCyan}RP(Ours)& \cellcolor{LightCyan} 58.35 & \cellcolor{LightCyan} 0.615 & \cellcolor{LightCyan} 58.49 & \cellcolor{LightCyan} 48.37 & \cellcolor{LightCyan} 47.08 & \cellcolor{LightCyan} 57.69 & \cellcolor{LightCyan} 54.42 & \cellcolor{LightCyan} 20.43 & \cellcolor{LightCyan} 57.83 & \cellcolor{LightCyan} 40.27 & \cellcolor{LightCyan} 73.20 & \cellcolor{LightCyan} 55.44 & \cellcolor{LightCyan} 50.35
\\

\cellcolor{LightCyan}Mix(Ours)& \cellcolor{LightCyan} 59.08 & \cellcolor{LightCyan} 0.615 & \cellcolor{LightCyan} 59.37 & \cellcolor{LightCyan} 49.00 & \cellcolor{LightCyan} 47.00 & \cellcolor{LightCyan} 58.06 & \cellcolor{LightCyan} 54.63 & \cellcolor{LightCyan} 21.44 & \cellcolor{LightCyan} 58.08 & \cellcolor{LightCyan} 41.50 & \cellcolor{LightCyan} 73.29 & \cellcolor{LightCyan} 56.68 & \cellcolor{LightCyan} 51.05

\\

\cellcolor{LightCyan}Ens(Ours)& \cellcolor{LightCyan} \bf{59.62} & \cellcolor{LightCyan} \bf{0.58} & \cellcolor{LightCyan} \bf{59.82} & \cellcolor{LightCyan} \bf{49.65} & \cellcolor{LightCyan} \bf{46.71} & \cellcolor{LightCyan} \bf{58.84} & \cellcolor{LightCyan} \bf{55.79} & \cellcolor{LightCyan} \bf{20.08} & \cellcolor{LightCyan} \bf{59.11} & \cellcolor{LightCyan} \bf{42.14} & \cellcolor{LightCyan} \bf{72.71} & \cellcolor{LightCyan} \bf{57.17} & \cellcolor{LightCyan} \bf{51.8}
\\
\hline
\multicolumn{1}{|c}{ }& \multicolumn{13}{c|}{LXMERT~\cite{tan2019lxmert}}\\
\hline
FRCNN&  38.43 & 93.21 & 60.87 & 35.22 & 93.88 & 60.49 & 35.73 &
93.72 & 59.94 & 31.00 & 94.62 & 58.76 & 35.09
\\
MSP&  42.39 & 76.25 & 60.87 & 42.60 & 78.92 & 60.49 & 47.30 & 61.79 & 59.94 & 38.12 & 85.14 & 58.76 & 42.60
\\
ODIN&  42.41 & 76.43 & 60.92 & 42.59 & 78.96 & 60.46 & 47.33 & 61.97 & 59.97 & 38.12 & 84.78 & 58.73 & 42.61
\\
Maha&  57.68 & 9.79 & 58.98 & 44.96 & 61.09 & 58.16 & 49.44 & 44.43 & 57.27 & 39.25 & 75.25 & 56.29 & 47.83
\\
Energy&  38.76 & 76.88 & 60.87 & 42.11 & 78.85 & 60.49 & 47.00 & 61.84 & 59.94 & 37.90 & 85.53 & 58.76 & 41.44

\\
\cellcolor{LightCyan}Q-C& \cellcolor{LightCyan} 60.39 & \cellcolor{LightCyan} 3.42 & \cellcolor{LightCyan} 60.87 & \cellcolor{LightCyan} 41.31 & \cellcolor{LightCyan} 68.72 & \cellcolor{LightCyan} 60.49 & \cellcolor{LightCyan} 53.11 & \cellcolor{LightCyan} 44.50 & \cellcolor{LightCyan} 59.94 & \cellcolor{LightCyan} 33.18 & \cellcolor{LightCyan} 85.65 & \cellcolor{LightCyan} 58.76 & \cellcolor{LightCyan} 46.99
\\
\cellcolor{LightCyan}Resample& \cellcolor{LightCyan} 60.47  & \cellcolor{LightCyan} 0.58 & \cellcolor{LightCyan}  60.66 & \cellcolor{LightCyan} 50.80 & \cellcolor{LightCyan} 46.49 & \cellcolor{LightCyan} 60.37 & \cellcolor{LightCyan} 55.74 & \cellcolor{LightCyan} 25.30 & \cellcolor{LightCyan} 59.84 & \cellcolor{LightCyan} 42.18 & \cellcolor{LightCyan} 76.78 & \cellcolor{LightCyan} 58.27 & \cellcolor{LightCyan} 52.29 
\\
\hline
\cellcolor{LightCyan}RP(w/ hard UQ)& \cellcolor{LightCyan} 53.60 & \cellcolor{LightCyan} 40.44 & \cellcolor{LightCyan} 60.15 & \cellcolor{LightCyan} 51.39 & \cellcolor{LightCyan} 47.80 & \cellcolor{LightCyan} 59.40 & \cellcolor{LightCyan} 46.95 & \cellcolor{LightCyan} 57.51 & \cellcolor{LightCyan} 58.74 & \cellcolor{LightCyan} 42.96 & \cellcolor{LightCyan} \bf68.56 & \cellcolor{LightCyan} 57.17 & 
\cellcolor{LightCyan} 48.72
\\

\cellcolor{LightCyan}RP(Ours)& \cellcolor{LightCyan} 60.51 & \cellcolor{LightCyan} 0.527 & \cellcolor{LightCyan} 60.66 & \cellcolor{LightCyan} 51.49 & \cellcolor{LightCyan} 45.02 & \cellcolor{LightCyan} 60.69 & \cellcolor{LightCyan} 56.08 & \cellcolor{LightCyan} 23.18 & \cellcolor{LightCyan} 59.74 & \cellcolor{LightCyan} 42.53 & \cellcolor{LightCyan} 75.78 & \cellcolor{LightCyan} 58.37 & \cellcolor{LightCyan} 52.65
\\

\cellcolor{LightCyan}Mix(Ours)& \cellcolor{LightCyan} 60.79 & \cellcolor{LightCyan} \bf{0.298} & \cellcolor{LightCyan} 61.03 & \cellcolor{LightCyan} 51.91 & \cellcolor{LightCyan} 43.43 & \cellcolor{LightCyan} 60.67 & \cellcolor{LightCyan} 56.83 & \cellcolor{LightCyan} 22.58 & \cellcolor{LightCyan} 60.40 & \cellcolor{LightCyan} 43.56 & \cellcolor{LightCyan} 73.02 & \cellcolor{LightCyan} 58.64 & \cellcolor{LightCyan} 53.27
\\

\cellcolor{LightCyan}Ens(Ours)& \cellcolor{LightCyan} \bf{61.03} & \cellcolor{LightCyan} 0.351 & \cellcolor{LightCyan} \bf{61.19} & \cellcolor{LightCyan} \bf{52.42} & \cellcolor{LightCyan} \bf{42.84} & \cellcolor{LightCyan} \bf{61.19} & \cellcolor{LightCyan} \bf{56.90} & \cellcolor{LightCyan} \bf{22.40} & \cellcolor{LightCyan}  \bf{60.47} & \cellcolor{LightCyan} \bf{43.75} & \cellcolor{LightCyan} {73.01} & \cellcolor{LightCyan} \bf{58.83} & \cellcolor{LightCyan} \bf{53.52}
\\
\hline
\end{tabular}
}
\label{tab:main_full}
\end{table*}

\end{document}